\ifdefined\pdfoutput\pdfoutput=1\fi
\documentclass[11pt,a4paper]{article}

\PassOptionsToPackage{svgnames}{xcolor}
\usepackage{style}

\usepackage[numbers,sort&compress]{natbib}
\usepackage{graphicx}
\usepackage[font=small,labelfont=bf]{caption}
\usepackage{subcaption}
\usepackage{booktabs}
\usepackage{colortbl}
\usepackage{multirow}
\usepackage{bigstrut}
\usepackage{float}
\usepackage[all]{hypcap}
\usepackage{pifont}
\usepackage{amsmath}
\usepackage{mathtools}
\usepackage{mathrsfs}
\usepackage{nicefrac}

\usepackage{algorithm}
\usepackage{algpseudocode}

\usepackage{microtype}
\usepackage{enumitem}
\usepackage{cleveref}
\usepackage{bxcoloremoji}
\usepackage{url}
\usepackage{wrapfig}
\usepackage{lipsum}
\usepackage{stackengine}
\usepackage{adjustbox}
\usepackage{rotating}
\usepackage{makecell}

\begin{document}

\Cover{
  title={Worldscape-MoE: A Unified Mixture-of-Experts World Model for Scalable Heterogeneous Action Control}, 
  authors={Jianjie Fang\textsuperscript{1,*}, Yongyan Xu\textsuperscript{1,*}, Ziyou Wang\textsuperscript{1,*}, Chen Gao\textsuperscript{1,*,\textdagger}, Yuchao Huang\textsuperscript{1}, Zhaolu Wang\textsuperscript{1}, Rongze Tang\textsuperscript{1}, Mingyuan Jia\textsuperscript{1}, Baining Zhao\textsuperscript{1}, Weichen Zhang\textsuperscript{1}, Xin Zhang\textsuperscript{2}, Haisheng Su\textsuperscript{2}, Yu Shang\textsuperscript{1}, Wei Wu\textsuperscript{2}, Xinlei Chen\textsuperscript{1}, Yong Li\textsuperscript{1,\textdagger}},
  affils={\textsuperscript{1}Tsinghua University \quad \textsuperscript{2}Manifold AI\\
  \textbf{\textsuperscript{*}Equal contribution. \quad \textsuperscript{\textdagger}Corresponding authors.}},
  abstract = {\textbf{Abstract.} World models are rapidly becoming a core infrastructure for embodied intelligence and interactive agents: they provide controllable simulators in which agents can perceive, act, forecast, and acquire scalable experience. Yet current video generation world models are still organized around isolated control interfaces, such as camera trajectories, robot actions, or hand-joint signals. This fragmentation is increasingly a scaling bottleneck. The central challenge is not the absence of controllable generators, but the lack of a unified and extensible learning framework that can absorb heterogeneous action supervision while preserving a shared model of world dynamics. In this work, we introduce \textbf{Worldscape-MoE}, a Mixture-of-Experts world model built on Diffusion Transformers for scalable heterogeneous action control. Our key observation is that different controls specify different interfaces to the same underlying world: although their representations differ, they constrain shared physical regularities, scene dynamics, and interaction semantics. Worldscape-MoE operationalizes this observation through modality-aware control injection, shared and control-specific experts, and a progressive MoE tuning strategy that supports continual extension to new action modalities. Experiments across locomotion, robotic manipulation, and egocentric hand control show that heterogeneous supervision improves rather than interferes with individual control capabilities. Worldscape-MoE achieves strong results on WorldArena, improves locomotion and hand-control metrics, exhibits robust out-of-distribution generalization, and demonstrates scaling behavior as additional control data and experts are integrated.},
  date={July 5, 2026},
  codeurl={https://worldscape-moe.com},
  pageurl={https://worldscape-moe.com},
  headerlogo={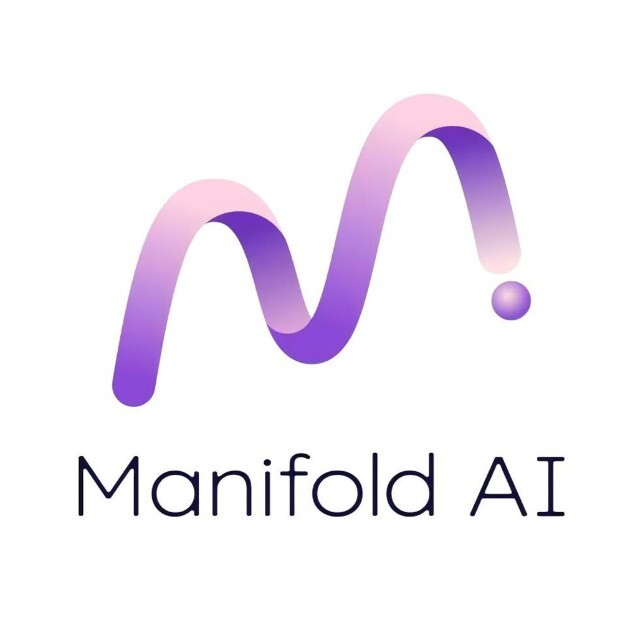},
  headerdate={July 5, 2026}
}

\section{Introduction}
\label{Intro}
World models are moving from a research prototype toward a general infrastructure for embodied intelligence and interactive agents. Built on recent progress in high-quality video generation~\cite{wan2025wan,wu2025hunyuanvideo,yang2024cogvideox}, they can simulate how visual scenes evolve under actions, support interactive exploration, and provide scalable experience beyond what can be collected directly in the physical world~\cite{ding2025understanding,team2025gigaworld}. This role is especially important for agents that must reason through the consequences of actions before execution: a world model is no longer merely a video generator, but a controllable substrate for perception, planning, evaluation, and policy learning.

Despite this progress, current world models remain largely fragmented by the form of control they accept. Interactive environment models, such as Genie~\cite{bruce2024genie}, the Matrix-Game series~\cite{he2025matrix,wang2026matrix,zhang2025matrix}, and HY-World~\cite{sun2025worldplay,hy2026hy}, primarily focus on camera or navigation-like controls. Embodied world models, such as Ctrl-World~\cite{guo2025ctrl} and Cosmos-Predict~\cite{ali2025world}, emphasize robot-action-conditioned manipulation~\cite{shang2026worldarena}. Egocentric interaction models, including Hand2World~\cite{wang2026hand2world} and Generated Reality~\cite{xie2026generated}, introduce hand-joint or hand-pose controls for first-person human interaction. Despite their individual merits, these lines of work remain largely siloed by control modality. Camera-control models, embodied manipulation models, and egocentric interaction models are each developed with distinct architectures, training objectives, and data pipelines. This siloed structure means that data, model capacity, and learned world knowledge are repeatedly partitioned along control boundaries, even when the underlying scenes and physical regularities they describe are fundamentally shared.

\begin{figure}[t]
	\centering
    \vspace{-10pt}
	\includegraphics[width=\linewidth]{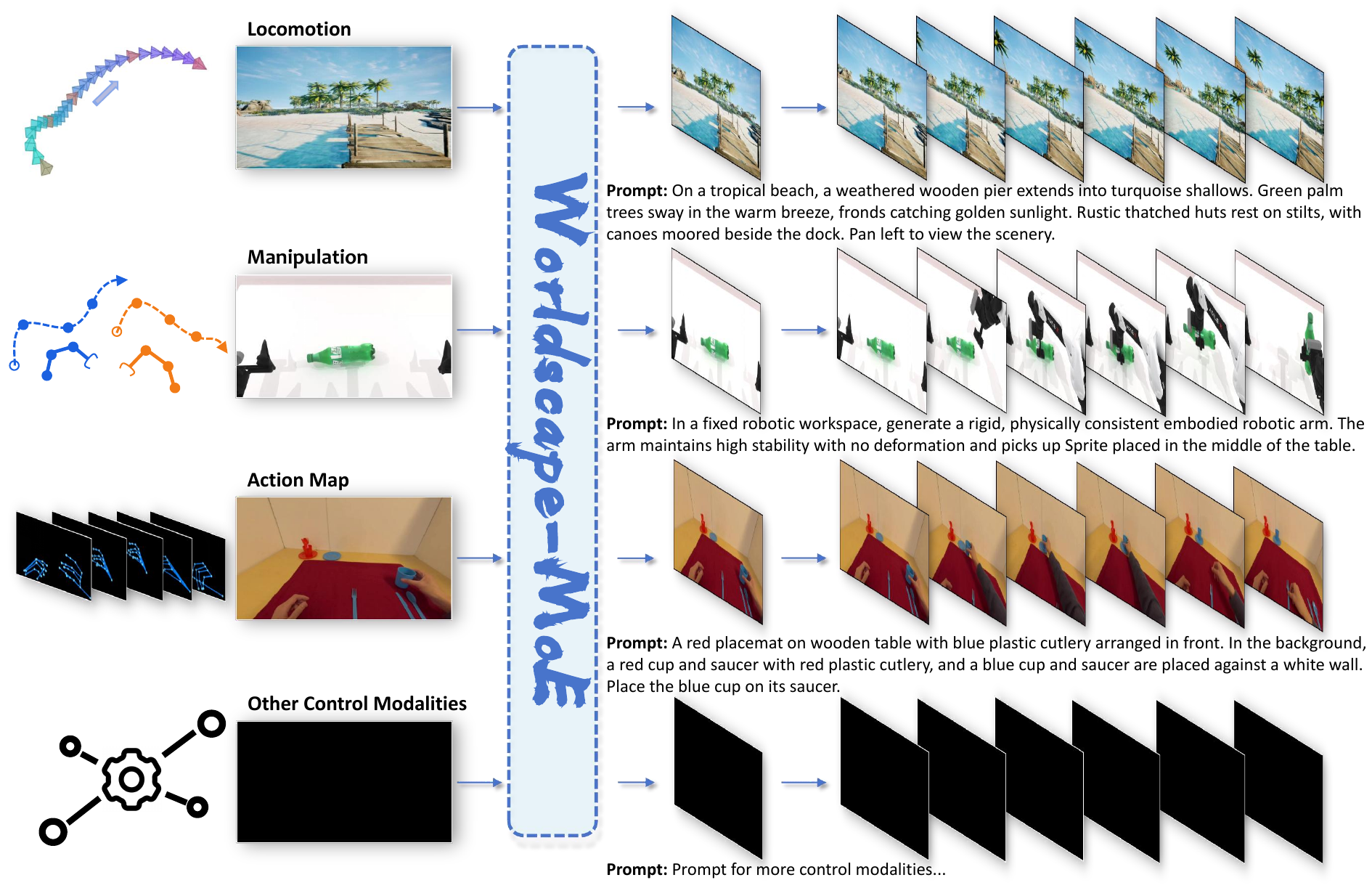} 
	\caption{\textbf{Worldscape-MoE Overview.} Worldscape-MoE supports three mainstream control modalities: \textbf{Locomotion} for trajectory-conditioned world navigation, \textbf{Manipulation} for robot-action-conditioned embodied tasks, and \textbf{Action Map} for hand-joint-conditioned egocentric interaction generation. The framework can also be extended to additional control injection settings.}
	\label{fig:overview}
    \vspace{-10pt}
\end{figure}

This limitation is fundamentally structural. The central challenge is not the absence of controllable world models for individual modalities, but the lack of a scalable learning framework capable of unifying heterogeneous action supervision within a shared representation of world dynamics. Camera trajectories, robot joint-action vectors, and hand-joint maps differ markedly in their representational form and conditioning pathway, yet they all constrain the same latent world: objects persist, interactions obey physical regularities, motion trajectories remain temporally coherent, and actions produce causally plausible scene changes. This motivates a design principle centered on factorization rather than isolation: instead of building separate models for separate controls, a world model should decompose the problem into shared world dynamics and control-specific action interpretation.

To examine whether heterogeneous action-control world modeling can be learned in a unified and scalable manner, we investigate four research questions.
\begin{itemize}
        \item \textbf{RQ1: Can a single world model support multiple heterogeneous action controls without sacrificing the performance of each individual control setting?}
        Since existing controllable world models are usually optimized for one control interface, it remains unclear whether camera trajectories, robot actions, and hand-joint controls can be learned within the same model without causing cross-control interference. Answering this question is necessary to determine whether unified heterogeneous training is a viable alternative to training separate single-control models.
        \item \textbf{RQ2: Does the MoE structure learn meaningful expert routing and specialization, or is the gain purely from increased capacity?}
        A larger model may improve performance simply by adding capacity, but heterogeneous world modeling requires a more structured form of capacity allocation. We therefore examine whether the shared expert captures common world dynamics while control-specific experts specialize to the action semantics of each modality.
        \item \textbf{RQ3: Does heterogeneous training improve scalability when new controls and data sources are added?}
        The central motivation of heterogeneous control learning is to turn fragmented datasets into a growing training resource. This raises the question of whether adding new controls improves overall capability, or whether it introduces optimization conflicts that degrade previously learned controls.
        \item \textbf{RQ4: Does the MoE-based unified world model generalize beyond the training distributions and support loco-manipulation scenarios?}
        A unified world model should not only fit the training controls independently, but also learn transferable world regularities that support out-of-distribution scenes and compositions of actions. We therefore test whether the learned representation can generalize to novel environments and generate coupled loco-manipulation behavior.
\end{itemize}

To answer these questions, we introduce \textbf{Worldscape-MoE}, shown in Fig.~\ref{fig:overview}, a unified Mixture-of-Experts world model for scalable heterogeneous action control. Worldscape-MoE uses a shared diffusion-transformer backbone with control-dependent injection paths and replaces dense feed-forward computation with a control-aware MoE structure. Each training sample activates a shared expert, which accumulates cross-control world knowledge, together with a modality-specific expert, which preserves the precision of the corresponding action interface. In this design, MoE is not used only as a capacity-scaling device; it is used as a mechanism for separating the shared laws of world evolution from the modality-specific details of action conditioning.

We further develop \textbf{Worldscape-MoE Tuning}, a progressive training strategy for extending a pretrained video diffusion model to multiple action controls. The strategy introduces heterogeneous control branches and experts in stages, initializes new experts from the current shared expert, and trains shared parameters with a conservative learning rate. This makes the model extensible: new control modalities can be absorbed without forcing all previous controls to be relearned from scratch. The resulting framework turns heterogeneous world-modeling data from a set of disconnected silos into a joint training resource.

In summary, our main contributions are as follows:
\begin{itemize}
        \item We formulate heterogeneous action-control world modeling as a scaling problem and identify the key obstacle as the lack of a unified learning framework rather than the lack of individual control models.
        \item We propose \textbf{Worldscape-MoE}, a DiT-based world model that combines modality-aware control injection, shared experts, and control-specific experts to learn from locomotion, robotic manipulation, and egocentric hand-control data in one architecture.
        \item We present \textbf{Worldscape-MoE Tuning}, a progressive and extensible training procedure that allows the shared expert to absorb cross-control world knowledge while new experts specialize to newly introduced control modalities.
        \item We conduct extensive experiments to answer the above research questions across heterogeneous control performance, expert routing, MoE effectiveness, scalability, out-of-distribution generalization, and coupled loco-manipulation. The results show consistent gains over dense mixed training and strong performance on manipulation, locomotion, and hand-motion evaluations.
\end{itemize}

\section{Methodology}
\label{Sec:method}
Worldscape-MoE is designed around a simple principle: heterogeneous controls should provide different interfaces to a shared world model, rather than define separate world models. The framework therefore separates three responsibilities. First, each control modality is represented through an injection pathway that matches its structure. Second, a shared diffusion-transformer backbone maintains common spatiotemporal and physical priors. Third, sparse experts separate cross-control world knowledge from modality-specific action interpretation. This design allows the model to scale with heterogeneous data while avoiding the destructive interference that often appears in direct dense mixed training.

\subsection{Heterogeneous Action-Control World Modeling}
\label{Sec:formulation}

We consider video generation world modeling under heterogeneous action supervision. Given the current observation frame $O_t$, a textual world description $C_t$, and an action condition $A_t$, the model predicts the subsequent world observation:
\begin{equation}
	W_{t+1} = f_{\theta}(O_t, C_t, A_t),
\end{equation}
where $A_t$ may take different forms of control input, including camera trajectory $A_{\mathrm{traj}}$, hand-joint representation $A_{\mathrm{hand}}$, dual-arm robot control vector $A_{\mathrm{act}}$, and other extensible action modalities $A_{\mathrm{other}}$, i.e.,
\begin{equation}
	A_t \in \left\{A_{\mathrm{traj}},\; A_{\mathrm{hand}},\; A_{\mathrm{act}},\; A_{\mathrm{other}}\right\}.
\end{equation}
The predicted $W_{t+1}$ is a future video segment composed of consecutive frames, and its last frame $O_{t+1}$ is fed back as the initial observation for the next prediction step, enabling long-horizon autoregressive world generation. Specifically, $A_{\mathrm{traj}}$ is a temporal camera-motion sequence, $A_{\mathrm{traj}} \in \mathbb{R}^{T_v \times d_{\mathrm{traj}}}$, which can be converted into dense camera-control features. $A_{\mathrm{hand}}$ is a dense action-map sequence, $A_{\mathrm{hand}} \in \mathbb{R}^{T_v \times H \times W \times C_{\mathrm{hand}}}$, while $A_{\mathrm{act}}$ is a low-dimensional manipulation sequence, $A_{\mathrm{act}} \in \mathbb{R}^{T_a \times d_{\mathrm{act}}}$. In our implementation, $T_a=17$ and $d_{\mathrm{act}}=14$, giving a $17 \times 14$ dual-arm action tensor per prediction window.

This setting imposes three design requirements. \textbf{Control faithfulness} requires the model to preserve the semantics of each action interface, because trajectory control, hand control, and robot control affect the scene through different causal paths. \textbf{Shared world learning} requires the model to reuse visual dynamics, object permanence, contact regularities, and temporal coherence across modalities. \textbf{Extensibility} requires the framework to incorporate new controls without rebuilding the entire model. Worldscape-MoE is built to satisfy these requirements jointly.

\subsection{Heterogeneous Control Dataset Construction}
\label{Sec:data_pipline}
Generating worlds under heterogeneous control requires carefully curated and diverse training data, including high-quality control annotations, informative captions, and temporally coherent frame sequences. Our training corpus consists of three categories of data, corresponding to three different interaction modalities. More detailed descriptions of data processing are provided in Appendix~\ref{app:data}.

\noindent\textbf{Camera-control data.}
For camera-trajectory-conditioned world generation, we curate data from multiple sources. \textbf{(I) Real-world data:} we use clips from RealEstate10K~\cite{zhou2018stereo} and DL3dv-10k~\cite{ling2024dl3dv} to cover diverse real-world scenes. \textbf{(II) Simulation data:} following the camera file design and data collection protocol of iWorld-Bench~\cite{fang2026iworldbench}, we perform stylized data collection in Unreal Engine using 81 camera motion patterns across diverse environments. \textbf{(III) Diverse world data:} we further select clips from Sekai~\cite{li2025sekai} and SpatialVid~\cite{wang2025spatialvid}, and re-annotate their trajectories using VIPE~\cite{huang2025vipe}. All collected data are then passed through a unified annotation pipeline. For textual descriptions, we adopt the caption annotation strategy introduced in Matrix-Game 3.0~\cite{wang2026matrix}.

\noindent\textbf{Robotic embodied data.}
For robot-action-conditioned embodied world modeling, we process a large collection of embodied clips. The data construction pipeline consists of two stages. \textbf{(I) Simulation data generation:} we build a large-scale simulation pipeline for robotic manipulation scenarios on the RoboTwin~\cite{mu2025robotwin,chen2025robotwin} platform. This part covers diverse robotic tasks and object-interaction scenarios. We segment the generated sequences along the temporal dimension into fixed-length samples using a sliding window. \textbf{(II) Data generalization:} to improve the diversity and generalization of embodied tasks, we further develop an embodied data augmentation pipeline based on Seedance 2.0~\cite{seedance2026seedance} and Qwen3.6-Plus~\cite{qwen36plus}, which varies materials, colors, and manipulated objects. For captions, we directly use the embodied textual descriptions provided by the RoboTwin platform. Additionally, we incorporate the LIBERO~\cite{liu2023libero} dataset for single-arm manipulation, processing its demonstrations into action-conditioned video clips with aligned task language and action trajectories to support the modality extension experiments described in Appendix~\ref{app:libero_single_arm}.

\noindent\textbf{Egocentric hand data.}
For hand-joint-conditioned world generation, we process a collection of egocentric hand-interaction clips. To construct the hand-control modality, we sample clips from large-scale egocentric video datasets, including EgoDex~\cite{hoque2025egodex} and Ego4D~\cite{grauman2022ego4d}. Since the datasets differ in annotation protocols and hand supervision formats, we apply HaMeR~\cite{pavlakos2024reconstructing} to all selected frames to perform monocular 3D hand reconstruction in a unified manner. This yields hand joints for each hand, which are then projected into 2D pixel coordinates using the camera intrinsics and extrinsics. Finally, we rasterize them into hand-joint action maps that are spatially aligned with the video frames at the pixel level, serving as the supervision signal for the hand-control expert. For text conditioning, we use the captions provided by the original datasets.

\subsection{Architecture of Worldscape-MoE}
\label{Sec:Uni_mdoel}
\begin{figure}[t]
	\centering
     \includegraphics[width=\linewidth]{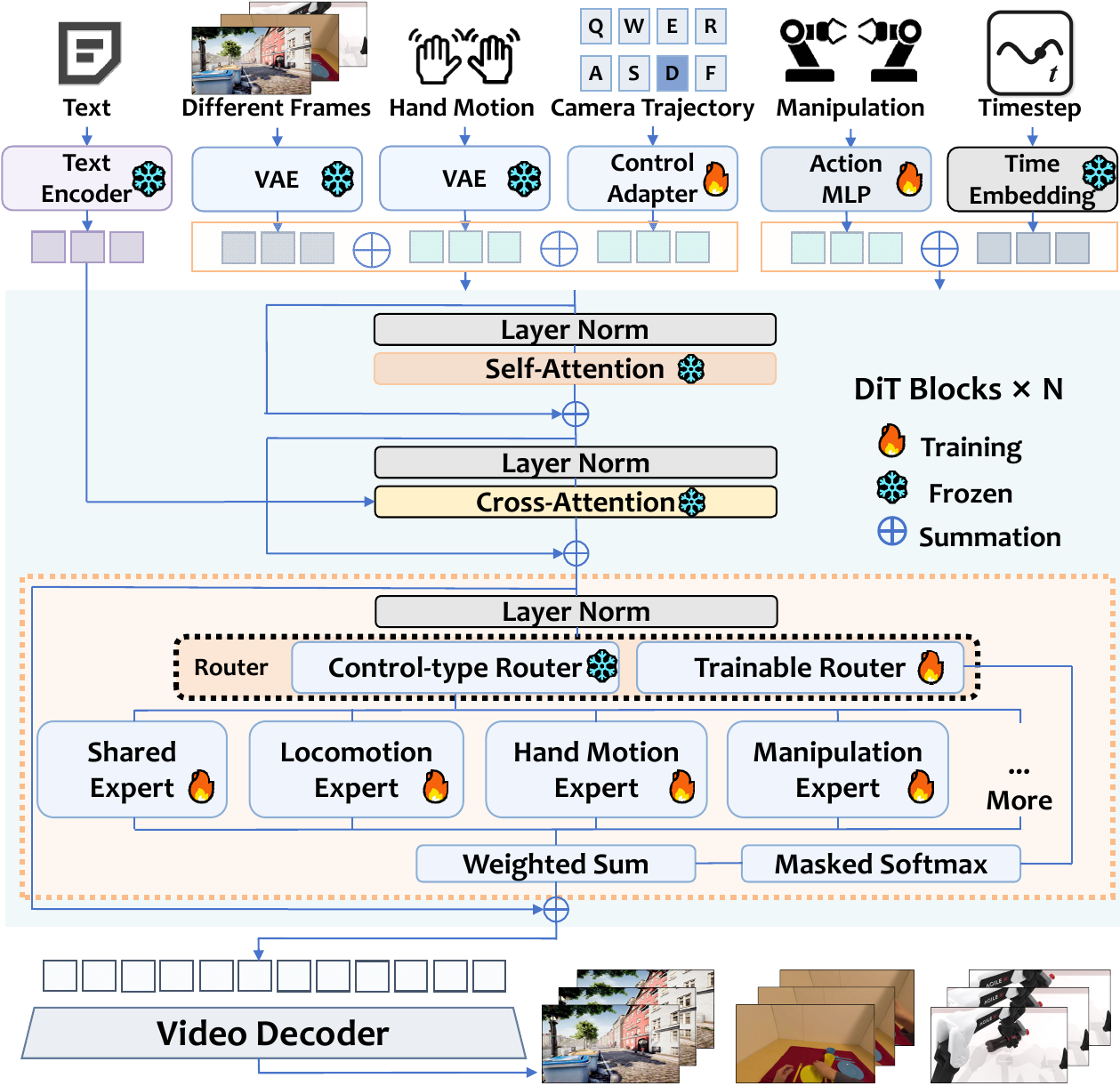}
	\caption{\textbf{Worldscape-MoE Architecture.} Given the current world observation and different forms of supervisory control, our framework generates world dynamics under heterogeneous control signals. It supports both egocentric world exploration and embodied task execution.}
	\label{fig:main}
\end{figure}

As shown in Fig.~\ref{fig:main}, Worldscape-MoE mainly consists of heterogeneous data representation, expert routing, a stack of DiT layers, and MoE blocks.

\noindent\textbf{Heterogeneous Data Representation.}
During training, we jointly optimize the model on mixed-control data and design dedicated injection pathways according to the structural properties of different control signals. Video frames and reference images are first encoded by a frozen VAE into latent representations, which serve as the basic visual inputs to the diffusion model. Text instructions are encoded by a frozen text encoder to extract semantic features and are injected into the backbone through the original cross-attention pathway. \textbf{Locomotion} mainly characterizes camera trajectories or local viewpoint changes. Its control signals are transformed into dense camera-control features and mapped by a lightweight trainable Control Adapter into a feature space aligned with the video latents, after which they are fused with patch-level video tokens. \textbf{Hand Motion} provides dense spatiotemporal constraints in the form of action maps. These signals are therefore first encoded by the frozen VAE into latent control features and then fused into the conditional latent branch as additional visual conditions for the diffusion backbone. \textbf{Manipulation}, in contrast, is represented as a low-dimensional action sequence that specifies the intended operation. We use a trainable Action MLP to encode it into compact action embeddings, which are further integrated into the timestep embedding and its projected modulation vectors, allowing the manipulation signal to affect the generated dynamics through the temporal conditioning pathway of DiT blocks. This asymmetric injection design is deliberate: spatially dense controls are injected through visual-token pathways, while compact action controls are injected through temporal modulation. These heterogeneous control signals can be unified as modality-dependent conditioning functions within a shared diffusion backbone:
\begin{equation}
	\hat{\epsilon} = f_{\theta}\bigl(z_v \oplus \phi_{\mathrm{vis}}(A),\; z_c,\; e_t + \phi_{\mathrm{tmp}}(A)\bigr),
\end{equation}
where $z_v$ denotes the VAE latent of video frames and reference images, $z_c$ denotes the text embedding, $e_t$ denotes the timestep embedding, $\phi_{\mathrm{vis}}(A)$ denotes the control-dependent visual injection function, and $\phi_{\mathrm{tmp}}(A)$ denotes the control-dependent temporal modulation function. For \textbf{Locomotion}, $\phi_{\mathrm{vis}}(A)$ is implemented by the Control Adapter and fused at the patch-token level; for \textbf{Hand Motion}, it corresponds to VAE-encoded action-map latents injected into the conditional visual branch; and for \textbf{Manipulation}, $\phi_{\mathrm{tmp}}(A)$ is implemented by the Action MLP and injected through the timestep modulation pathway. Through this unified formulation, heterogeneous controls are incorporated into different pathways of the same DiT backbone, enabling multi-control modeling while preserving the priors of the pretrained video generator.

\noindent\textbf{Expert Routing and MoE Forward.}
We replace the feed-forward network in each DiT block with a control-aware MoE-FFN, enabling shared generative computation and per-modality specialization under heterogeneous supervision. For a model supporting $M$ control modalities, the expert set at block $l$ is $\mathcal{E}^{(l)}=\{\mathcal{E}^{(l)}_0,\mathcal{E}^{(l)}_1,\ldots,\mathcal{E}^{(l)}_M\}$, where $\mathcal{E}^{(l)}_0$ is shared across all samples and $\mathcal{E}^{(l)}_r$ is specialized for the $r$-th control modality. All experts are initialized from the pretrained FFN or the current shared expert (Sec.~\ref{Sec:uni_train}) rather than from random weights, preserving the video-generation prior.

For sample $i$, let $H^{(l)}_i=[h^{(l)}_{i,1},\ldots,h^{(l)}_{i,N}]\in\mathbb{R}^{N\times d}$ denote the hidden tokens at block $l$. We encode the available controls by a binary vector $\mathbf{u}_i=[u_{i,1},\ldots,u_{i,M}]\in\{0,1\}^{M}$ and construct an eligibility mask
\begin{equation}
	\mathbf{m}_i=[1,u_{i,1},\ldots,u_{i,M}]\in\{0,1\}^{M+1},
	\qquad
	\mathcal{A}_i=\{k\in\{0,\ldots,M\}\mid m_{i,k}=1\},
\end{equation}
where the shared expert is always eligible and each modality-specific expert is activated only when its corresponding control signal is present.

The router then predicts sample-wise logits from mean-pooled hidden states and computes routing weights by a masked softmax over the eligible experts:
\begin{equation}
	\mathbf{z}^{(l)}_i=\frac{1}{N}\sum_{n=1}^{N}h^{(l)}_{i,n},
	\qquad
	\mathbf{a}^{(l)}_i=\mathbf{W}^{(l)}_{\mathrm{route}}\,\mathbf{z}^{(l)}_i,
\end{equation}
\begin{equation}
	\alpha^{(l)}_{i,k}
	=
	\frac{m_{i,k}\exp(a^{(l)}_{i,k}/\tau)}
	{\sum_{j=0}^{M}m_{i,j}\exp(a^{(l)}_{i,j}/\tau)},
	\qquad
	k\in\{0,\ldots,M\},
\end{equation}
where $\mathbf{W}^{(l)}_{\mathrm{route}}\in\mathbb{R}^{(M+1)\times d}$ is a learnable routing matrix and $\tau$ is a temperature hyperparameter. By construction, ineligible experts receive zero probability and do not participate in the forward computation or the sample-wise gradient update.

The MoE-FFN output is then aggregated over the active expert set $\mathcal{A}_i$:
\begin{equation}
	\mathrm{MoE}^{(l)}\!\left(H^{(l)}_i,\mathbf{u}_i\right)
	=
	\sum_{k\in\mathcal{A}_i}\alpha^{(l)}_{i,k}\,
	\mathcal{E}^{(l)}_k\!\left(H^{(l)}_i\right).
\end{equation}
The router is implemented as a bias-free linear layer with zero initialization, yielding uniform weights over $\mathcal{A}_i$ at the start of training. We do not introduce an auxiliary load-balancing loss; instead, expert specialization is induced by deterministic modality eligibility and optimized end-to-end with the diffusion objective. This routing rule guarantees that every sample updates the shared expert, which is essential for accumulating cross-control world knowledge, and prevents unrelated modality experts from receiving noisy gradients from controls they are not designed to interpret.

\subsection{Worldscape-MoE Tuning}
\label{Sec:uni_train}
\begin{figure}[htbp]
	\centering
     \includegraphics[width=0.90\linewidth]{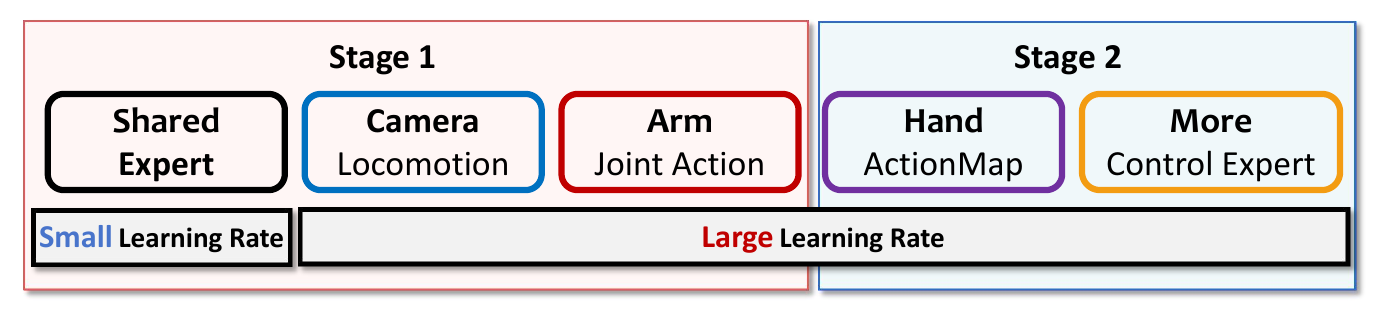}
	\caption{\textbf{Worldscape-MoE Tuning.} The shared expert is optimized with a smaller learning rate, while each modality-specific expert uses a larger learning rate for control specialization.}
    \label{fig:tuning}
    \vspace{-3pt}
\end{figure}

\noindent\textbf{Grouped Learning.}
To rapidly adapt the model to new control capabilities while preserving the generation priors inherited from pretraining, we adopt a grouped learning-rate strategy, as illustrated in Fig.~\ref{fig:tuning}. Newly introduced or strongly modality-specific modules are optimized with the base learning rate, whereas shared components inherited from the pretrained model, such as the Shared Expert and part of the adapter parameters, are fine-tuned with a smaller learning rate. This prevents the shared expert from overfitting to the most recent or most frequent control type while allowing newly added experts and adapters to acquire control-specific behavior quickly.

\noindent\textbf{Progressive Control Expansion.}
To stably extend the model to multiple control capabilities, we adopt a progressive modality-incremental strategy rather than introducing all experts and control branches at once. The process consists of two stages. In \textbf{Stage I}, starting from a pretrained video generation model, we train the model with camera-control and robotic-arm signals to establish reliable generation priors, stable camera-trajectory control, and embodied task capability. In \textbf{Stage II}, we progressively introduce new control modalities together with their corresponding experts. Each newly added expert is initialized from the current Shared Expert rather than from random weights, so the new branch inherits the accumulated world prior before specializing to its own control interface. This stage-wise design makes Worldscape-MoE a continually extensible framework that can absorb heterogeneous supervisory signals for world simulation.

\section{Experiments}
\label{Sec:Experiments}

\subsection{Experimental Setup}
 To ensure a fair and reproducible comparison, we standardize the evaluation protocol across models whenever their official implementations permit the corresponding control and resolution settings. The evaluation covers three representative control regimes: camera-trajectory locomotion, robotic manipulation, and egocentric hand motion.

\noindent\textbf{Locomotion Setting.}
Following the dataset and metric design of iWorld-Bench~\citep{fang2026iworldbench}, we construct a 500-case locomotion evaluation set and adopt its trajectory-following, generation-quality, and consistency metrics. This setting measures whether the model can follow camera-level actions while maintaining visual quality and temporal smoothness.

\noindent\textbf{Manipulation Setting.}
For robotic-arm manipulation, we follow the WorldArena evaluation protocol~\citep{shang2026worldarena}. We directly adopt the released WorldArena results and use the corresponding inference configurations for manipulation baselines. This setting evaluates whether generated videos preserve functional action outcomes, object interactions, and embodied task consistency.
  
\noindent\textbf{Hand Motion Setting.}
For hand-motion control, we follow the egocentric hand-motion evaluation setting and randomly select 100 hand-object interaction samples from the EgoDex test set~\citep{hoque2025egodex}. We evaluate generated videos with FID-VID, FVD, image-level FID, and Image Quality to measure frame-level appearance alignment, temporal coherence, and visual fidelity under dense hand-joint constraints.

\subsection{Overall Evaluation}
\label{sec:exp}

\begin{table*}[b]
\centering

\begin{minipage}{\textwidth}
\centering
\footnotesize
\setlength{\tabcolsep}{3pt}
\renewcommand{\arraystretch}{1.0}
\caption{\textbf{Locomotion Experiments on iWorldBench.} Comparison of metric scores for different models on generation quality and trajectory following tasks.}
\label{tab:example_table_large_font}
\begin{tabular}{r|c|cccc|cc}
\hline
\multirow{3}{*}{} & 
\multicolumn{1}{c|}{}  & 
\multicolumn{4}{c|}{Generation Quality} & 
\multicolumn{2}{c}{Trajectory Following} \\ \cline{2-8} 
& 
& Image & Brightness & Color Temp. & Sharpness 
& Motion & Trajectory \\  
Method 
& Avg.$\uparrow$
& Quality $\uparrow$ & Consistency $\uparrow$ & Constraint $\uparrow$ & Retention $\uparrow$ 
& Smoothness $\uparrow$ & Accuracy $\uparrow$ \\ \hline
\rowcolor[HTML]{E6F2FF} \textbf{Worldscape-MoE} 
& \textbf{0.7556}
& 0.6955 &0.7758 &\textbf{0.7639}  &0.6745
&\textbf{0.9941}   & 0.6300  \\
\rowcolor[HTML]{E6F2FF} \textbf{w/o MoE} 
& 0.6869
& 0.6710 &0.6993 &0.6613  &0.4865
&0.9930   & 0.6100  \\
Matrix-game 3.0 
& 0.6232
& 0.5633  &0.6180  &0.6353  & 0.3852 
& 0.9660 & 0.5714 \\
HY-World 1.5 
& 0.7322 
& \textbf{0.7128} & 0.7027 & 0.7477 & 0.5545 
& 0.9908 & 0.6844 \\ 
CameraCtrl 
& 0.5521 
& 0.4602 & 0.4812 & 0.3076 & 0.4833 
& 0.9832 & 0.5970 \\
MotionCtrl 
& 0.5562 
& 0.4583 & 0.5296 & 0.2421 & 0.5182 
& 0.9776 & 0.6115 \\
CamI2V 
& 0.6137 
& 0.5150 & 0.5904 & 0.4513 & 0.5255 
& 0.9886 & 0.6115 \\
RealCam-I2V 
& 0.7063 
& 0.6530 & 0.5712 & 0.6197 & 0.6987 
& 0.9901 & 0.7050 \\
videox-fun-Wan 
& 0.7443 
& 0.6684 & 0.6856 & 0.6640 & 0.6934 
& 0.9899 & \textbf{0.7645} \\
AC3D 
& 0.7262 
& 0.4884 & \textbf{0.7764} & 0.7050 & \textbf{0.7213 }
& 0.9934 & 0.6729 \\
ASTRA 
& 0.6072 
& 0.5600 & 0.5916 & 0.5088 & 0.5625 
& 0.9826 & 0.4379 \\ \hline
\end{tabular}
\end{minipage}

\vspace{3mm}

\begin{minipage}[t]{0.42\textwidth}
\centering
\small
\setlength{\tabcolsep}{5pt}
\renewcommand{\arraystretch}{0.75}
\caption{\textbf{Manipulation Experiments on the WorldArena Benchmark.} EWM Score for dual-arm manipulation across 16 metrics.}
\label{tab:benchmark_ewm}
\begin{tabular}{l|c}
\toprule
\textbf{Models} & \textbf{EWM Score} \\
\midrule
\rowcolor[HTML]{E6F2FF} \textbf{Worldscape-MoE}\strut & \textbf{62.84} \\
\rowcolor[HTML]{E6F2FF} \textbf{w/o MoE}\strut & 61.88 \\
CtrlWorld & 59.98 \\
Wan 2.6 & 59.80 \\
CogvideoX & 58.79 \\
Veo 3.1 & 57.77 \\
IRASim & 56.14 \\
TesserAct & 54.62 \\
Cosmos-Predict 2.5 (action) & 54.29 \\
Cosmos-Predict 2.5 (text) & 53.06 \\
Vidar & 51.92 \\
Wan 2.2 & 51.71 \\
GigaWorld-0 & 50.96 \\
RoboMaster & 50.35 \\
\bottomrule
\end{tabular}
\end{minipage}
\hfill
\begin{minipage}[t]{0.55\textwidth}
\centering
\small
\setlength{\tabcolsep}{5pt}
\renewcommand{\arraystretch}{1.6}
\caption{\textbf{Hand-motion Experiments.} Comparison of hand-motion control measured by FID-VID, FVD, image-level FID, and Image Quality.}
\label{tab:hand_motion_metrics}
\begin{tabular}{l|cccc}
\toprule
\textbf{Model} & \makecell{\textbf{FID-VID} $\downarrow$} & \makecell{\textbf{FVD} $\downarrow$} & \makecell{\textbf{FID} $\downarrow$}  & \makecell{\textbf{Image}\\\textbf{Quality} $\uparrow$}\\
\midrule
\rowcolor[HTML]{E6F2FF} \textbf{Worldscape-MoE} & \textbf{3.80} & \textbf{110.94} & \textbf{5.78} & \textbf{0.7325} \\
\rowcolor[HTML]{E6F2FF} \textbf{w/o MoE} & 5.39 & 128.87 & 15.34 & 0.7250\\ 
HunyuanVideo-1.5 & 23.18 & 517.42 & 56.31 & 0.6419 \\
Cosmos-Predict 2.5 & 15.02 & 628.96 & 51.36 & 0.6158 \\
MimicMotion & 26.74 & 589.47 & 48.92 & 0.5324 \\
MagicPose & 65.93 & 1498.65 & 91.78 & 0.5739 \\
LOME & 144.58 & 1794.84 & 67.82 & 0.5281 \\
\bottomrule
\end{tabular}
\end{minipage}
\end{table*}

\noindent\textbf{Results and analysis.}
Table~\ref{tab:example_table_large_font} shows that Worldscape-MoE attains the best overall locomotion score of $0.7556$, improving over the strongest baseline VideoX-Fun-Wan by $0.0113$ while leading on motion smoothness with $0.9941$. This indicates that heterogeneous training does not weaken camera-level action following.

On robotic-arm manipulation, Table~\ref{tab:benchmark_ewm} reports the highest EWM score of $62.84$, surpassing CtrlWorld and Wan 2.6 by $+2.86$ and $+3.04$, respectively, with detailed per-metric breakdowns provided in Appendix Tables~\ref{tab:benchmark_part1} and~\ref{tab:benchmark_part2}.

For egocentric hand-motion control, Table~\ref{tab:hand_motion_metrics} shows that Worldscape-MoE achieves the best performance across all evaluated metrics, obtaining $3.80$ FID-VID, $110.94$ FVD, $5.78$ image-level FID, and $0.7325$ Image Quality. Compared with the w/o MoE variant, it reduces FID-VID, FVD, and image-level FID by $1.59$, $17.93$, and $9.56$, respectively.

\begin{figure}[t]
    \centering
    \includegraphics[width=0.85\linewidth]{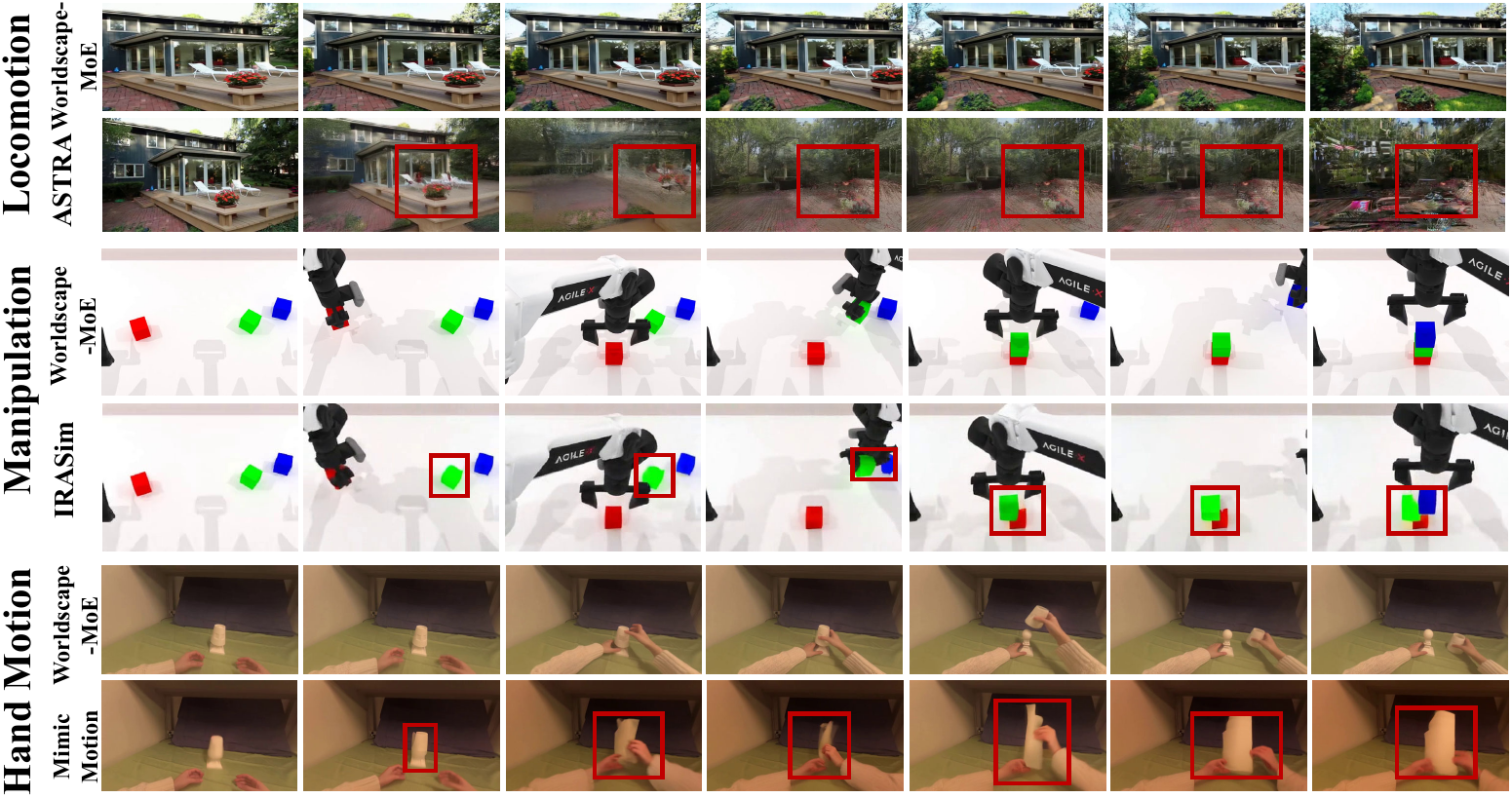}
    \caption{\textbf{Qualitative comparison under heterogeneous controls.} Across locomotion, manipulation, and hand-motion settings, Worldscape-MoE produces more controllable and physically coherent world dynamics than representative baselines.}
    \label{fig:qualitative-heterogeneous-comparison}
\end{figure}

Across the three heterogeneous control regimes, Worldscape-MoE ranks first on every primary metric, demonstrating a level of cross-modality consistency rarely achieved by prior controllable world models. The gains are not limited to the largest data modality, which is important for interpreting the result as scaling behavior rather than simple data imbalance. Locomotion benefits from broader visual dynamics, manipulation benefits from a stronger shared physical prior, and hand control benefits from improved temporal coherence. These observations support the central claim that different control signals can reinforce one another when the architecture separates shared world knowledge from control-specific specialization. The same pattern is visible in the qualitative comparison in Fig.~\ref{fig:qualitative-heterogeneous-comparison}.

\subsection{Quantitative Analysis of MoE}
\label{sec:quantitative-routing}

To verify whether Worldscape-MoE learns control specialization rather than merely scaling capacity, we record the gate-weighted expert workload at each MoE-FFN layer on the action-only evaluation cases. This analysis reveals how computation is allocated across the shared and modality-specific experts during inference. Concretely, we attach forward hooks to all MoE-FFN routers, accumulate each expert's routing weight over all tokens and denoising steps, and normalize the resulting layer-wise weighted token counts to draw the all-expert, dedicated-expert, and shared-vs-dedicated loading panels.

\begin{figure}[htbp]
    \centering
    \includegraphics[width=\linewidth,height=0.72\textheight,keepaspectratio]{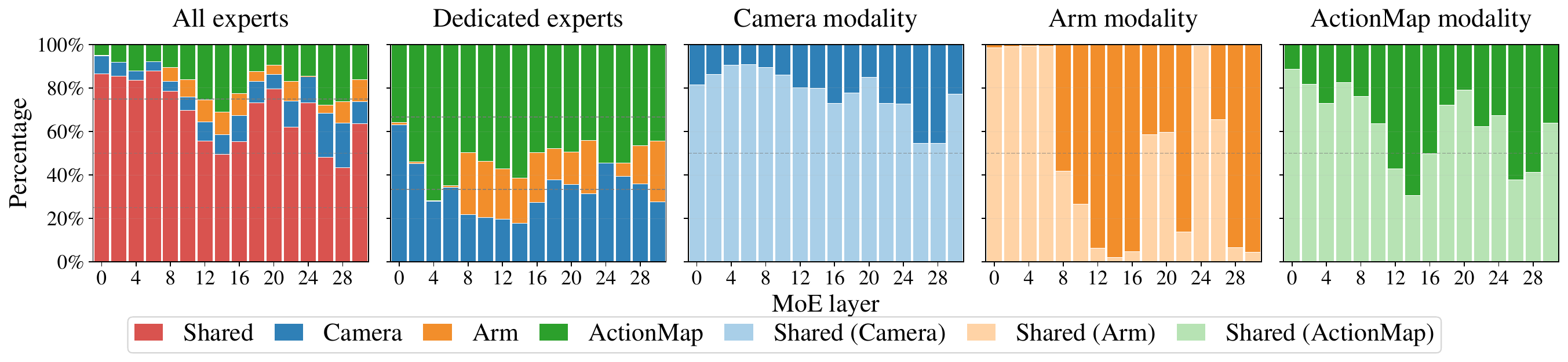}
    \caption{\textbf{Layer-wise MoE expert loading analysis.}
    Gate-weighted expert workload across MoE layers. The panels show all-expert loading, dedicated-expert loading, and shared-vs-dedicated routing for locomotion, manipulation, and hand-motion control.}
    \label{fig:moe-routing-analysis}
\end{figure}

As shown in Fig.~\ref{fig:moe-routing-analysis}, the shared expert remains the dominant path in the overall workload, accounting for 69.48\% of gate-weighted computation across the sampled action-only cases. The dedicated experts exhibit clear modality-dependent usage: they account for 20.91\% of the workload for locomotion, 47.95\% for manipulation, and 35.97\% for hand motion. This pattern is technically meaningful. Camera-like locomotion can be largely handled by the shared spatiotemporal prior with moderate expert-specific correction, while manipulation and action-map hand control require stronger modality-specific computation because they impose more direct constraints on object contact and fine-grained hand-object interaction. The routing statistics therefore support the intended decomposition: shared experts model common world dynamics, while dedicated experts specialize in the interface-specific action semantics.

\subsection{MoE Ablation and Modality Extension}
\begin{figure}[htbp]
    \centering
    \vspace{-10pt}
    \includegraphics[width=0.90\linewidth]{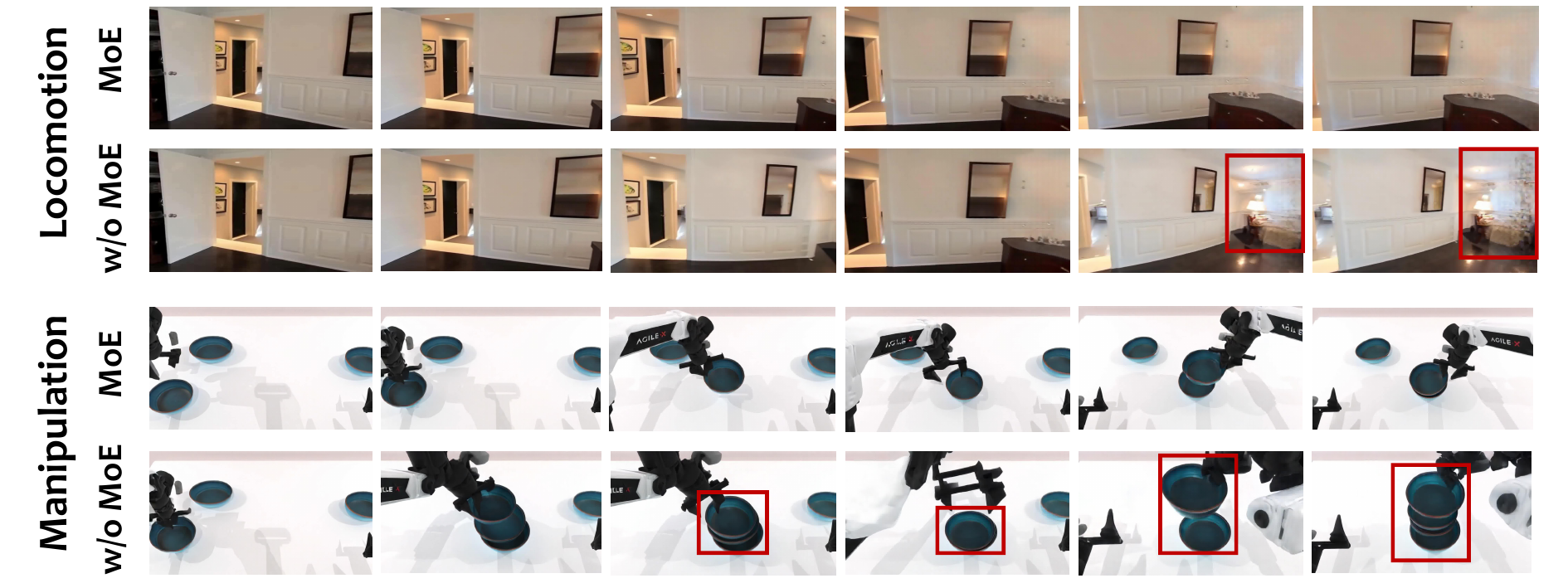}
    \caption{\textbf{Qualitative evidence for the effectiveness of MoE.} Worldscape-MoE produces more stable results than direct mixed training without MoE under heterogeneous controls.}
    \label{fig:moe_effectiveness}
    \vspace{-5pt}
\end{figure}

\noindent\textbf{Effectiveness of MoE.} 
To isolate the contribution of the MoE design, we compare Worldscape-MoE with a direct mixed-training baseline without MoE that jointly optimizes all control modalities using the same dense backbone. As shown in Fig.~\ref{fig:moe_effectiveness} and Fig.~\ref{fig:WOmoe}, the dense baseline suffers from cross-modal control interference, resulting in weaker control fidelity and less coherent world dynamics. Although this w/o MoE baseline remains competitive overall, Worldscape-MoE achieves stronger scores across locomotion, manipulation, and hand motion under the same training budget. The reason is that dense mixed training forces all controls to update the same feed-forward pathway, while Worldscape-MoE allows the shared expert to learn common dynamics and the dedicated experts to absorb control-specific residuals. This architectural separation is what turns heterogeneous supervision into a scaling resource rather than a source of gradient conflict.

\noindent\textbf{Scalability of MoE.}
We further evaluate whether the proposed framework can be extended to additional control modalities without compromising previously acquired capabilities. After introducing a new modality, Worldscape-MoE successfully learns the new control signal while largely preserving generation quality and control fidelity on existing modalities. We observe a brief degradation in locomotion performance at the early stage of extension, consistent with a temporary adaptation phase as shared and modality-specific parameters are re-balanced to absorb the new control knowledge. This transient effect is gradually mitigated as training proceeds, and the original locomotion capability recovers. We also conduct an additional extension experiment on single-arm LIBERO manipulation, with the setting and qualitative results provided in Appendix~\ref{app:libero_single_arm}. Additional evidence from parameter-update dynamics is provided in Appendix~\ref{app:weight}. Overall, these results suggest that Worldscape-MoE provides a practical scaling path: new controls can be added by extending the expert set and control branch while retaining the shared world prior.

\subsection{Out-of-Distribution Generalization and Loco-Manipulation Control}

\begin{figure}[htbp]
	\centering
	\includegraphics[width=0.90\linewidth]{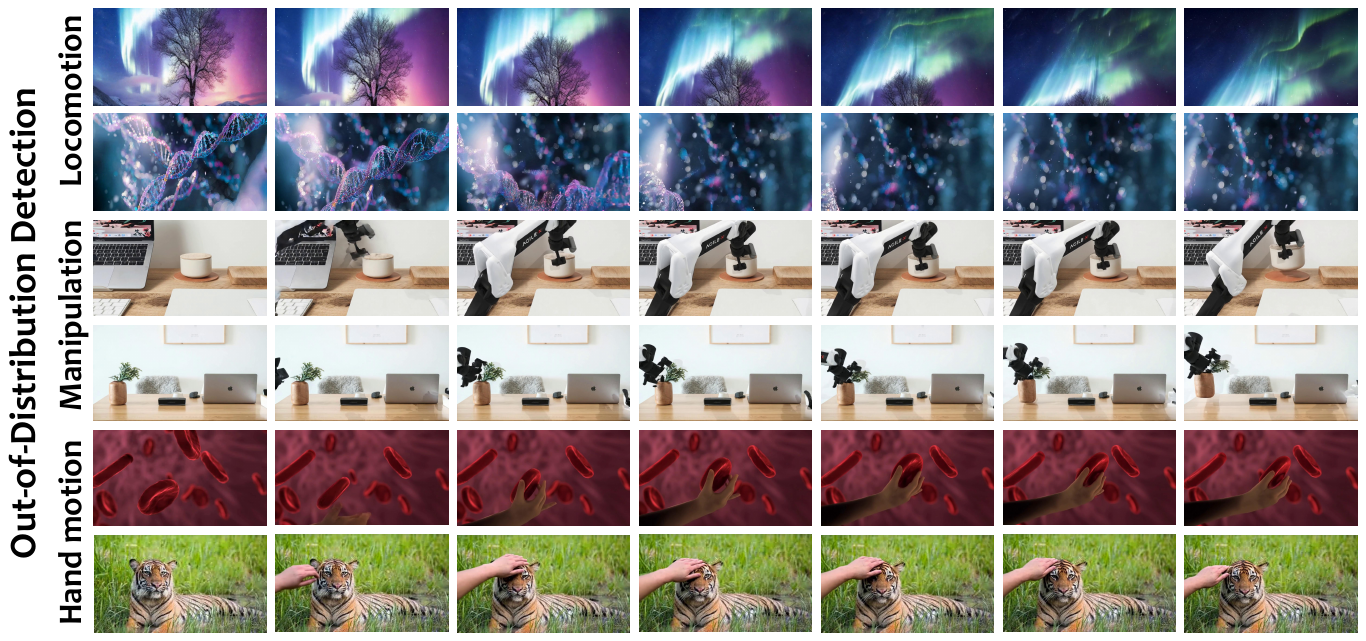}
    \vspace{-5pt}
    \caption{\textbf{Out-of-distribution detection across three motion domains.}
Representative OOD across locomotion, robotic manipulation, and hand-motion tasks.}
    \label{fig:ood}
\end{figure}

As shown in Fig.~\ref{fig:ood}, Worldscape-MoE demonstrates strong out-of-distribution generalization across locomotion, robotic manipulation, and hand-motion domains. It generates coherent locomotion rollouts under large visual shifts, generalizes simulated Robotwin arms to realistic furniture-service scenarios, and transfers hand-pose control trained from a single dataset to diverse unseen scenes. These results suggest that the model has not merely memorized modality-specific data distributions. Instead, heterogeneous training encourages the shared expert to represent reusable scene dynamics and physical regularities, which can be transferred to unseen domains.

Beyond single-modality control, Worldscape-MoE benefits from generating from broad and diverse world knowledge rather than being constrained to one specific control form. This leads to stronger modeling of physical regularities, as illustrated in Fig.~\ref{fig:physics}, and further enables coupled reasoning across different actions. In particular, we observe that the model can synthesize loco-manipulation behaviors in diverse environments, as shown in Fig.~\ref{fig:loco-mani}. This is an important qualitative indicator for scalability: once the model learns multiple controls in a shared world representation, it can begin to compose capabilities that were previously isolated. More qualitative out-of-distribution cases are provided in Appendix~\ref{sec:More Qualitative Results}.

\vspace{-5pt}
\section{Related Work}
\label{related_work}

\noindent\textbf{Video generation foundations for world modeling.}
World modeling has recently benefited from the rapid improvement of large-scale video generation backbones, including Wan~\cite{wan2025wan}, HunyuanVideo~\cite{wu2025hunyuanvideo}, CogVideoX~\cite{yang2024cogvideox}, and Cosmos~\cite{alhaija2025cosmos}. These models provide strong visual priors, temporal generation ability, and scalable diffusion-transformer architectures. However, general video generators do not by themselves solve world modeling: an embodied or interactive agent requires controllable prediction under actions, not only visually plausible continuation. The key research question is therefore how action supervision should be represented, injected, and scaled.

\noindent\textbf{Interactive and locomotion-oriented world models.}
One line of work studies interactive environments controlled by camera motion, navigation commands, or game-like actions. Genie~\cite{bruce2024genie}, Matrix-Game~\cite{he2025matrix,zhang2025matrix,wang2026matrix}, HY-World~\cite{sun2025worldplay,hy2026hy}, CameraCtrl~\cite{he2025cameractrl}, MotionCtrl~\cite{wang2024motionctrl}, CamI2V~\cite{zheng2024cami2v}, RealCam-I2V~\cite{li2025realcam}, VideoX-Fun-Wan~\cite{videoxfun2024}, and AC3D~\cite{bahmani2025ac3d} demonstrate that trajectory-like controls can guide video generation toward interactive scene exploration. These methods are important for world navigation and scene simulation, but the learned control interface is usually tied to camera motion. As a result, their data and learned dynamics are difficult to reuse directly for robot-action or hand-joint control.

\noindent\textbf{Embodied manipulation and hand-control world models.}
Another line of work focuses on embodied action control. GigaWorld-0~\cite{team2025gigaworld}, Cosmos-Predict~\cite{ali2025world}, and WorldArena~\cite{shang2026worldarena} emphasize robot-action-conditioned future prediction and functional utility for manipulation. Hand2World~\cite{wang2026hand2world}, Generated Reality~\cite{xie2026generated}, MimicMotion~\cite{zhang2024mimicmotion}, MagicPose~\cite{chang2023magicpose}, and LOME~\cite{gao2026lome} explore hand or human-motion-conditioned generation. These models expand the scope of world modeling beyond camera motion, but they still generally treat each action representation as a separate modeling problem. The consequence is a set of specialized control islands: each model may be effective within its own interface, yet the field lacks a mechanism for pooling heterogeneous supervision into a shared, extensible world model.

\noindent\textbf{Mixture-of-Experts for heterogeneous modeling.}
Mixture-of-Experts (MoE) architectures were originally introduced as adaptive computation mechanisms~\cite{jacobs1991adaptive} and have become a central scaling strategy for modern Transformer models~\cite{shazeer2017olln,lepikhin2020Ggshard,fedus2022st,dai2024deepseekmoe}. They have also been applied to multimodal systems, such as MoE-LLaVA~\cite{lin2024moellava} and DeepSeek-VL2~\cite{wu2024deepseekvl2}, where expert specialization helps manage heterogeneous visual-language inputs. In generative modeling, Nucleus-Image~\cite{akiti2026nucleus} shows the potential of sparse experts for image generation, while Astra~\cite{zhu2026astra} and Wan-Animate~\cite{cheng2025wananimate} introduce expert-style specialization in action or denoising contexts. These works motivate MoE as a useful tool, but existing applications do not directly address the central problem of heterogeneous action-control world modeling. Worldscape-MoE uses MoE to explicitly factor a world model into shared world dynamics and control-specific action interpretation, pushing the field from single-control islands toward a unified and scalable multi-control framework.

\section{Conclusion and Future Work}
In this work, we presented Worldscape-MoE, the first world model framework that unifies heterogeneous action controls within a single model. By jointly learning from heterogeneous control data, Worldscape-MoE improves physical consistency, temporal coherence, and overall world-simulation quality over single-control training, offering a promising direction for alleviating the data bottleneck of current world models. Through the combination of a shared expert for cross-control world knowledge and modality-specific experts for control specialization, our framework achieves strong generalization and competitive performance across diverse benchmarks and downstream tasks. In future work, we plan to further improve its efficiency and scalability by compressing the DiT architecture and extending the framework to broader control modalities.
\bibliography{references}

\newpage
\appendix
\section{Broader Impacts}
\label{app:broder}
This work may have several positive broader impacts. By unifying heterogeneous control modalities within a single world model, Worldscape-MoE provides a scalable framework for learning from broader and more diverse data sources, which may facilitate the development of more general and data-efficient intelligent agents. The resulting improvements in simulation fidelity and controllability could benefit robotics, embodied AI, and interactive decision-making, while reducing reliance on costly real-world data collection. In addition, the shared-expert design may promote transferable physical knowledge across tasks and control settings, supporting more robust downstream learning.

At the same time, the high fidelity of generated videos also requires careful ethical consideration. In particular, such models could be misused to generate deceptive synthetic videos, including deepfakes, or to misrepresent safe procedures and robot behaviors. To mitigate these risks, future work should explore safeguards such as synthetic-content detection, provenance or watermarking mechanisms, and anomaly detection methods for identifying unsafe or non-standard behaviors. We also plan to incorporate stronger safety-oriented generation constraints and detection mechanisms into future public releases of our models to support more responsible deployment in safety-critical applications.

\section{Limitation}
\label{app:limit}
Despite the strong performance of our framework, Worldscape-MoE remains computationally expensive to train and requires substantial GPU resources. In addition, the training speed decreases as more control modalities are incorporated, which limits the practical scalability of the current framework. Improving training efficiency is therefore an important direction for future work. One possible solution is to pre-encode VAE features in advance to reduce the computational overhead during training. Another limitation is that the current framework is not yet optimized for real-time deployment. In particular, how to distill Worldscape-MoE into a faster model and adapt it to existing acceleration or distillation pipelines remains an open problem. Addressing these challenges is essential for building a real-time, high-performance world model that supports heterogeneous control at scale.

\section{Data Process Details}
\label{app:data}

\subsection{Open Sources Datasets Details} 
\noindent\textbf{RealEstate10K.}
RealEstate10K~\cite{zhou2018stereo} is a large-scale dataset providing camera trajectories from video clips, with camera poses organized as continuous trajectories describing spatial position and orientation over time. These poses are estimated via large-scale reconstruction pipelines based on SLAM and bundle adjustment.

\noindent\textbf{DL3DV-10K.}
DL3DV-10K~\cite{ling2024dl3dv} is a real-world dataset with high-quality videos and manual annotations of key scene structures and complexity. In addition to camera poses, it includes rich geometric information such as NeRF-based depth, point clouds, and 3D meshes. The dataset supports tasks including 3D reconstruction, scene consistency modeling, and vision-language learning.

\noindent\textbf{Sekai.}
Sekai~\cite{li2025sekai} is a large-scale, diverse video dataset that focuses on rich scene dynamics and varied visual content, making it suitable for tasks involving motion understanding and generation. The dataset contains a wide range of environments and object interactions, providing complex camera movements and temporal variations. Each video clip is accompanied by structured annotations, including captions and motion-related information, which facilitate downstream tasks such as trajectory modeling and video synthesis.

\noindent\textbf{SpatialVID.}
SpatialVID~\cite{wang2025spatialvid} is a large-scale video dataset built for multi-modal generation tasks such as text-to-video and image-to-video. It contains video data along with corresponding annotations organized into multiple subsets. Each clip is paired with structured metadata that can be easily processed, and the annotations include captions, dynamic masks, camera parameters, poses, and motion instructions, supporting detailed spatial-temporal modeling.

\noindent\textbf{EgoDex.}
EgoDex~\cite{hoque2025egodex} is an egocentric video dataset designed for learning dexterous human manipulation from first-person observations. It covers a wide range of tabletop manipulation tasks such as grasping, placing, folding, and arranging objects. Each video is paired with frame-level 3D pose annotations for the head, upper body, and both hands, along with camera intrinsics, camera transforms, joint SE(3) transforms, confidence values, and natural-language annotations; from EgoDex, we obtain hand-joint control training clips.

\noindent\textbf{Ego4D.}
Ego4D~\cite{grauman2022ego4d} is a large-scale open-domain egocentric video dataset capturing unscripted daily-life activities across diverse environments. It spans household activities, workplaces, outdoor scenes, leisure activities, social interactions, and commuting. Although Ego4D provides broad scene diversity and supports multiple first-person perception benchmarks, it does not include native frame-aligned 3D hand-joint poses for all videos, so we generate hand-joint control supervision using the same automatic hand reconstruction pipeline and obtain training clips.

\noindent\textbf{LIBERO.}
LIBERO~\cite{liu2023libero} is a language-conditioned benchmark and dataset for lifelong single-arm robot manipulation. It provides an extendible procedural generation pipeline together with high-quality human-teleoperated demonstrations for multiple tasks organized into several task suites. These suites are designed to evaluate knowledge transfer under controlled changes in spatial relations, object categories, task goals, and their entangled combinations. For our single-arm setting, we process the released LIBERO demonstrations into robot-action-conditioned video clips with aligned task language and action trajectories, and use the corresponding train/validation metadata to study transfer to held-out manipulation cases.

\subsection{Synthetic Data}
\label{app:add_data}

In this study, we adopt the 81 camera configurations and data collection strategies from iWorld-Bench~\cite{fang2026iworldbench}, and perform large-scale stylized data acquisition in Unreal Engine. Specifically, we leverage diverse scenes spanning both indoor and outdoor environments, covering a wide range of visual styles such as urban, natural, and fantasy settings. Across these scenes, we sample multiple camera viewpoints, and further generate training clips through systematic trajectory execution and multi-view capture. We present a representative subset of the selected scenes, along with their scene categories, illustrating the diversity and coverage of our data collection process.

\begin{table*}[htbp]
	\centering
	\caption{Overview of simulation environments with scene types.}
	\label{tab:scene_overview}
	\renewcommand{\arraystretch}{1.1}
	\setlength{\tabcolsep}{3pt}
	
	\newcommand{\sceneitem}[3]{
		\begin{minipage}[t]{0.18\textwidth}
			\centering
			\includegraphics[width=\linewidth,height=0.105\textheight,keepaspectratio]{#1}
			
			\vspace{1.5mm}
			{\footnotesize \textbf{#2}}\\
			{\scriptsize #3}
		\end{minipage}
	}
	
	\begin{tabular}{ccccc}
		
		\sceneitem{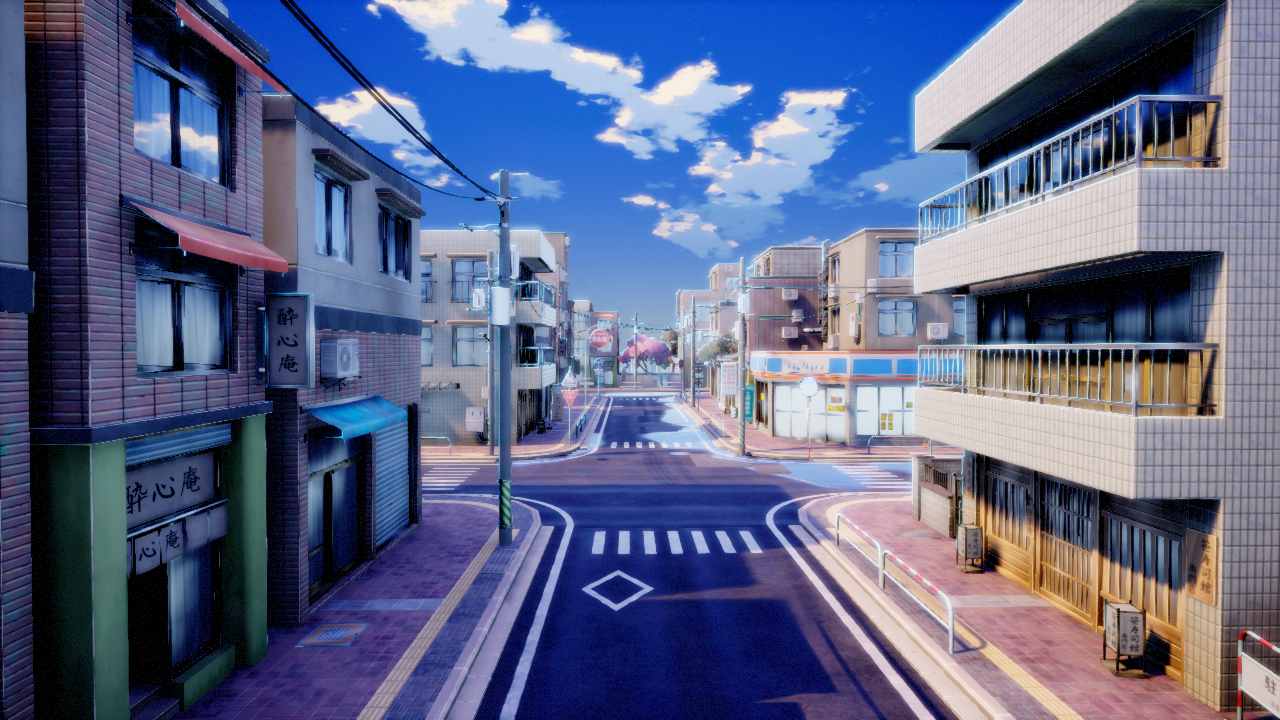}{Cartoon Street}{Outdoor} &
		\sceneitem{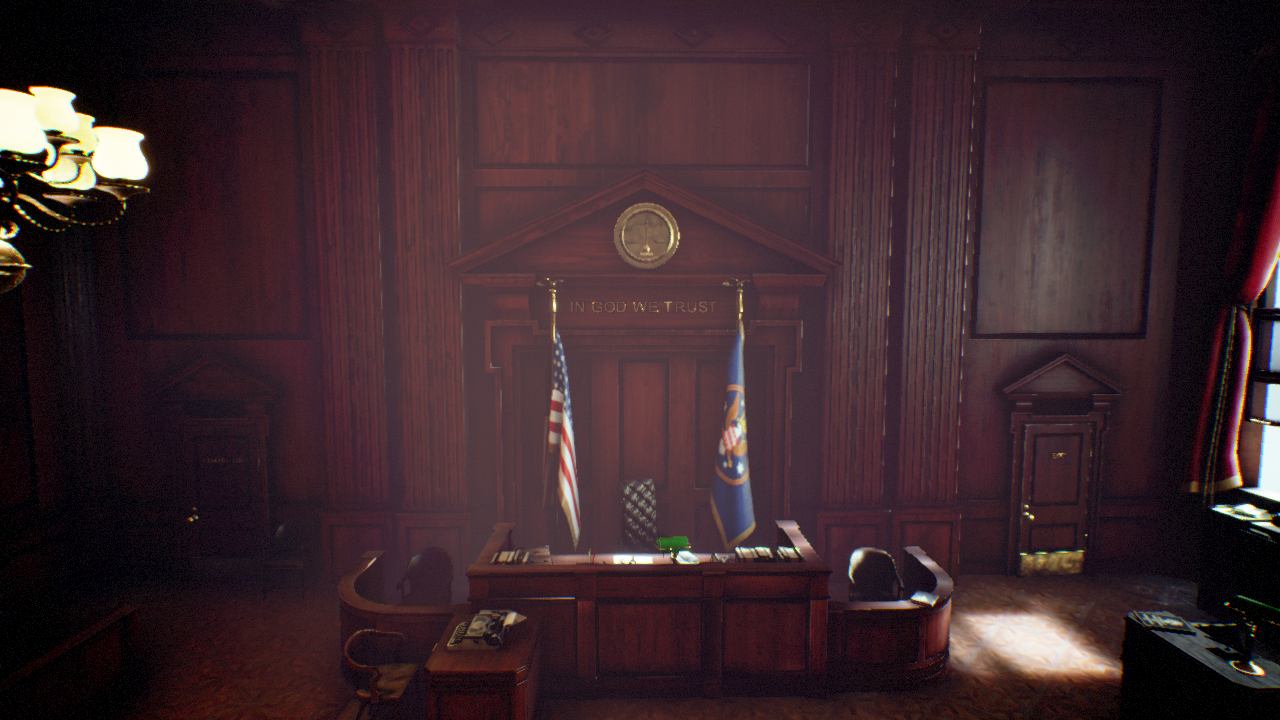}{Courtroom}{Indoor} &
		\sceneitem{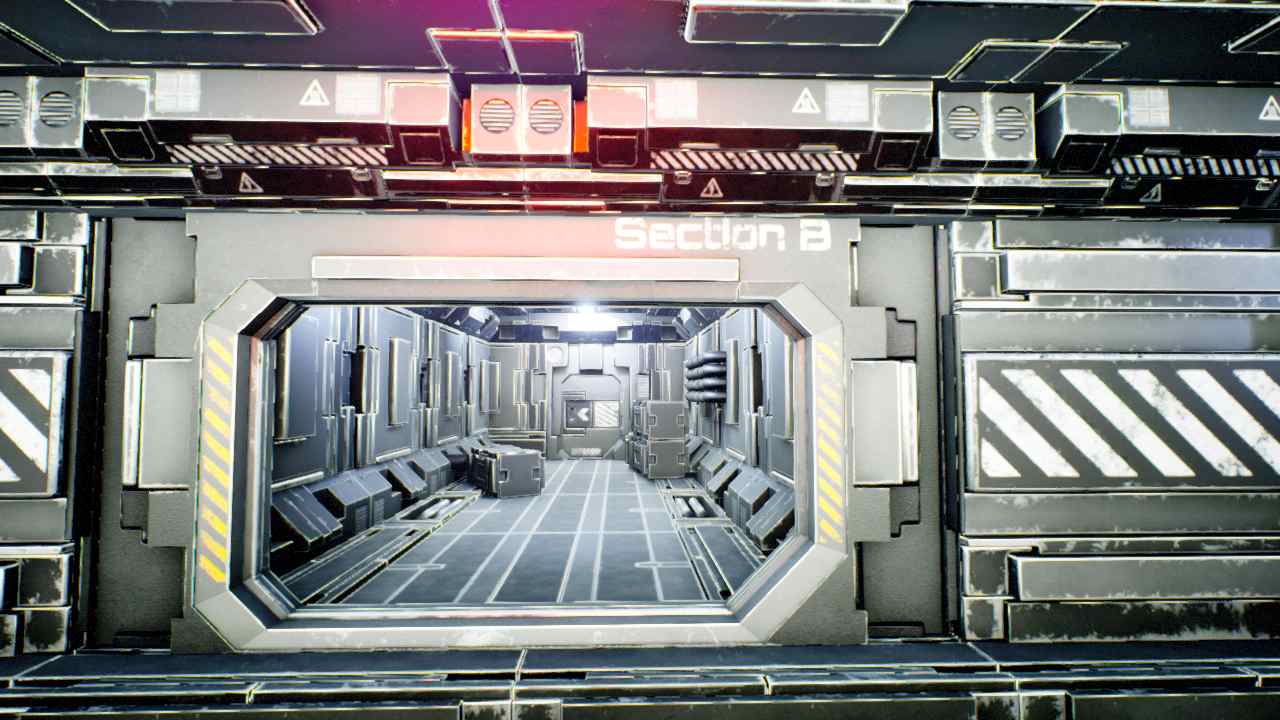}{Sci-Fi Station}{Indoor} &
		\sceneitem{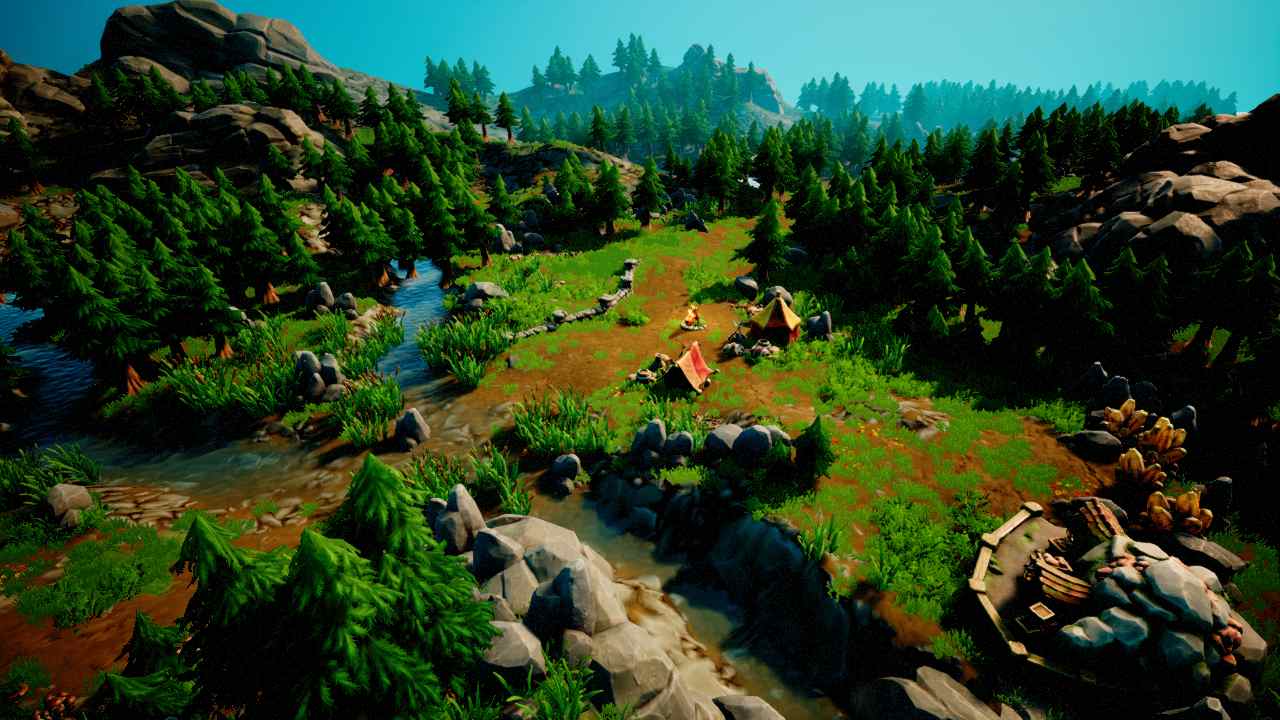}{Fantasy Forest}{Outdoor} &
		\sceneitem{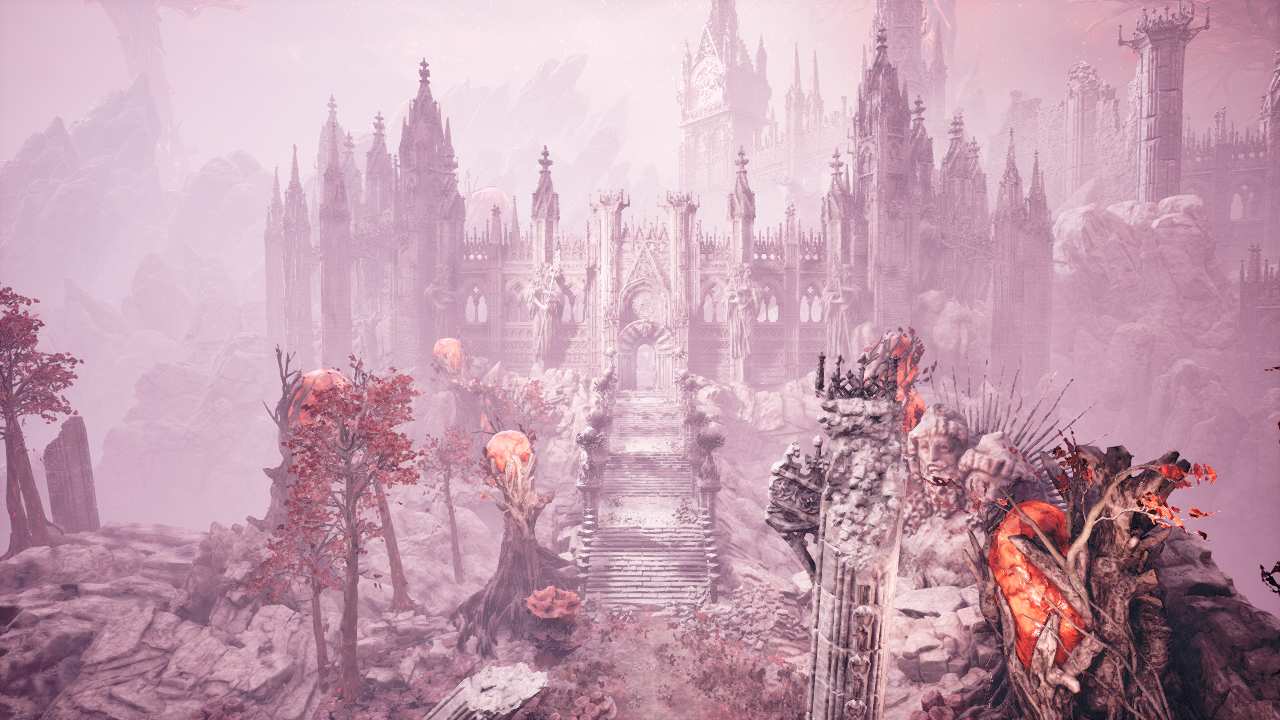}{Alien Castle}{Outdoor} \\
		
		\vspace{-3mm} \\
		
		\sceneitem{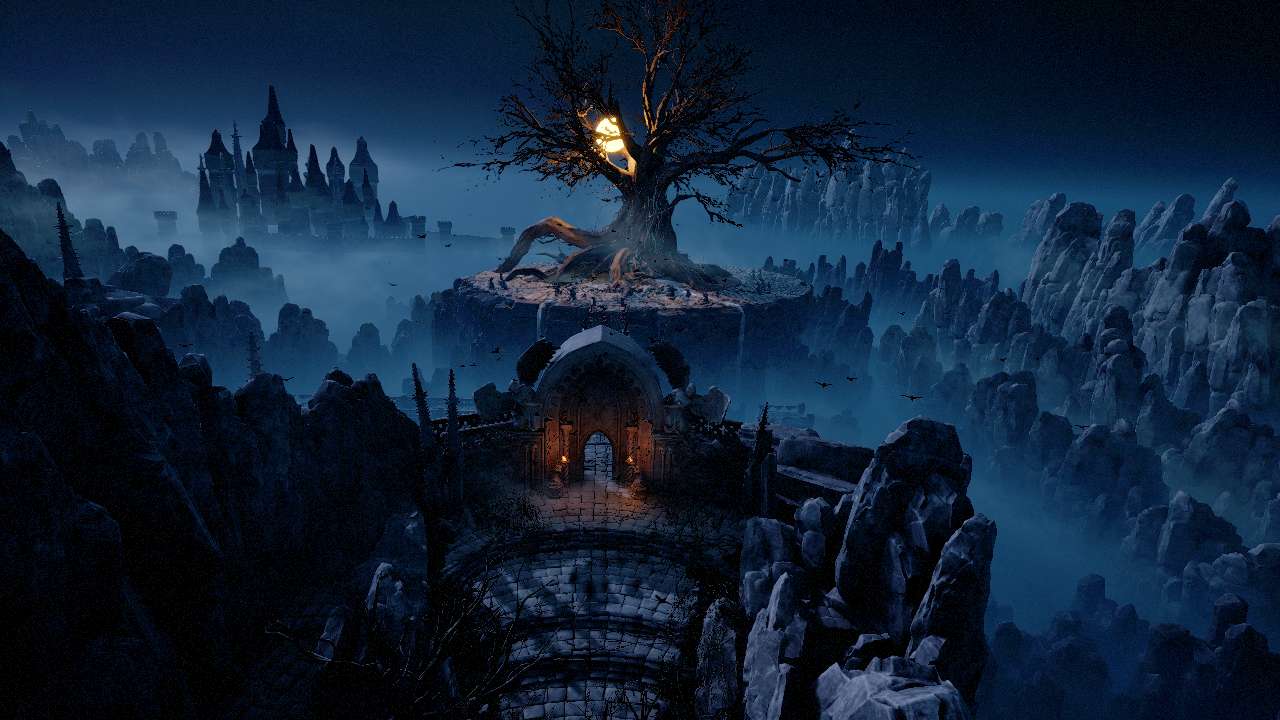}{Giant Tree Forest}{Outdoor} &
		\sceneitem{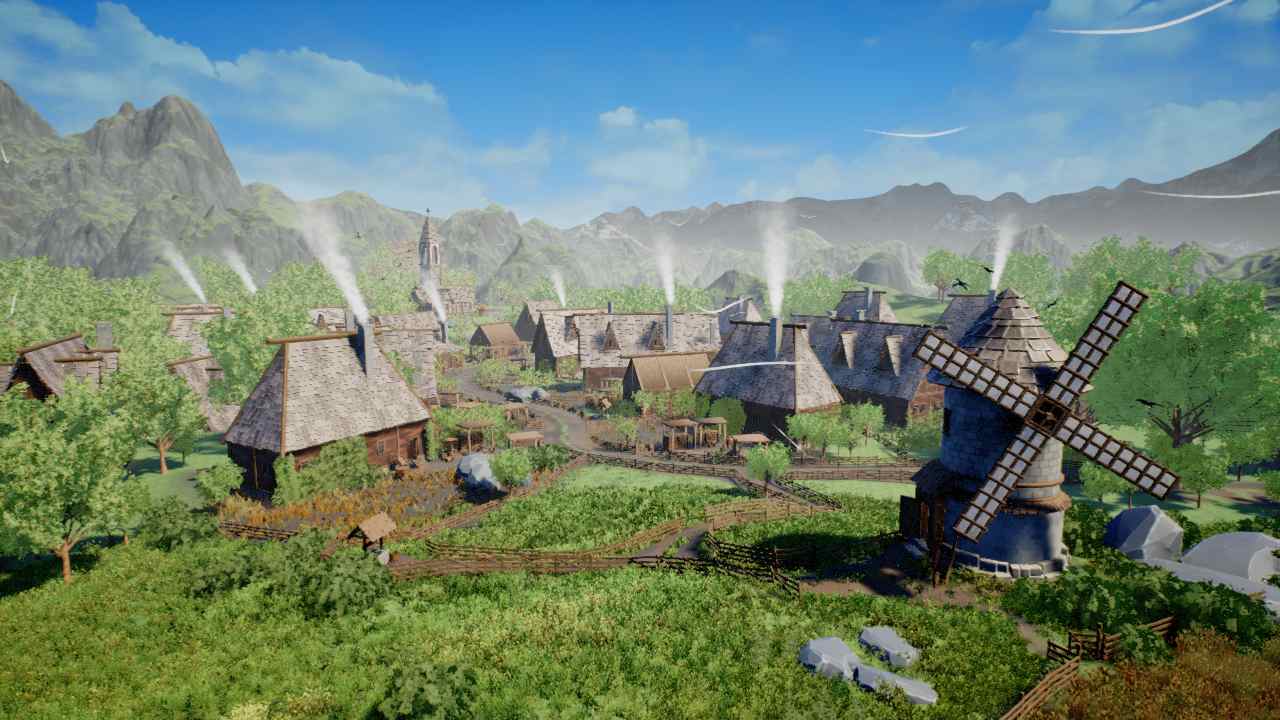}{Medieval Village}{Outdoor} &
		\sceneitem{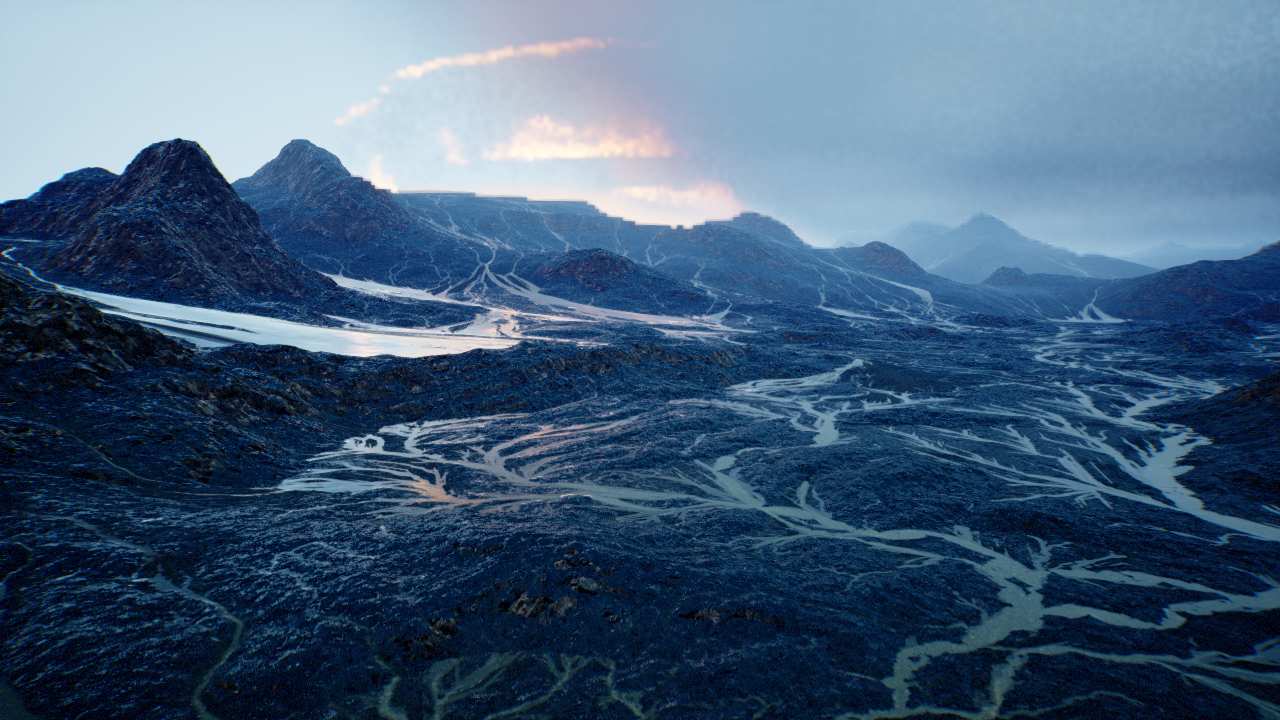}{Dark Rock Terrain}{Outdoor} &
		\sceneitem{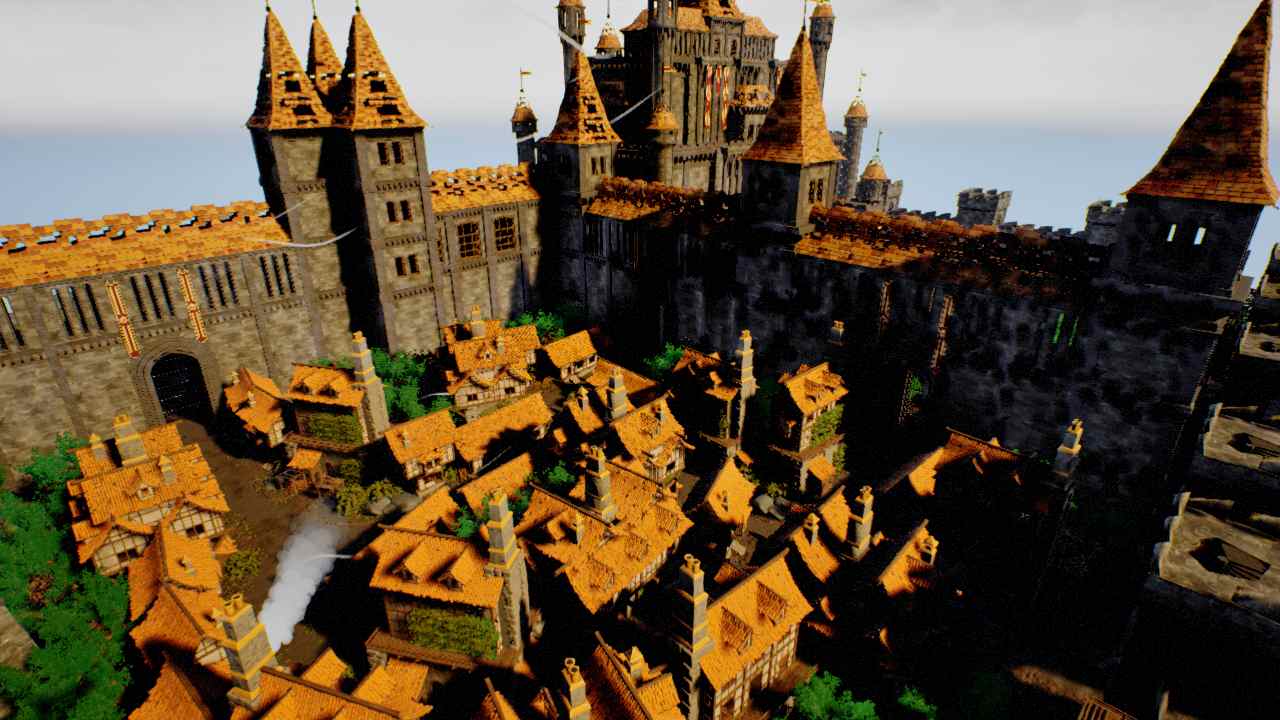}{Medieval Kingdom}{Outdoor} &
		\sceneitem{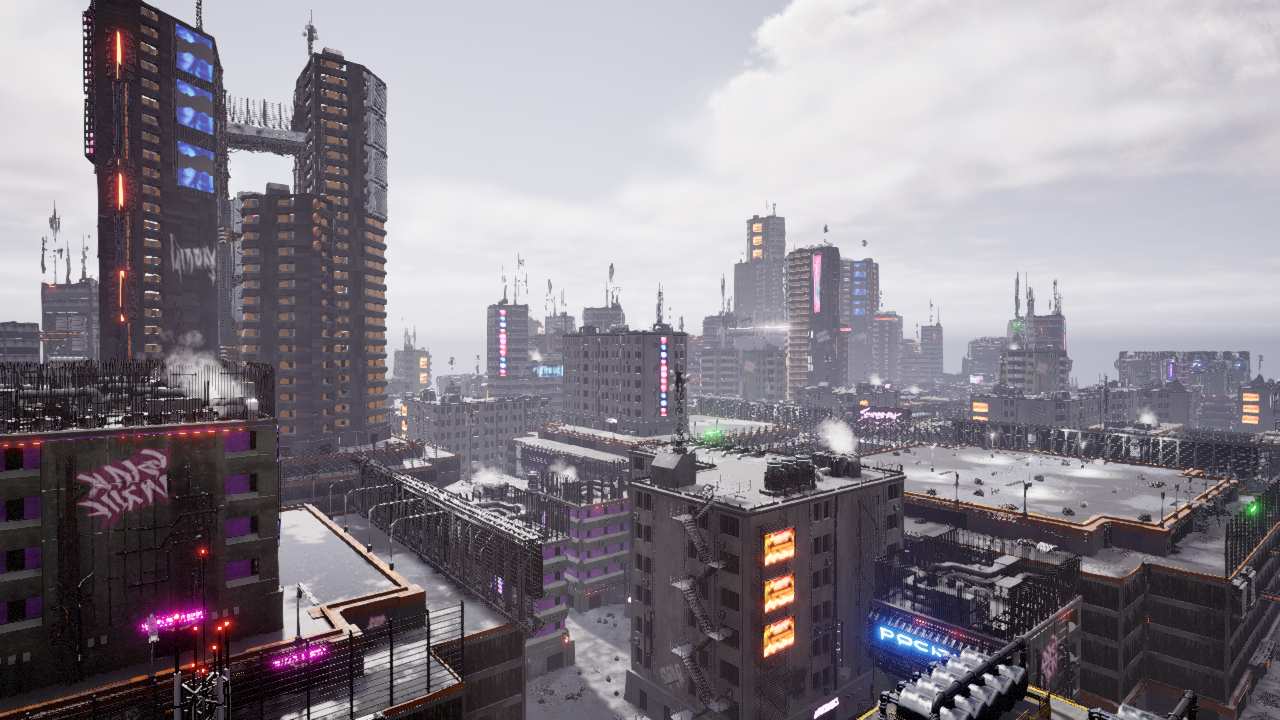}{Cyberpunk City}{Outdoor} \\
		
		\vspace{-3mm} \\
		
		\sceneitem{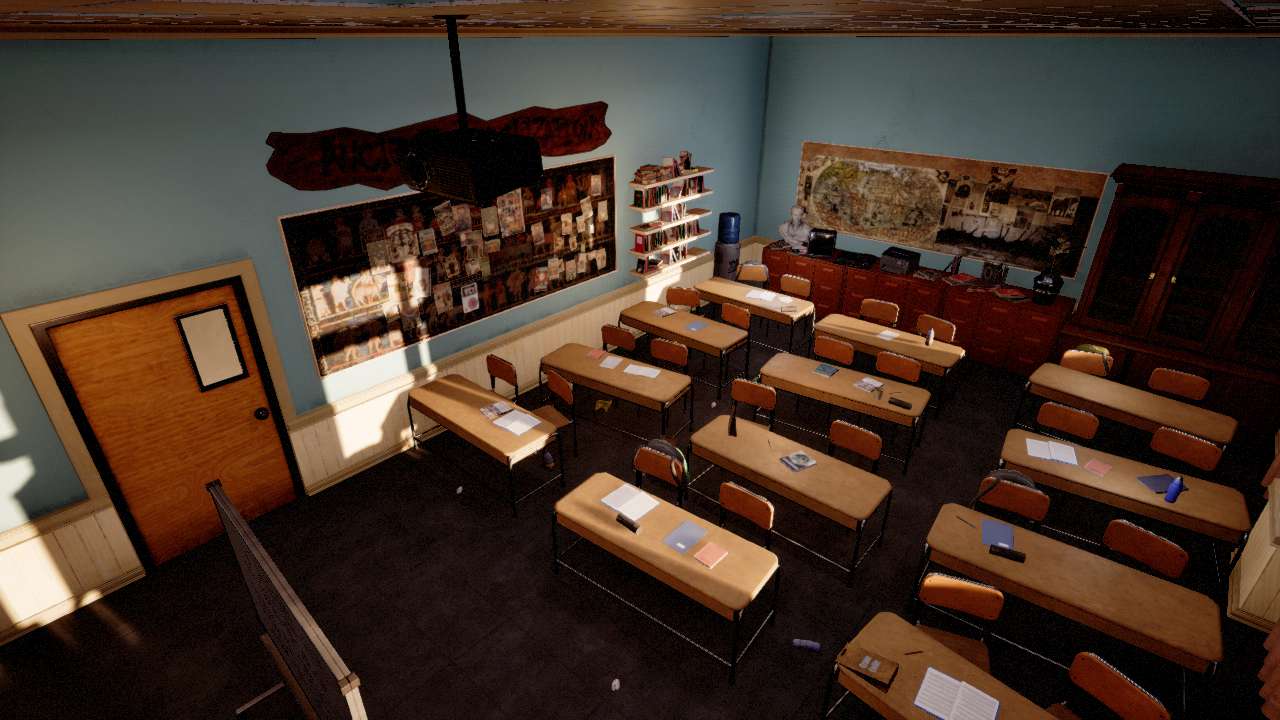}{School Classroom}{Indoor} &
		\sceneitem{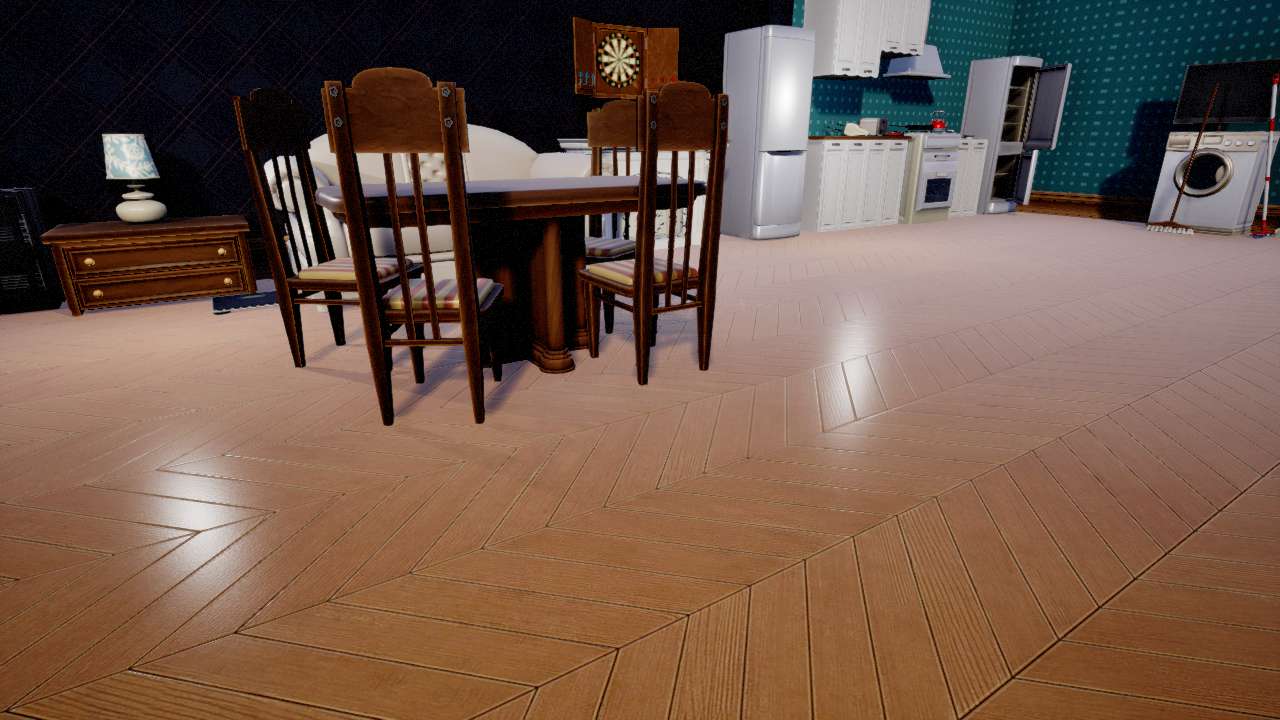}{Stylized Interior}{Indoor} &
		\sceneitem{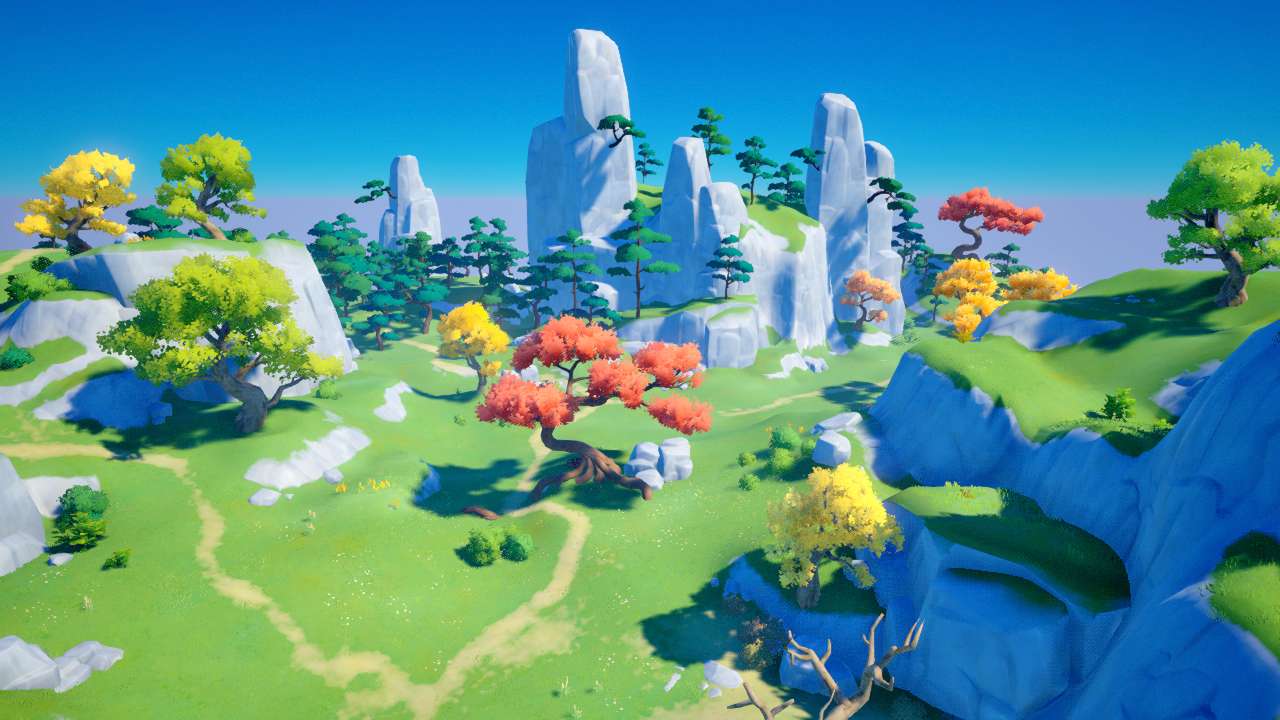}{Asian Forest}{Outdoor} &
		\sceneitem{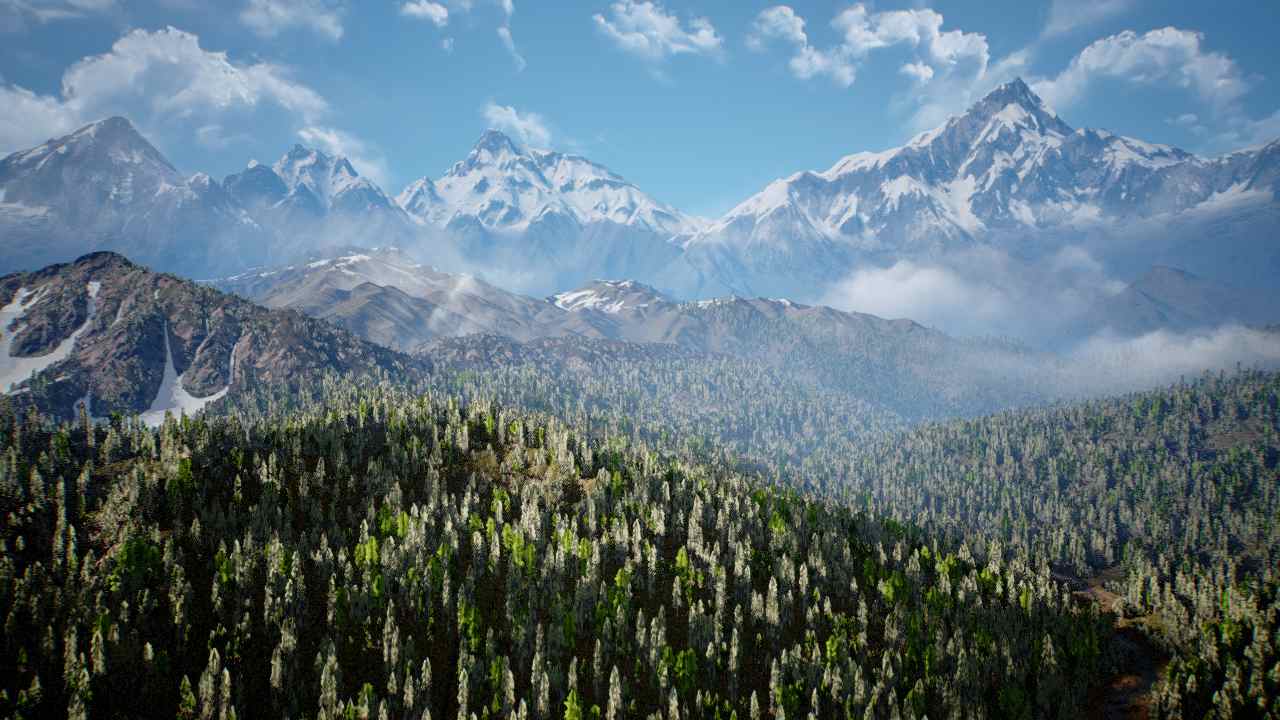}{Mountain Forest}{Outdoor} &
		\sceneitem{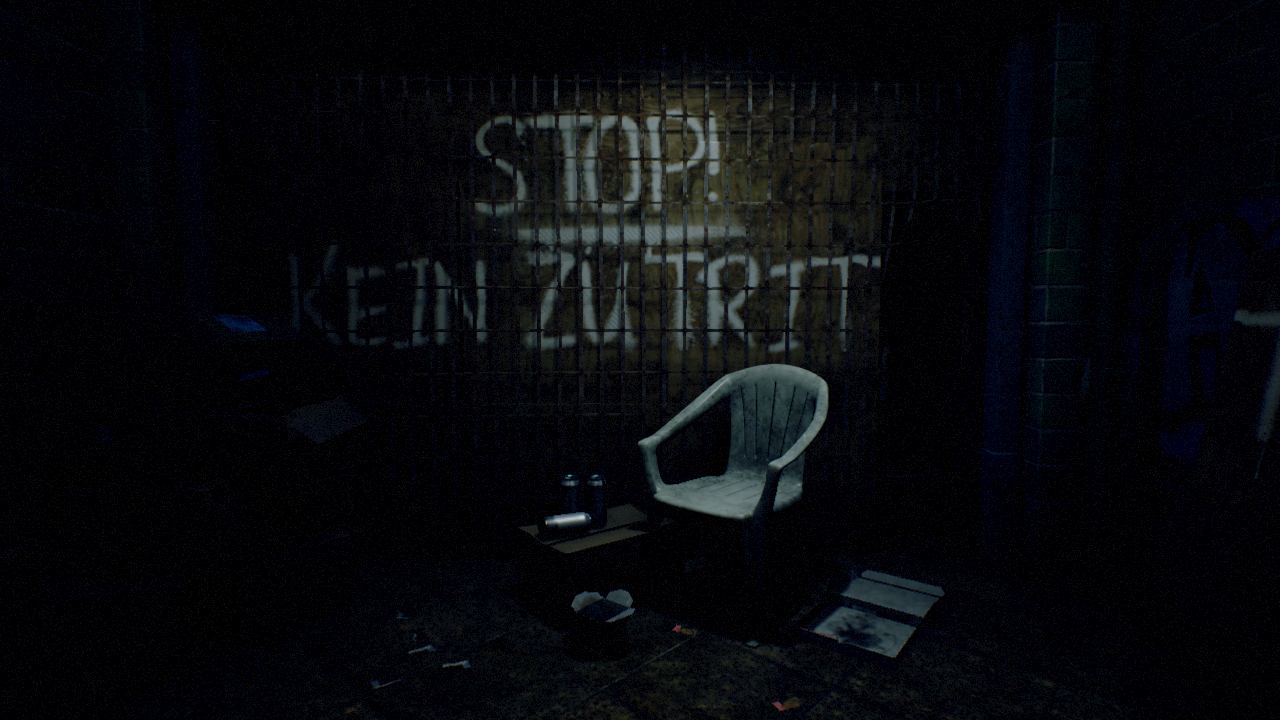}{Abandoned Subway}{Indoor} \\
		
		\vspace{-3mm} \\
		
		\sceneitem{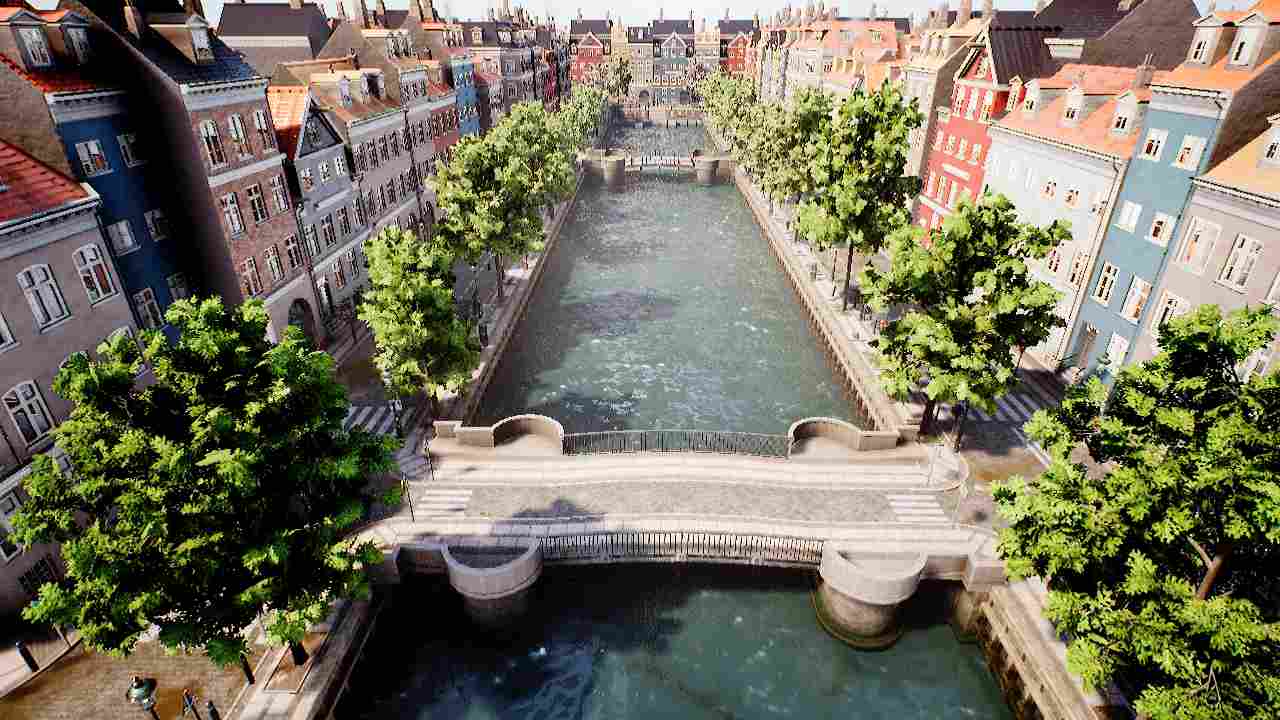}{Water Town}{Outdoor} &
		\sceneitem{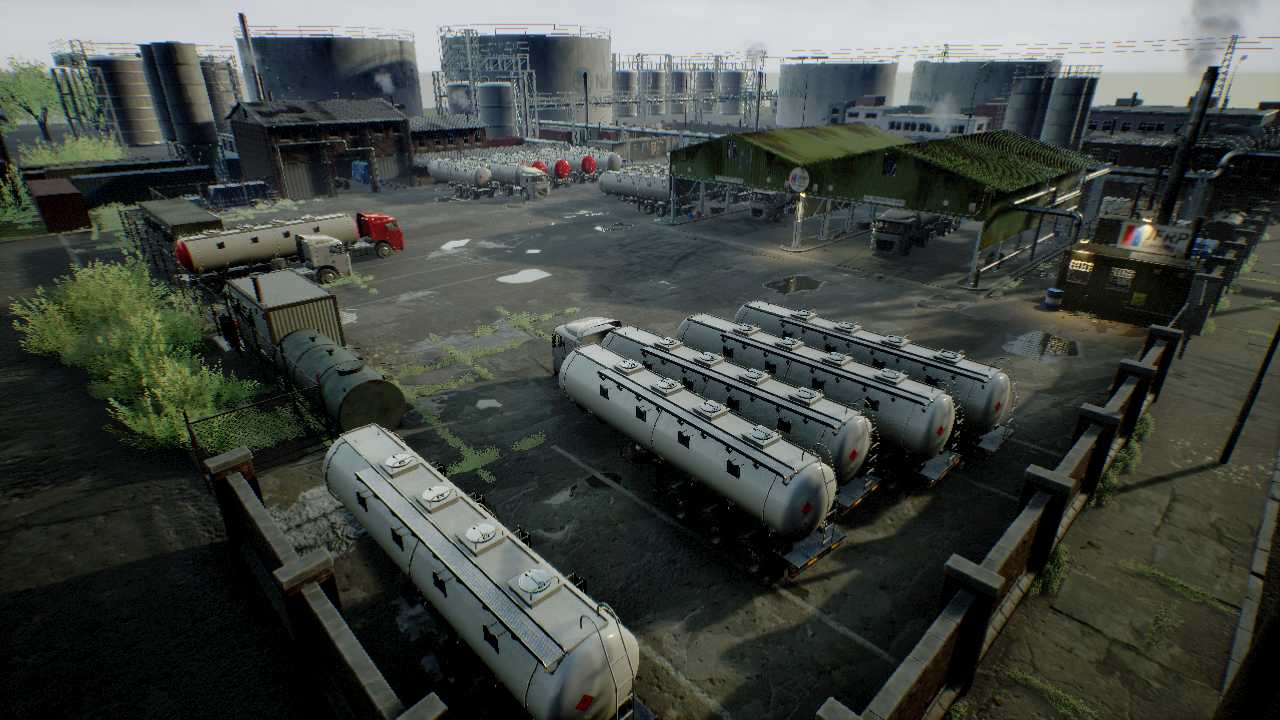}{Nafutsha Factory}{Outdoor} &
		\sceneitem{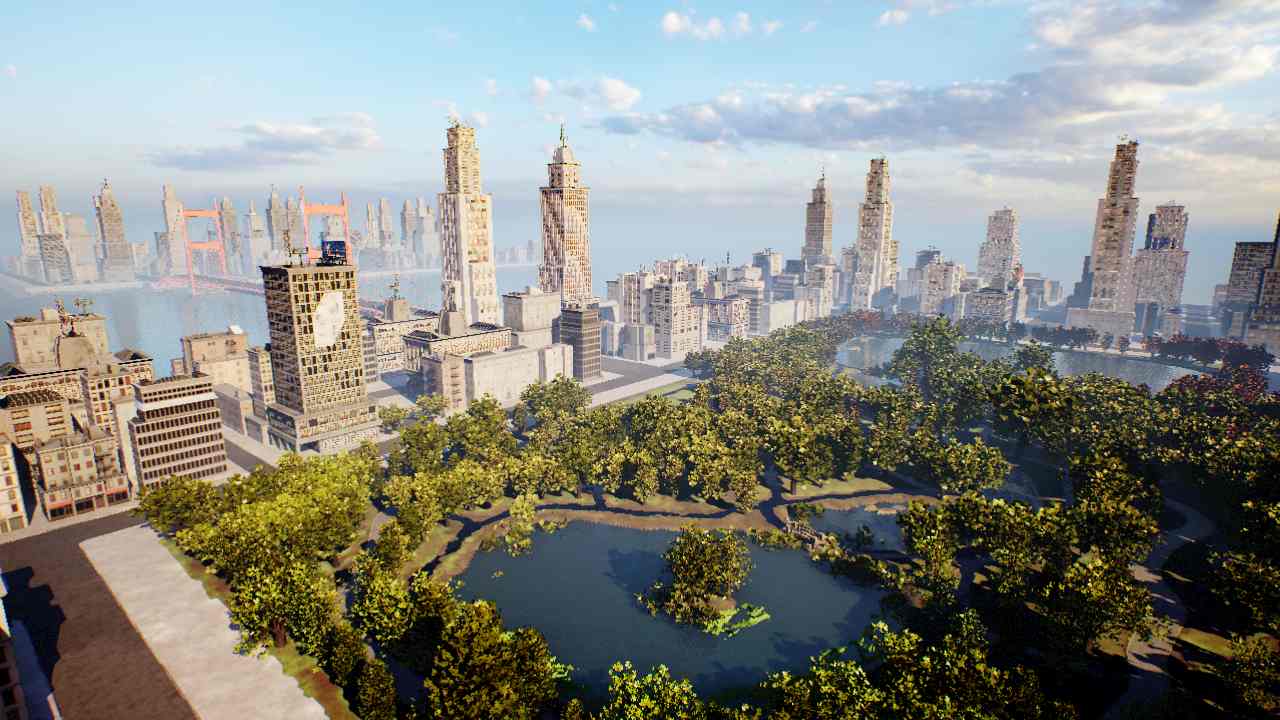}{City}{Outdoor} &
		\sceneitem{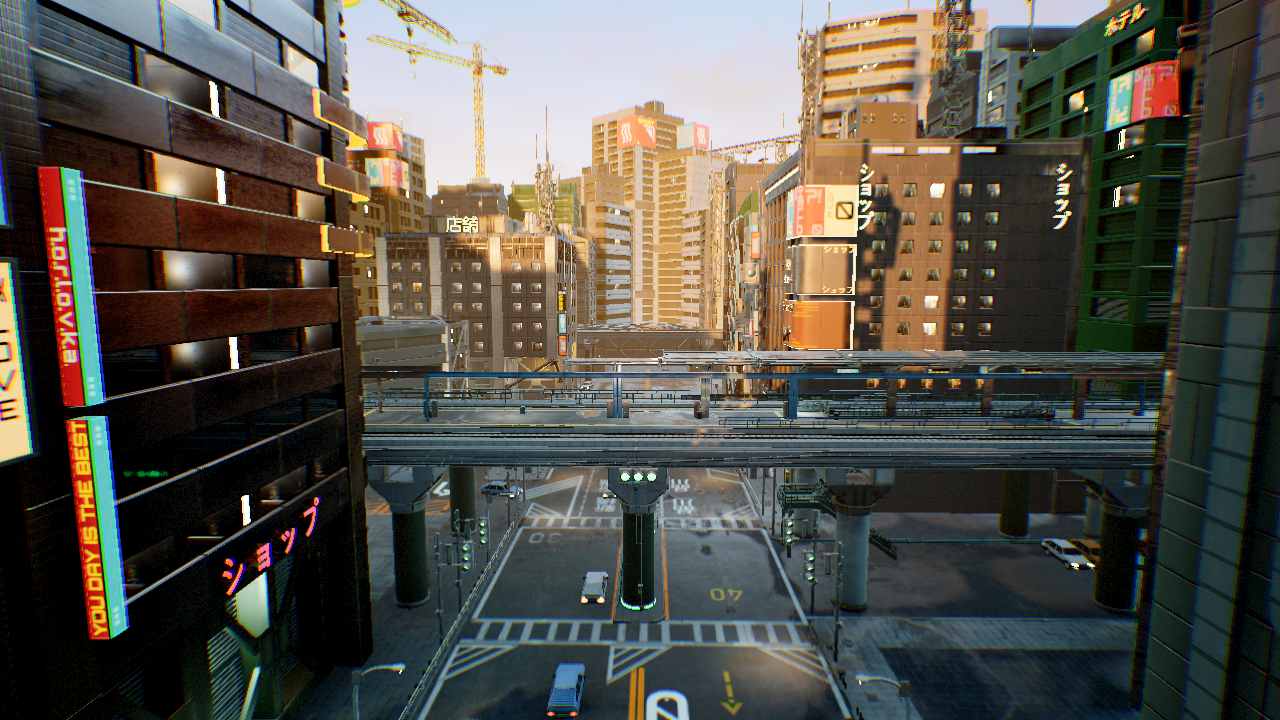}{Roadside Vehicles}{Outdoor} &
		\sceneitem{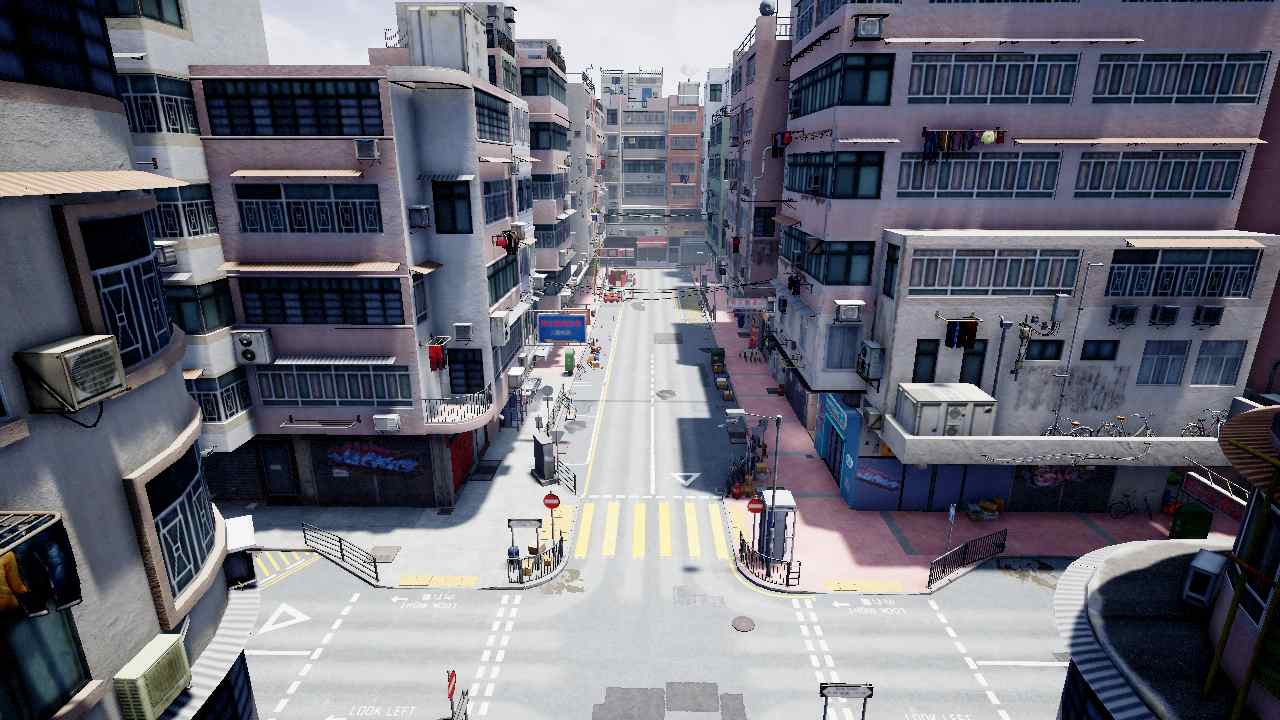}{Town}{Outdoor} \\
		
	\end{tabular}
\end{table*}

\subsection{Embodied Data Collection on the RoboTwin Platform}
\noindent\textbf{RoboTwin 2.0.}\cite{chen2025robotwin}
RoboTwin 2.0 is an embodied platform for bimanual manipulation in simulation, integrating unified interfaces, robot embodiments, and success-checking logic. By instantiating tasks from predefined configurations, the simulator records synchronized observations, actions, and metadata for every successful episode. This design ensures scalable data collection while preserving the spatial-temporal and semantic information essential for downstream policy learning and reasoning.

Training data is collected using a distributed pipeline that replays successful seeds. For each episode, the system archives multi-modal data, including HDF5 files, rendered videos, and action trajectories. All successfully parsed videos are rendered at a standard resolution and frame rate, with video durations varying across episodes.

\subsection{Seedance 2.0 Embodied Data Augmentation Pipeline}
\label{app:seedance}
\begin{figure}[htbp]
	\centering
	\includegraphics[width=\linewidth]{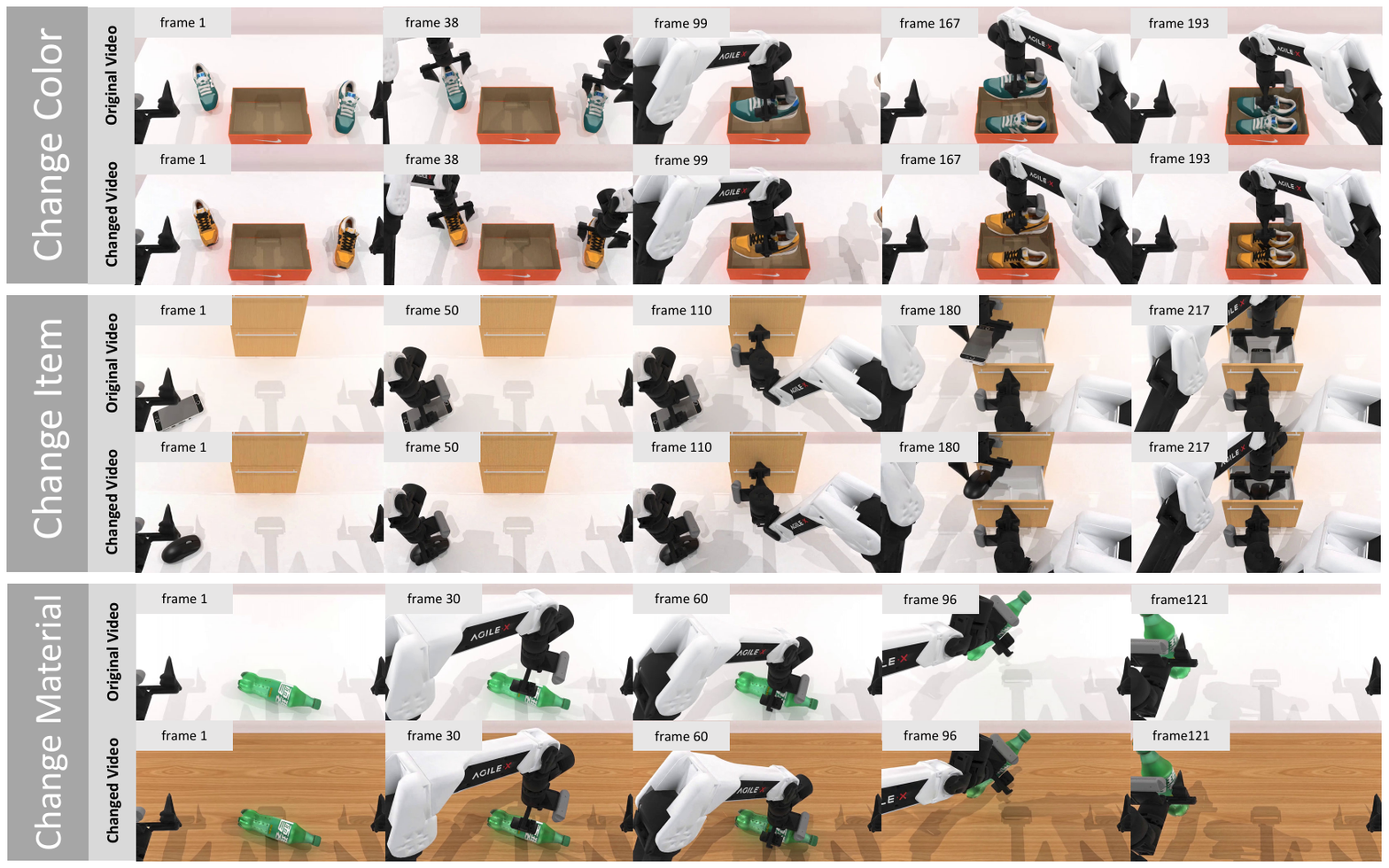}
	\caption{\textbf{Seedance 2.0 Data Augmentation.} We use Seedance 2.0 to diversify the embodied data in terms of object color, tabletop material, and manipulated objects. The results demonstrate consistent generation quality and interaction dynamics while keeping the robotic arm motion unchanged.}
	\label{fig:seedance_aug}
\end{figure}

Collecting real-world embodied interaction videos is costly and often suffers from limited visual diversity. Models trained solely on raw data are therefore prone to overfitting to specific object colors, appearances, materials, or scene textures, rather than learning the underlying relationship between control signals and physical interaction dynamics. To improve visual generalization while preserving the validity of action supervision, we build an embodied data augmentation pipeline based on Seedance 2.0~\cite{seedance2026seedance}, as illustrated in Fig.~\ref{fig:seedance_aug}.

Given an original embodied interaction video together with its control annotations, we use Seedance 2.0 as an instruction-guided video editing model to generate visually augmented versions of the same interaction process. We use Qwen3.6-Plus~\cite{qwen36plus} to batch-generate such instructions. The editing process is explicitly restricted to local appearance variations. For example, the color of the manipulated object can be changed, such as replacing the color of a shoe, while keeping surrounding objects, the background, object geometry, camera trajectory, hand or robot motion, contact relations, and the final task state unchanged. Similarly, for grasping or lifting sequences, the appearance of the manipulated object can be replaced or modified while preserving the original grasp trajectory, the object motion pattern, and the final lifted pose. In this way, the augmented samples introduce new visual variations without altering the underlying embodied interaction behavior.

Since the augmentation process is designed to preserve the physical interaction process, the generated videos inherit the control annotations of the original samples, including camera trajectories, action maps, and low-dimensional robot action vectors. When the modified visual attributes are explicitly mentioned in the text description, we update the caption accordingly; otherwise, we retain the original high-level task description. In addition, we filter out edited samples that change the task semantics, alter object motion, break contact relations, modify the final task outcome, or introduce obvious temporal inconsistency artifacts, thereby ensuring that the augmented data remain aligned with the original demonstrations in both temporal consistency and physical semantics.

This pipeline provides a practical way to scale the training data for embodied world models without recollecting robot or human demonstrations. By decoupling visual appearance variation from action supervision, the model can observe the same control signal under different visual realizations, which encourages it to learn action-conditioned physical dynamics rather than memorizing specific object textures, colors, or scene appearances. As a result, this data augmentation strategy improves robustness to object appearance changes, scene style variations, and out-of-distribution visual conditions.

\subsection{Hand-Joint Data Construction}

\noindent\textbf{Unified Hand Annotation with HaMeR.}
To remove annotation-format discrepancies between EgoDex and Ego4D, we process all selected video frames with a unified HaMeR annotation pipeline. HaMeR \citep{pavlakos2024reconstructing} is a fully Transformer-based method for hand mesh recovery. It uses a large-scale Vision Transformer backbone and a Transformer decoder to regress hand parameters from a monocular RGB image. The method adopts the MANO parametric hand model \citep{romero2017mano} to represent hand geometry. Given a hand crop $I_t$, HaMeR predicts the MANO pose parameters, shape parameters, and camera parameters:
\[
\Theta_t = f_{\mathrm{HaMeR}}(I_t) = \{\theta_t, \beta_t, \pi_t\}.
\]
Here, $\theta_t$ denotes the articulated hand pose, $\beta_t$ denotes the hand shape, and $\pi_t$ denotes the camera-related parameters. The MANO layer then produces a hand mesh and 21 3D joints:
\[
(M_t, X_t) = \mathrm{MANO}(\theta_t, \beta_t), \quad X_t \in \mathbb{R}^{21 \times 3}.
\]
We process the left and right hands separately. For each frame $t$ and hand $h \in \{L,R\}$, the annotation consists of 3D joint coordinates $X_{t,h} \in \mathbb{R}^{21 \times 3}$, joint pose matrices $R_{t,h} \in SO(3)^{21}$, and 2D pixel coordinates $U_{t,h} \in \mathbb{R}^{21 \times 2}$ obtained by projecting the 3D joints into the video plane. Both hands preserve the standard 21-joint hand topology, including the wrist and finger joints. Frames with failed hand detection, severe occlusion, implausible reprojection, or unstable temporal behavior are filtered using detection confidence, joint reprojection validity, and temporal consistency checks.

\noindent\textbf{Camera-Aware 3D-to-2D Projection.}
To align the hand-control signal with the RGB video pixels, we transform the reconstructed 3D joints into the camera coordinate system of each video frame and project them to the image plane with the corresponding camera intrinsics. Let $X^{w}_{t,h,j} \in \mathbb{R}^{3}$ be the 3D coordinate of the $j$-th joint of hand $h$ at frame $t$ in the world or reconstruction coordinate system, let $T^{c \leftarrow w}_t = [R^{c \leftarrow w}_t \mid \mathbf{t}^{c \leftarrow w}_t]$ be the camera extrinsic matrix, and let $K_t$ be the camera intrinsic matrix. The camera-space joint is computed as
\[
\tilde{X}^{c}_{t,h,j} = R^{c \leftarrow w}_t X^{w}_{t,h,j} + \mathbf{t}^{c \leftarrow w}_t.
\]
The homogeneous image projection is
\[
\tilde{u}_{t,h,j} \sim K_t \tilde{X}^{c}_{t,h,j},
\]
and the pixel coordinate is obtained by perspective division:
\[
u_{t,h,j} =
\left(
\frac{\tilde{u}^{(x)}_{t,h,j}}{\tilde{u}^{(z)}_{t,h,j}},
\frac{\tilde{u}^{(y)}_{t,h,j}}{\tilde{u}^{(z)}_{t,h,j}}
\right).
\]
For EgoDex, the original release provides camera intrinsics and per-frame camera/joint SE(3) transforms, which allows direct reprojection into the video frame. For Ego4D, we apply the same projection procedure using the available video metadata or estimated camera parameters. After this step, each hand joint is represented by a 2D coordinate in the same pixel coordinate system as the corresponding RGB frame.

\noindent\textbf{Action-Map Representation.}
Given the projected left- and right-hand joints
\[
U_t = \{U_{t,L}, U_{t,R}\}, \quad U_{t,h} \in \mathbb{R}^{21 \times 2},
\]
we construct a hand-joint action map $A_t$ for each video frame. The action map is a pixel-aligned control condition that binds the 3D hand motion signal to the 2D video space. We rasterize the 42 hand joints from both hands in the original image coordinate system while preserving hand identity and joint topology. The action map can be represented as multi-channel joint heatmaps, sparse coordinate maps, or skeleton-rendered maps. During Worldscape-MoE training, $A_t$ shares the same temporal index and spatial coordinate frame as the RGB frame $I_t$, and can therefore be directly used as the control input for the diffusion world model. For a video clip $\{I_t\}_{t=1}^{T}$, we obtain a synchronized control sequence $\{A_t\}_{t=1}^{T}$, enabling the model to learn the dynamic correspondence among hand motion, object response, and scene evolution.

\section{Additional Experimental Details}
\label{app:expriment}

\subsection{Model Inference Configurations}
Table~\ref{tab:world-model-details} summarizes the detailed inference configurations of the evaluated world-generation models, including model version, capability type, generation resolution, video length, and frame rate. These settings are provided to ensure a fair comparison across models with different official releases and inference protocols.
\begin{table*}[htbp]
	\centering
	\caption{Detailed inference configurations of world generation models evaluated in our benchmark. Version denotes the official release date of each model.}
	\label{tab:world-model-details}
	\small
	\renewcommand{\arraystretch}{1.1}
	\setlength{\tabcolsep}{5pt}
	\begin{tabular}{l c c c c c}
		\toprule
		Method & Version & Ability & Resolution & Length (s) & FPS \\
		\midrule
		Matrix-Game 3.0 & 25.08.12 & I2V & 640$\times$352 & 3.4 & 24 \\
		HY-World 1.5 & 25.12.17 & I2V & 832$\times$480 & 3.2 & 24 \\
		CameraCtrl & 24.04.02 & I2V & 256$\times$256 & 7.6 & 10 \\
		MotionCtrl & 23.12.06 & I2V & 256$\times$256 & 7.6 & 10 \\
		CamI2V & 25.07.12 & I2V & 512$\times$320 & 7.6 & 10 \\
		RealCam-I2V & 25.07.12 & I2V & 896$\times$512 & 5 & 16 \\
		VideoX-Fun-Wan & 25.10.16 & I2V & 832$\times$480 & 5 & 16 \\
		AC3D & 25.04.01 & I2V & 720$\times$480 & 6 & 8 \\
		ASTRA & 25.12.09 & I2V & 832$\times$480 & 4 & 20 \\
		\midrule
        HunyuanVideo-1.5 & 25.11.20 & I2V & 848$\times$480 & 5 & 24 \\
		Cosmos-Predict 2.5 & 25.10.06 & I2V & 1280$\times$704 & 5 & 16 \\
		MimicMotion & 24.07.08 & I2V & 1024$\times$576 & 4.8 & 15 \\
                MagicPose & 24.04.03 & I2V & 512$\times$512 & 4 & 15 \\
		LOME & 26.04.05 & I2V & 832$\times$480 & 5 & 15 \\
		\bottomrule
	\end{tabular}
\end{table*}

\subsection{Details of EWM Score}
\label{app:ewm_score_details}

Embodied World Models (EWMs) are generative models designed to predict future environment states conditioned on current observations and external instructions or robot actions. Unlike general-purpose video generation models, EWMs must capture physically grounded and action-consistent dynamics, effectively functioning as mental simulators for action planning or environment proxies for scalable robotic training.

To systematically evaluate the capabilities of our proposed framework and existing baselines, we adopt the EWMScore, a holistic metric originating from the WorldArena \cite{shang2026worldarena} benchmark. WorldArena provides a unified evaluation framework specifically designed for embodied world models, assessing both their perceptual fidelity and functional utility in downstream decision-making tasks. The EWMScore integrates multi-dimensional performance into a single interpretable index by computing the arithmetic mean of 16 normalized video metrics. 

\begin{table}[htbp]
\centering

\caption{Video quality evaluation results across visual quality, motion quality and content consistency dimensions.}
\label{tab:benchmark_part1}
\small
\setlength{\tabcolsep}{6pt}
\resizebox{\textwidth}{!}{
\begin{tabular}{l|ccc|ccc|ccc}
\toprule
\multirow{2}{*}{\textbf{Models}} & \multicolumn{3}{c|}{\textbf{Visual Quality}} & \multicolumn{3}{c|}{\textbf{Motion Quality}} & \multicolumn{3}{c}{\textbf{Content Consistency}} \\
\cmidrule(lr){2-4} \cmidrule(lr){5-7} \cmidrule(lr){8-10}
 & \makecell{Image\\Quality} & \makecell{Aesthetic\\Quality} & \makecell{JEPA\\Similarity} & \makecell{Dynamic\\Degree} & \makecell{Flow\\Score} & \makecell{Motion \\ Smoothness} & \makecell{Subject \\ Consist.} & \makecell{Background \\ Consist.} & \makecell{Photometric \\ Consist.}  \\
\midrule
\rowcolor[HTML]{E6F2FF} \textbf{Worldscape-MoE}\strut &0.4566  &0.3795  &0.8920 &0.4373  &0.2632  &0.7717  &0.8333  &0.9043  &0.1439  \\
\rowcolor[HTML]{E6F2FF} w/o MoE &0.5220 &0.4053 &0.8779 &0.4432 &0.2457 &0.7776 &0.8282 &0.8990 &0.1126 \\
GigaWorld-0 &0.5041  &0.3991  &0.4413  &0.6709  &0.3118  &0.7811  &0.7303  &0.8563  &0.1756  \\
TesserAct&0.3322  &0.4590  &0.4579  &0.5150  &0.2447  &0.7579  &0.8250  &\textbf{0.9238}  &0.2491  \\
RoboMaster &0.3487  & 0.3842 & 0.2966 &0.6124  & 0.1484 &0.6940  & 0.8295 &0.9123  &0.3356  \\
Vidar &0.4145  &0.4068  &0.5608  & 0.2767 &0.1426  & 0.7973 & 0.7629 &0.8300  &0.2350  \\
Cosmos-Predict 2.5 (text) &0.6668  &0.4501  &0.3126  &0.5911  &0.4302  &0.7882  &0.7488  &0.8511  &0.1383  \\
Cosmos-Predict 2.5 (action) &0.4489  &0.3576  & 0.9296 &0.3994  &0.0573  &0.7100  &0.8197  &0.8894  & 0.3528 \\
CtrlWorld &0.3522  &0.3893  &0.9185  &0.4257  &0.3449  &0.7377  &\textbf{0.8411}  &0.9057  &0.1729  \\
Wan 2.2 &0.3884  &0.3963  &0.7575  & 0.4349 &0.1269  &0.7019  &0.8388  &0.9042  &\textbf{0.4776}  \\
CogvideoX &0.3582  &0.3777  & \textbf{0.9384} &0.3166  &0.2189  &0.7391  &0.8083  & 0.8773 &0.3580  \\
IRASim &0.3489  &0.3623 &0.9330  &0.4139  &0.2083 &0.7052 & 0.8312 &0.9068 &0.3522\\
Veo 3.1 & 0.6605 &\textbf{0.4632}  &0.5694  &0.5450  & 0.1396 &0.6989  &0.7878  &0.8710  &0.3247  \\
Wan 2.6 &\textbf{0.6824}  &0.4433  &0.7229  &\textbf{0.7421}  &\textbf{0.4532}&\textbf{0.8539}  & 0.7517 &0.8687  &0.1904    \\
\bottomrule
\end{tabular}
}

\vspace{6mm}

\caption{Video quality evaluation results across physics adherence, 3D accuracy and controllability dimensions.}
\label{tab:benchmark_part2}
\small
\setlength{\tabcolsep}{2.3pt}
\begin{tabular}{l|cc|cc|cccc}
\toprule
\multirow{2}{*}{\textbf{Models}} & \multicolumn{2}{c|}{\textbf{Physics Adherence}} & \multicolumn{2}{c|}{\textbf{3D Accuracy}} & \multicolumn{3}{c}{\textbf{Controllability}} \\
\cmidrule(lr){2-3} \cmidrule(lr){4-5} \cmidrule(lr){6-8}
 & \makecell{Interaction \\ Quality} & \makecell{Trajectory \\ Acc.}  & \makecell{Depth \\ Acc.} & Perspectivity & \makecell{Instruction \\ Following} & \makecell{Semantic \\ Alignment} &  \makecell{Action \\ Following} \\
\midrule
\rowcolor[HTML]{E6F2FF} \textbf{Worldscape-MoE}\strut &\textbf{0.8008} &0.4610 &0.9030 &0.9686 &\textbf{0.9348} &\textbf{0.9039}  &0.0955  \\
\rowcolor[HTML]{E6F2FF} w/o MoE &0.7622 &0.3540 &0.9038 &\textbf{0.9744} &0.8703 &0.8914 &0.0324 \\
GigaWorld-0 &0.5368  &0.1552 &0.6316  &0.7596  &0.6156 &0.8591  &0.1134  \\
TesserAct&0.5800  &0.1396 &0.7159  &0.7920  &0.6152 &0.8783  &0.0311  \\
RoboMaster &0.5364  &0.1158 &0.8335  &0.7588  &0.5772 &0.8761  &0.0352  \\
Vidar &0.5348  &0.1928 &0.7872  &0.7592  &0.5912 &0.8826  &0.0819  \\
Cosmos-Predict 2.5 (text) &0.3872  &0.0816 & 0.7051 &0.7964  &0.2664 & 0.7733 &\textbf{0.1418}  \\
Cosmos-Predict 2.5(action) &0.5500  &0.2945 &0.8862  &0.7644  &0.5840 &0.8879  &0.0133  \\
CtrlWorld &0.6212  &\textbf{0.4766} &0.9300  &0.7960  &0.7272 & 0.8912 &0.0210  \\
Wan 2.2 &0.5184  &0.1627 &0.7768  &0.7660  &0.5376 &0.8877  & 0.0512 \\
CogvideoX &0.5940  &0.3526 & 0.9097 &0.7828  &0.7268 &0.8977  &0.0076  \\
IRASim & 0.5656 &0.3639 &\textbf{0.9312}  &0.7788 &0.6604 &0.8849&0.0526  \\
Veo 3.1 &0.7872  &0.1231 &0.7421  &0.8276  &0.9328 & 0.8607 &0.0852  \\
Wan 2.6 & 0.7280 &0.1182 &0.7144  &0.8032  &0.8536 &0.8728  & 0.0992 \\
\bottomrule
\end{tabular}

\end{table}

These metrics are systematically categorized into six major evaluation dimensions:

\begin{itemize}
    \item \textbf{Visual Quality}: Assesses whether the generated videos are perceptually reliable for embodied scenarios, evaluating low-level fidelity, perceptual appeal, and similarity to real data. The sub-metrics include Image Quality, Aesthetic Quality, and JEPA Similarity.
    \item \textbf{Motion Quality}: Reflects whether the model captures physically meaningful and temporally coherent dynamics. The sub-metrics consist of Dynamic Degree, Flow Score, and Motion Smoothness.
    \item \textbf{Content Consistency}: Measures the stability of objects and scenes throughout the video at both semantic and appearance levels. The sub-metrics are Subject Consistency, Background Consistency, and Photometric Consistency.
    \item \textbf{Physics Adherence}: Evaluates whether generated behaviors conform to real-world physical constraints rather than merely appearing visually plausible. The sub-metrics are Interaction Quality and Trajectory Accuracy.
    \item \textbf{3D Accuracy}: Assesses the preservation of real-world spatial structures and geometric consistency beyond 2D image appearance. The sub-metrics include Depth Accuracy and Perspectivity.
    \item \textbf{Controllability}: Measures the model's ability to accurately respond to and execute external action instructions. The sub-metrics consist of Instruction Following, Semantic Alignment, and Action Following.
\end{itemize}

The detailed quantitative evaluation results of our model and all other baselines across these dimensions are provided in the appendix tables. Specifically, Table~\ref{tab:benchmark_part1} reports the video quality evaluation results across the visual quality, motion quality, and content consistency dimensions. Meanwhile, Table~\ref{tab:benchmark_part2} details the evaluation results across the physics adherence, 3D accuracy, and controllability dimensions.

\paragraph{Analysis of Worldscape-MoE on EWMScore.}
As shown in Tables~\ref{tab:benchmark_part1} and \ref{tab:benchmark_part2}, Worldscape-MoE delivers consistently leading performance on the dimensions that most directly characterize an \emph{embodied} world model. In terms of \textbf{Physics Adherence}, it attains the best Interaction Quality (\textbf{0.8008}) among all baselines, surpassing the strongest closed-source models Veo~3.1 (0.7872) and Wan~2.6 (0.7280) and exceeding action-conditioned EWMs such as Cosmos-Predict~2.5 (action) (0.5500) and IRASim (0.5656) by a substantial margin, while its Trajectory Accuracy (0.4610) ranks second only to CtrlWorld (0.4766). On \textbf{3D Accuracy}, the Without MoE baseline obtains the highest Perspectivity (0.9744), while Worldscape-MoE remains highly competitive on Perspectivity (0.9686) and Depth Accuracy (0.9030), demonstrating that the generated futures faithfully preserve real-world spatial structure rather than relying on 2D appearance shortcuts. On \textbf{Controllability}, Worldscape-MoE further establishes new state-of-the-art results on both Instruction Following (\textbf{0.9348}) and Semantic Alignment (\textbf{0.9039}), outperforming all baselines, which indicates that Worldscape-MoE grounds external textual instructions far more reliably than appearance-driven generators. On the appearance-oriented dimensions, Worldscape-MoE remains competitive, with a strong JEPA Similarity (0.8920) and high Subject Consistency (0.8333) and Background Consistency (0.9043). Taken together, these results show that Worldscape-MoE advances the frontier of EWMs along the physically grounded, geometry-consistent, and instruction-controllable axes that the EWMScore is specifically designed to measure.

\subsection{MoE Scalability Analysis}
\label{app:weight}

\begin{figure}[htbp]
	\centering
	\includegraphics[width=\linewidth]{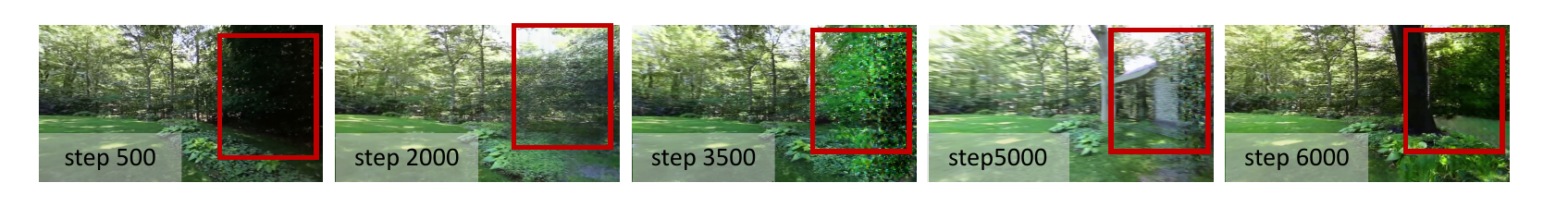}
	\caption{\textbf{Transient Degradation and Recovery During Modality Extension.} Qualitative results from 500 to 6500 training steps show the transient degradation of locomotion capability after introducing a new control modality, followed by gradual recovery as training proceeds. This effect is particularly pronounced in scenarios involving substantial scene changes.}
	\label{fig:transient_degradation}
\end{figure}

We analyzed the model convergence status and feature evolution under multimodal control by tracking the weight updates of each expert module across different training checkpoints.

\begin{table}[ht]
\centering
\small
\setlength{\tabcolsep}{3pt}
\renewcommand{\arraystretch}{1.0}
\caption{Expert weight L2 norm variation across different training step segments.}
\label{tab:module_segments}
\begin{tabular}{l|*{8}{c}}
\toprule
\textbf{Expert} & \textbf{250-750} & \textbf{750-1.25k} & \textbf{1.25-1.75k} & \textbf{1.75-2.25k} & \textbf{2.25-2.75k} & \textbf{2.75-3.25k} & \textbf{3.25-3.75k} & \textbf{3.75-4.25k} \\
\midrule
Shared       & 67.78 & 70.25 & 71.13 & 65.81 & 64.35 & 64.50 & 67.92 & 67.99 \\
Locomotion   & 115.60 & 122.62 & 125.02 & 116.48 & 116.34 & 119.70 & 138.31 & 130.17 \\
Manipulation & 134.89 & 137.50 & 139.52 & 126.24 & 123.10 & 122.23 & 126.33 & 125.71 \\
ActionMap    & 155.30 & 144.04 & 135.25 & 134.97 & 134.43 & 132.53 & 144.95 & 138.01 \\
\bottomrule
\end{tabular}
\end{table}

\paragraph{Weight update analysis.}
We track the cumulative L2 variation of each expert module across consecutive 500-step intervals in Table~\ref{tab:module_segments}.
The Shared expert remains the most stable throughout training, while Locomotion, Manipulation, and ActionMap exhibit larger but comparable update magnitudes, with ActionMap showing the strongest early adaptation and all three modality-specific experts gradually stabilizing after the first few segments. This pattern indicates that joint mixed training enables balanced co-adaptation across experts, rather than suppressing the locomotion pathway when new control modalities are introduced.

\noindent\textbf{Joint Training Strategy.}
Experimental results show that the expert initialization strategy significantly impacts the multimodal decoupling effect.If parameters are cloned from a pre-trained Locomotion expert to initiate mixed training for other experts, it leads to a significantly smaller magnitude of weight updates for the Locomotion expert. This approach distorts the original feature space, resulting in a partial collapse of locomotion control performance. In contrast, uniformly cloning and initializing all experts from a Dense Model backbone for multi-expert joint training facilitates a more stable and balanced decoupling effect across all experts.

\noindent\textbf{Transient Degradation During Modality Extension.}
When a new control modality is introduced, we observe a short period of locomotion degradation in the early stage of training, as shown in Fig.~\ref{fig:transient_degradation}. We attribute this behavior to a temporary adaptation phase, during which the shared and modality-specific parameters are jointly re-adjusted to absorb the newly introduced control knowledge. This phenomenon is particularly pronounced in locomotion cases that require large scene changes. Importantly, the degradation is not persistent: as training continues, the model gradually restores its original locomotion capability while retaining the newly learned modality. In the following, we provide qualitative cases from 500 to 6500 training steps to visualize this recovery process.

\subsection{Generalization to Single-Arm LIBERO Manipulation}
\label{app:libero_single_arm}

\begin{figure}[htbp]
	\centering
	\includegraphics[width=\linewidth]{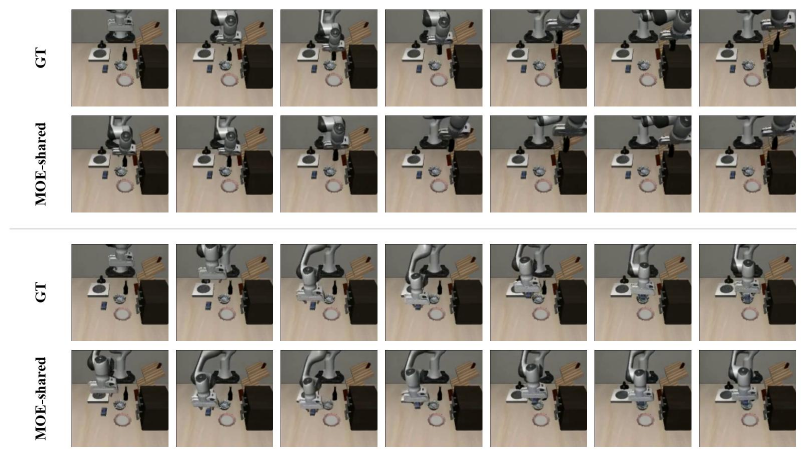}
	\caption{\textbf{Single-arm LIBERO generalization.} Each panel compares the ground-truth demonstration with the model initialized from the MoE shared expert after 5K fine-tuning steps. The figure is pre-rendered from normalized comparison videos by uniformly sampling seven frames from each rollout.}
	\label{fig:libero_single_arm_generalization}
\end{figure}

To further examine whether the MoE-trained shared expert learns transferable embodied dynamics, we conduct an additional extension experiment on the LIBERO single-arm manipulation benchmark. In this setting, each sample consists of an initial observation, a task language instruction, and the corresponding low-dimensional single-arm robot action trajectory from the LIBERO demonstrations. We process these demonstrations into action-conditioned video clips and use the official train/validation metadata for fine-tuning and held-out evaluation. Specifically, we initialize a single-FFN model from the MoE shared expert backbone and fine-tune it on the LIBERO single-arm split. As shown in Fig.~\ref{fig:libero_single_arm_generalization}, after only 5K fine-tuning steps, the shared-initialized model already produces action-conditioned rollouts that closely follow the ground-truth demonstrations in object motion, end-effector movement, and scene consistency. Importantly, this adaptation does not compromise the model's generative capability on other control modalities, which remains effective after fine-tuning, suggesting that the shared expert preserves broadly reusable world dynamics rather than overfitting to the single-arm manipulation domain.

This result provides qualitative evidence that the shared expert does not merely serve the original heterogeneous-control training tasks. Instead, it captures manipulation-relevant priors that can be transferred to a new single-arm setting with limited additional training. We emphasize this as a qualitative generalization study rather than a large-scale quantitative comparison: the goal is to show that the MoE-shared backbone can quickly adapt to unseen LIBERO manipulation cases and generate physically plausible futures under robot-action conditioning.

\subsection{More Qualitative Results of Worldscape-MoE}
\label{sec:More Qualitative Results}

In this section, we present more qualitative results of Worldscape-MoE.

\label{app:Qualitative Results}

\begin{figure}[htbp]
	\centering
	\includegraphics[width=\linewidth]{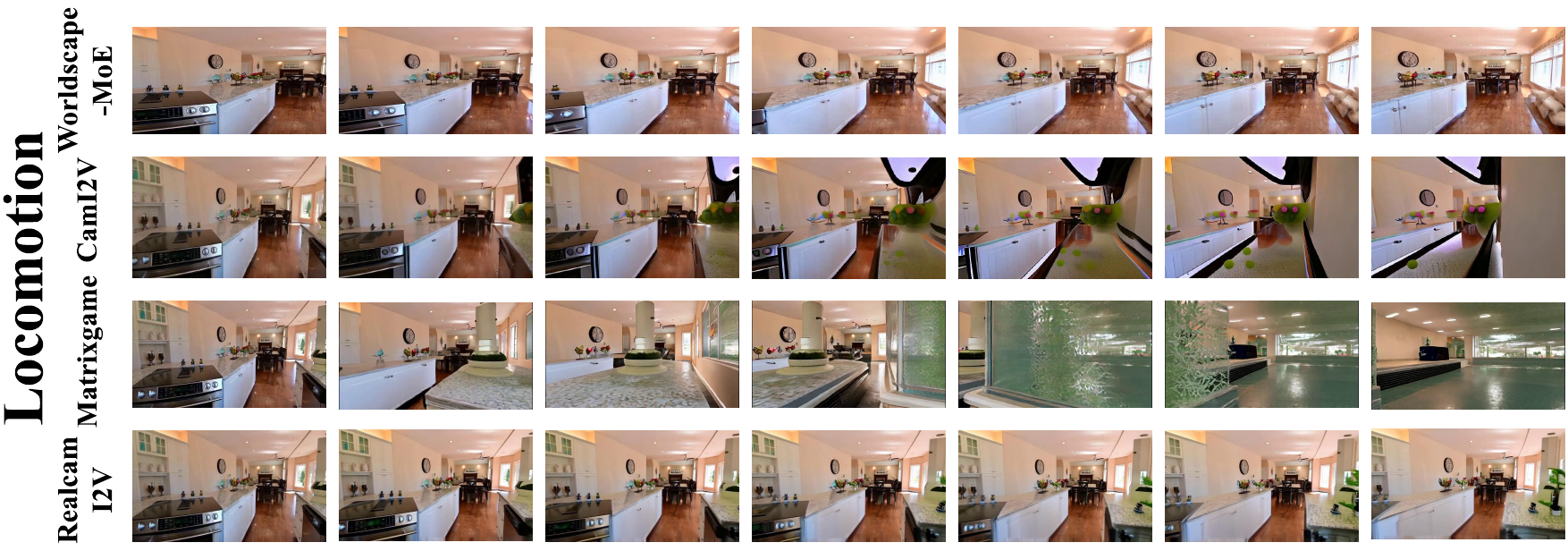}
    \caption{\textbf{Locomotion comparison.} Each row compares video frames generated by Worldscape-MoE, CamI2V, Matrixgame, and RealCam-I2V under the same locomotion trajectory, with columns uniformly sampled from each rollout.}
	\label{fig:case1}
    \vspace{-5pt}
\end{figure}
\begin{figure}[htbp]
	\centering
	\includegraphics[width=\linewidth]{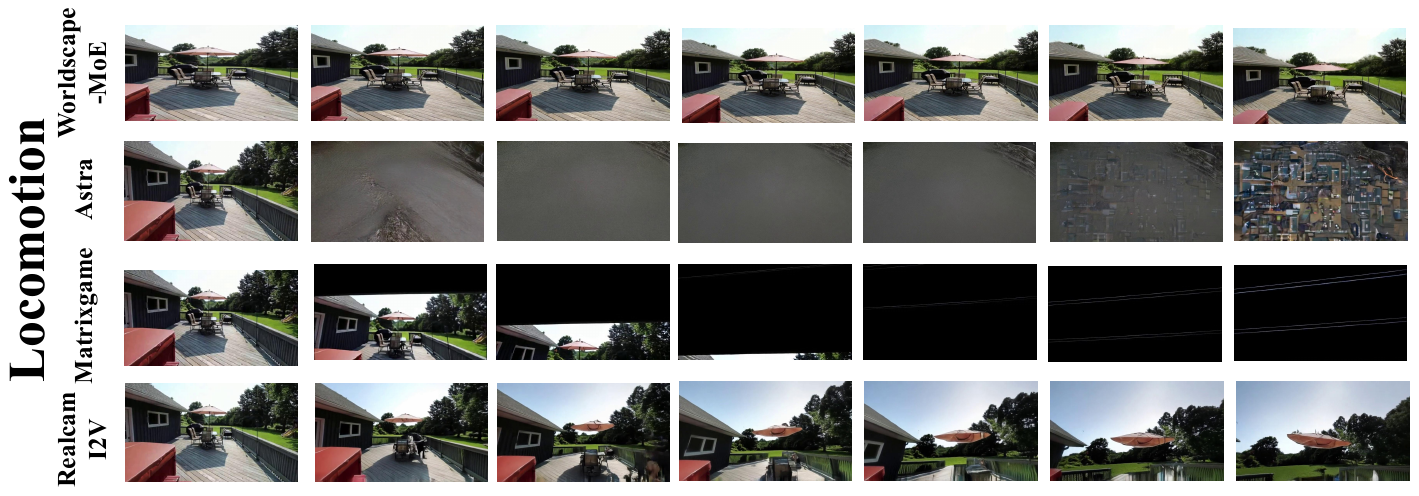}
	\caption{\textbf{Locomotion comparison.} Each row compares video frames generated by Worldscape-MoE, Astra, Matrixgame, and RealCam-I2V under the same locomotion trajectory, with columns uniformly sampled from each rollout.}
    \label{fig:case1-2}
    \vspace{-5pt}
\end{figure}
\begin{figure}[htbp]
	\centering
	\includegraphics[width=\linewidth]{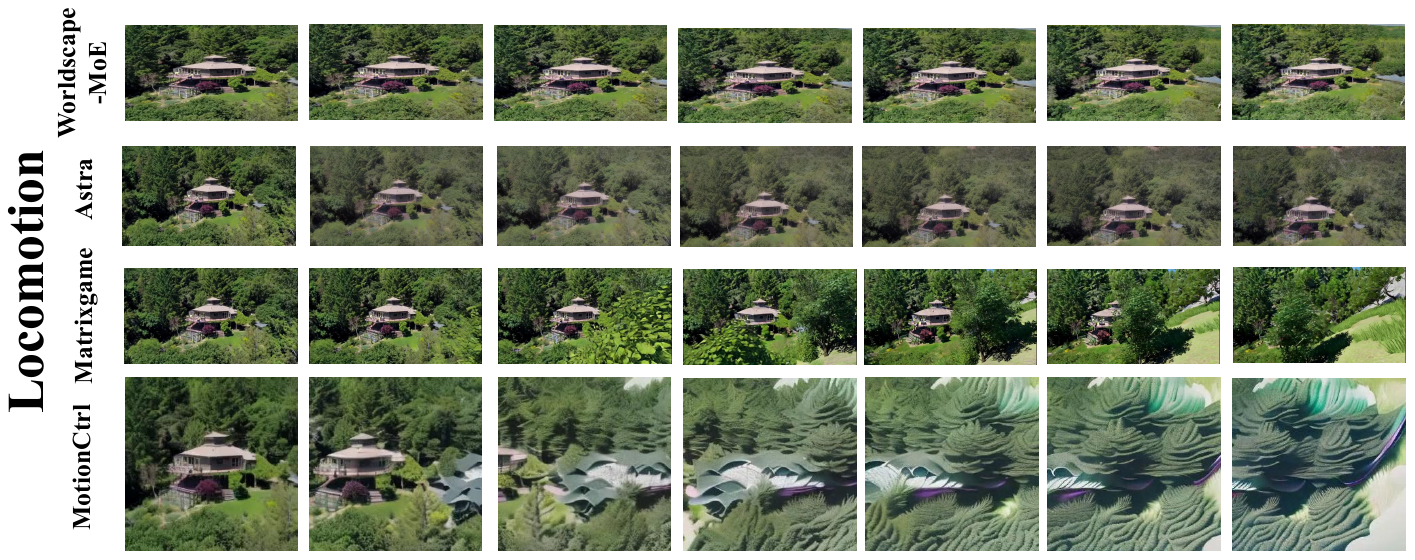}
    \caption{\textbf{Locomotion comparison.} Each row compares video frames generated by Worldscape-MoE, Astra, Matrixgame, and MotionCtrl under the same locomotion trajectory, with columns uniformly sampled from each rollout.}
	\label{fig:case1-3}
    \vspace{-5pt}
\end{figure}
\begin{figure}[htbp]
	\centering
	\includegraphics[width=\linewidth]{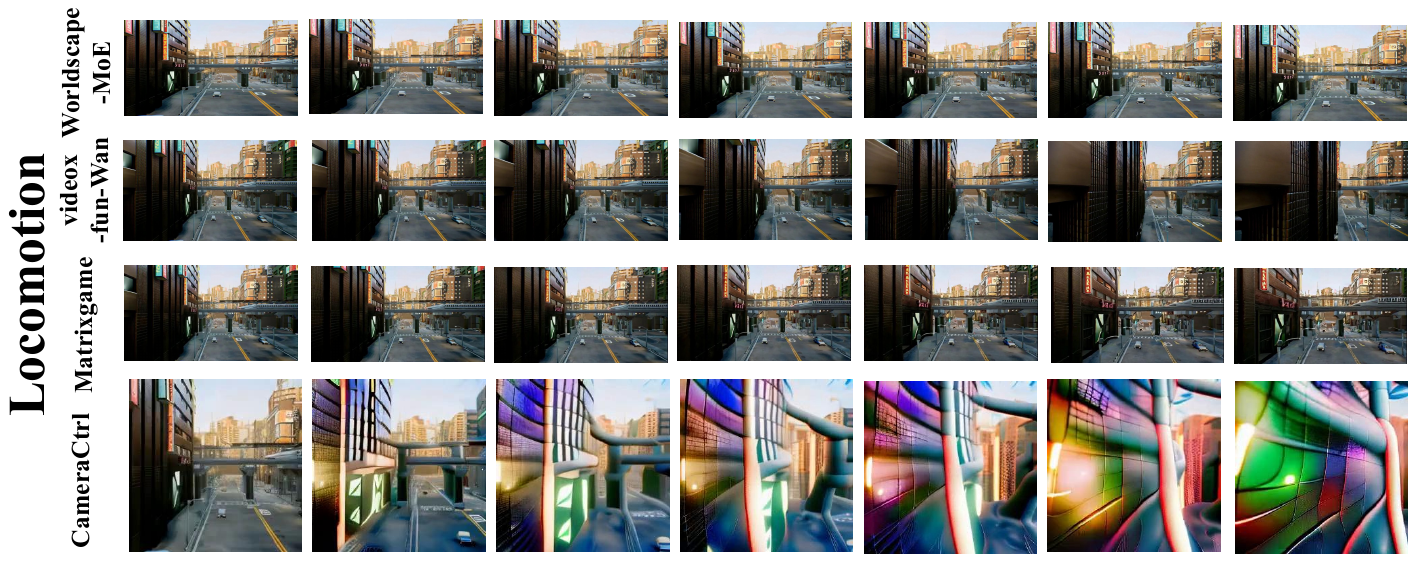}
    \caption{\textbf{Locomotion comparison.} Each row compares video frames generated by Worldscape-MoE, VideoX-fun-Wan, Matrixgame, and CameraCtrl under the same locomotion trajectory, with columns uniformly sampled from each rollout.}
	\label{fig:case1-4}
    \vspace{-5pt}
\end{figure}
\begin{figure}[htbp]
	\centering
	\includegraphics[width=\linewidth]{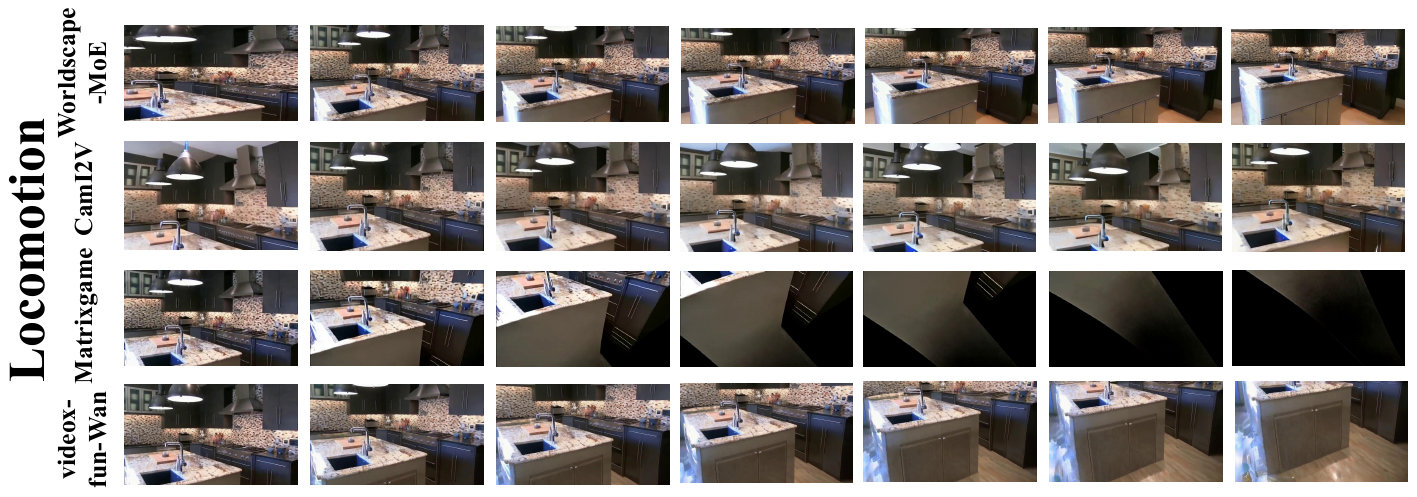}
    \caption{\textbf{Locomotion comparison.} Each row compares video frames generated by Worldscape-MoE, CamI2V, Matrixgame, and VideoX-fun-Wan under the same locomotion trajectory, with columns uniformly sampled from each rollout.}
	\label{fig:case1-5}
    \vspace{-5pt}
\end{figure}
\begin{figure}[htbp]
	\centering
	\includegraphics[width=\linewidth]{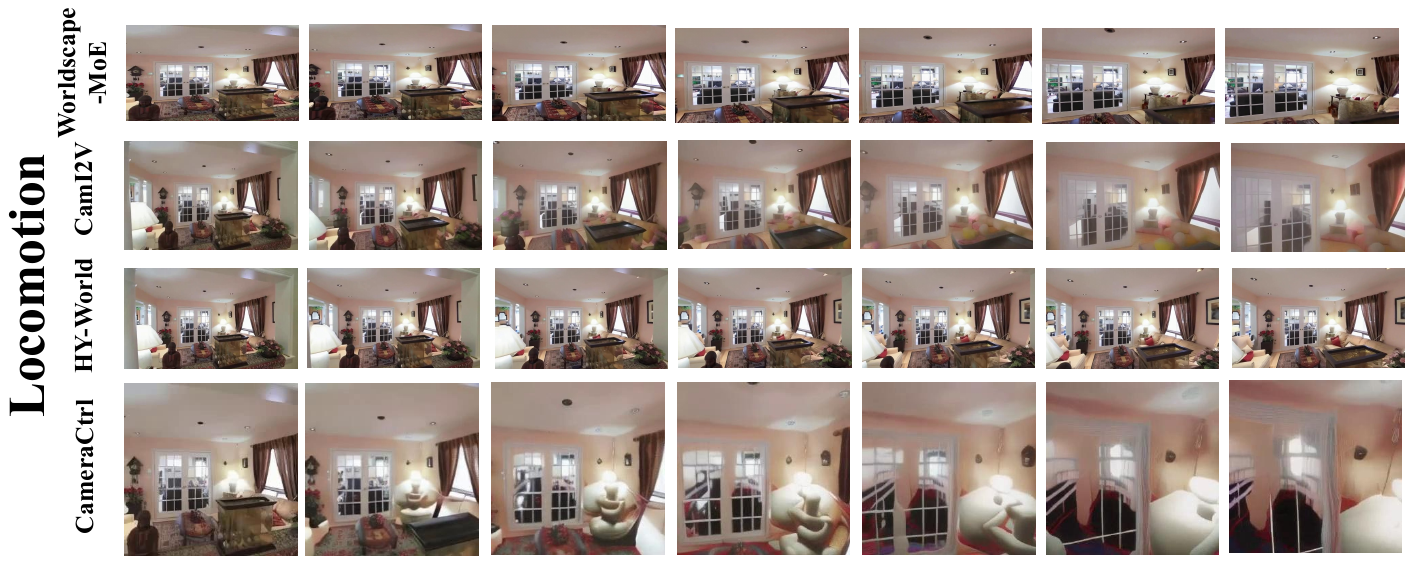}
    \caption{\textbf{Locomotion comparison.} Each row compares video frames generated by Worldscape-MoE, CamI2V, HY-World, and CameraCtrl under the same locomotion trajectory, with columns uniformly sampled from each rollout.}
	\label{fig:case1-6}
    \vspace{-5pt}
\end{figure}
\begin{figure}[htbp]
	\centering
	\includegraphics[width=\linewidth]{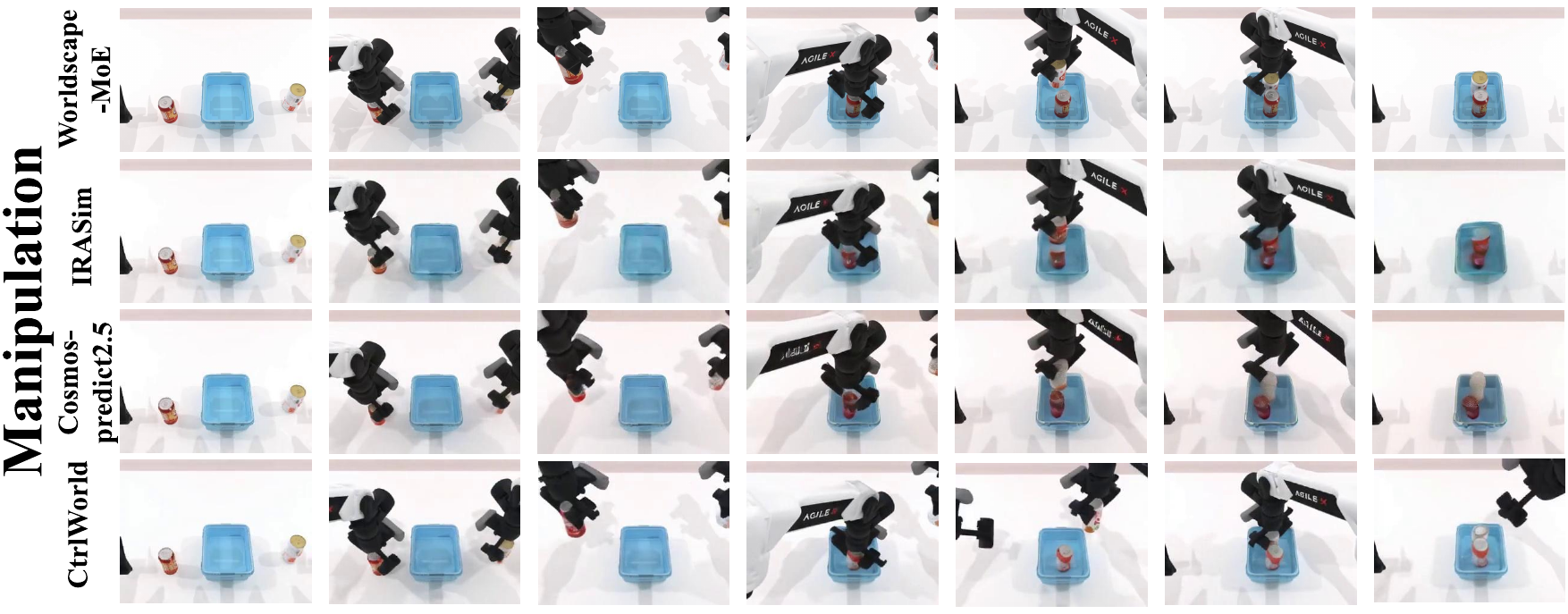}
    \caption{\textbf{Manipulation comparison.} Each row compares video frames generated by Worldscape-MoE, IRASim, Cosmos-predict2.5, and CtrlWorld under the same manipulation trajectory, with columns uniformly sampled from each rollout.}
	\label{fig:case2}
    \vspace{-5pt}
\end{figure}
\begin{figure}[htbp]
	\centering
	\includegraphics[width=\linewidth]{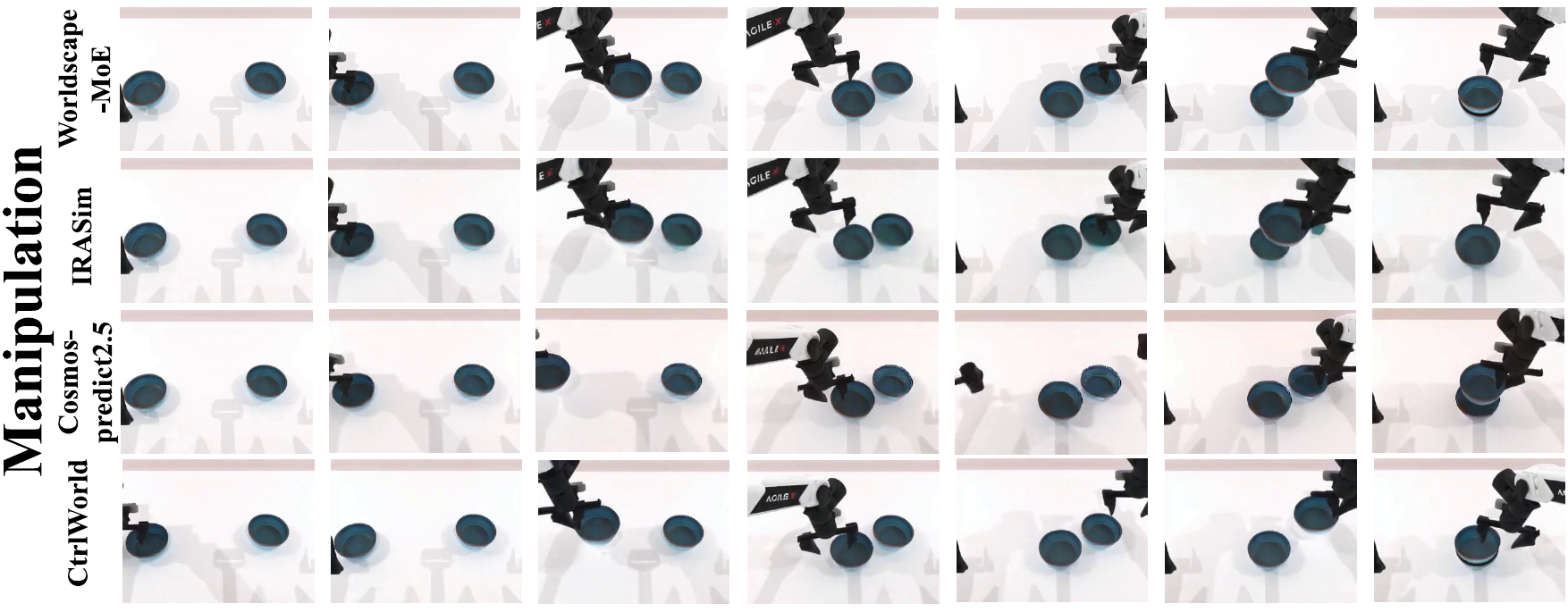}
    \caption{\textbf{Manipulation comparison.} Each row compares video frames generated by Worldscape-MoE, IRASim, Cosmos-predict2.5, and CtrlWorld under the same manipulation trajectory, with columns uniformly sampled from each rollout.}
	\label{fig:case2-2}
    \vspace{-5pt}
\end{figure}
\begin{figure}[htbp]
	\centering
	\includegraphics[width=\linewidth]{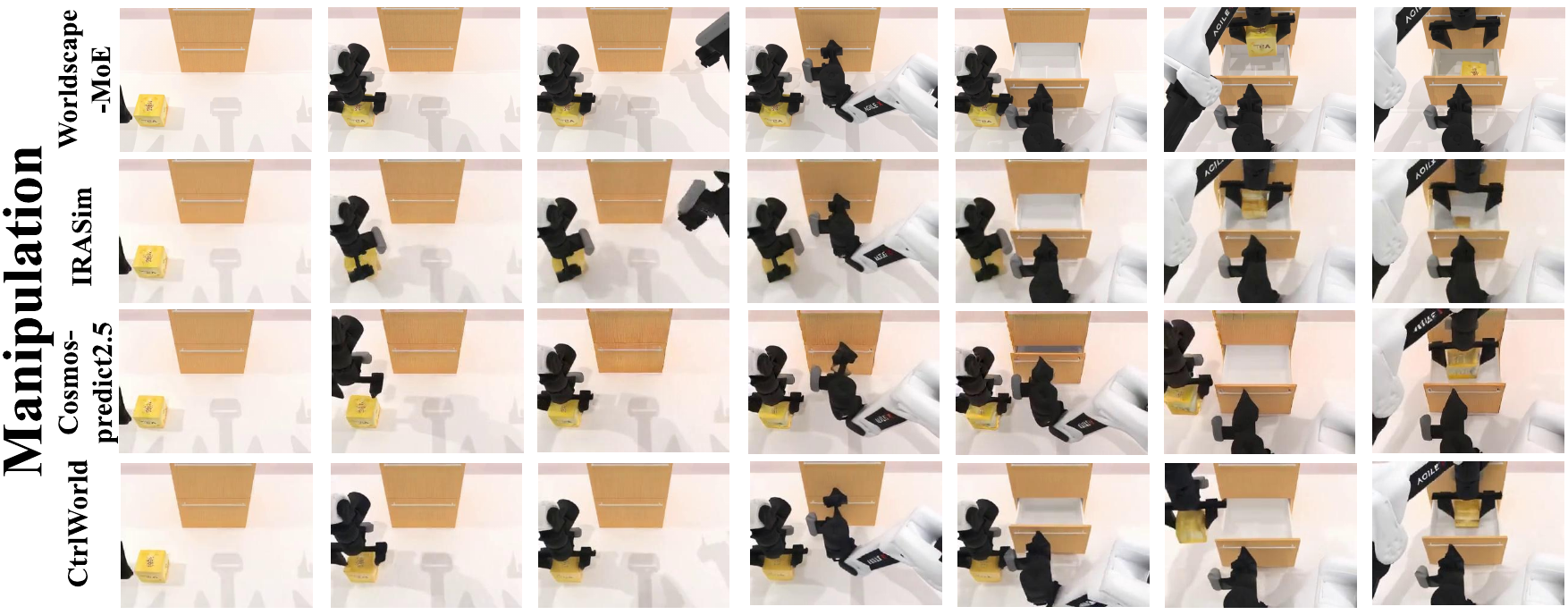}
    \caption{\textbf{Manipulation comparison.} Each row compares video frames generated by Worldscape-MoE, IRASim, Cosmos-predict2.5, and CtrlWorld under the same manipulation trajectory, with columns uniformly sampled from each rollout.}
	\label{fig:case2-3}
    \vspace{-5pt}
\end{figure}
\begin{figure}[htbp]
	\centering
	\includegraphics[width=\linewidth]{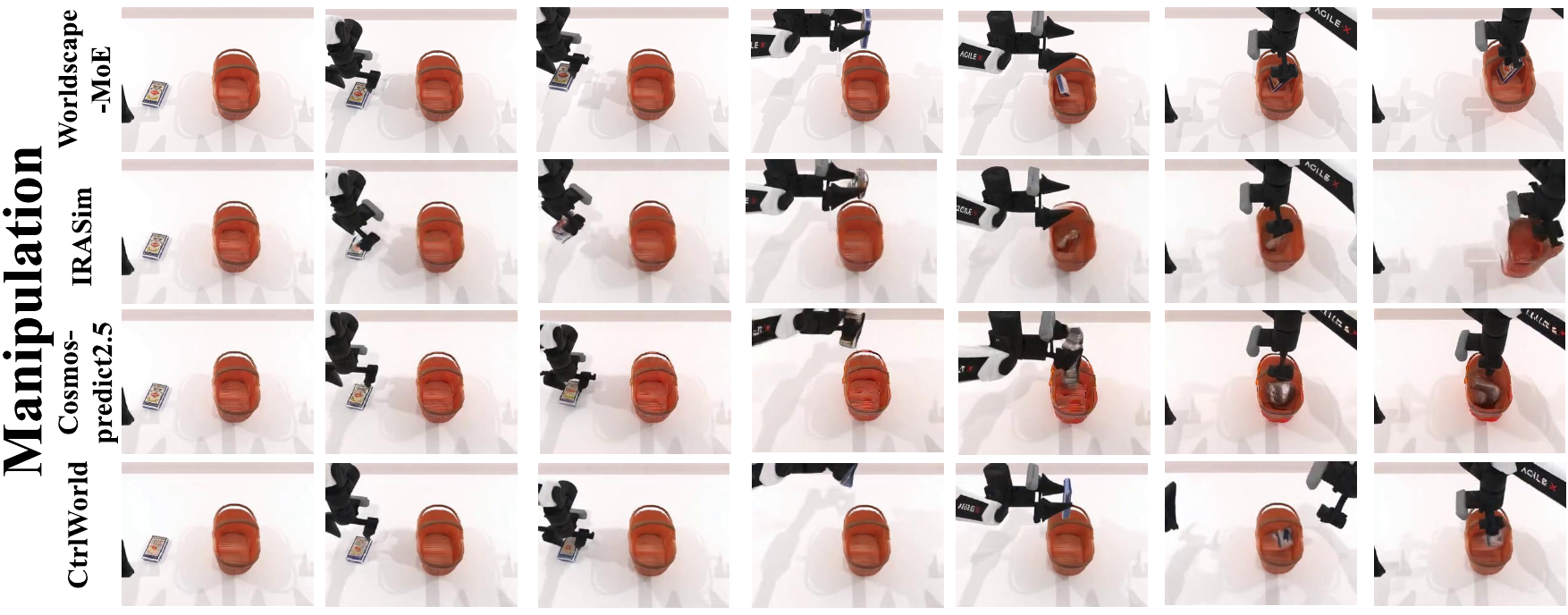}
    \caption{\textbf{Manipulation comparison.} Each row compares video frames generated by Worldscape-MoE, IRASim, Cosmos-predict2.5, and CtrlWorld under the same manipulation trajectory, with columns uniformly sampled from each rollout.}
	\label{fig:case2-4}
    \vspace{-5pt}
\end{figure}
\begin{figure}[htbp]
	\centering
	\includegraphics[width=\linewidth]{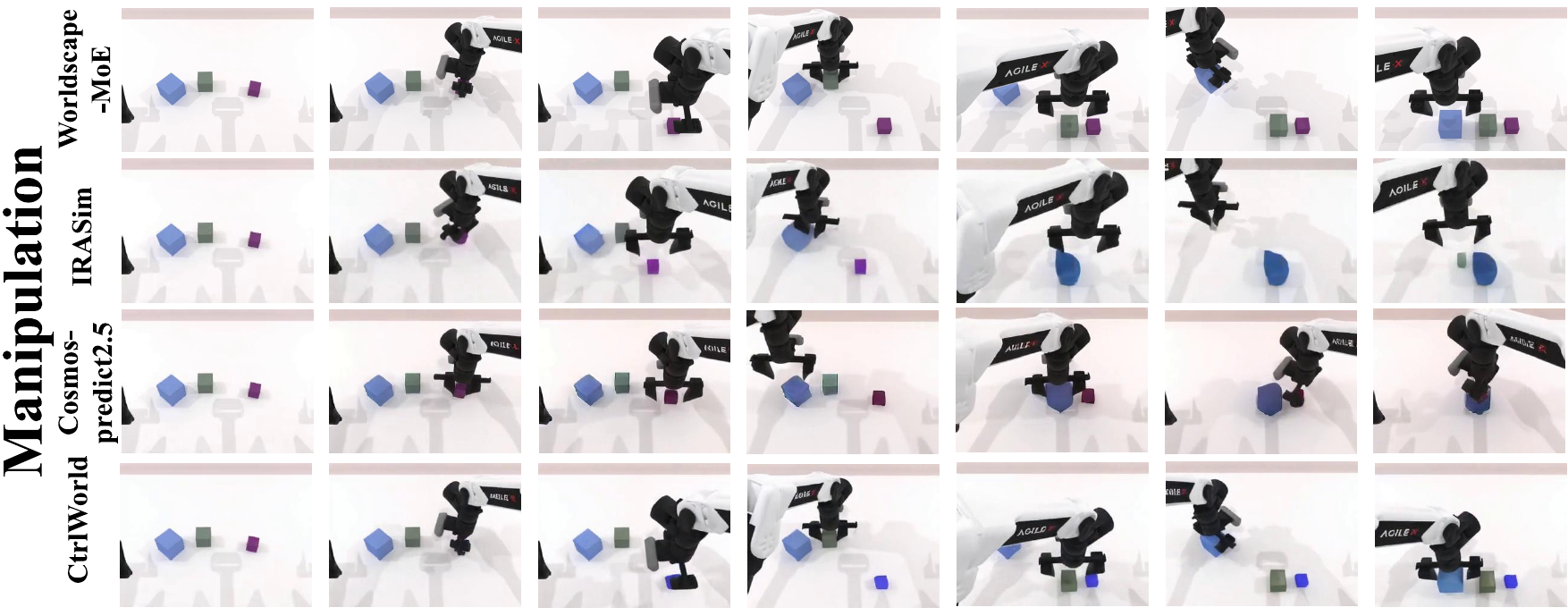}
    \caption{\textbf{Manipulation comparison.} Each row compares video frames generated by Worldscape-MoE, IRASim, Cosmos-predict2.5, and CtrlWorld under the same manipulation trajectory, with columns uniformly sampled from each rollout.}
	\label{fig:case2-5}
    \vspace{-5pt}
\end{figure}
\begin{figure}[htbp]
	\centering
	\includegraphics[width=\linewidth]{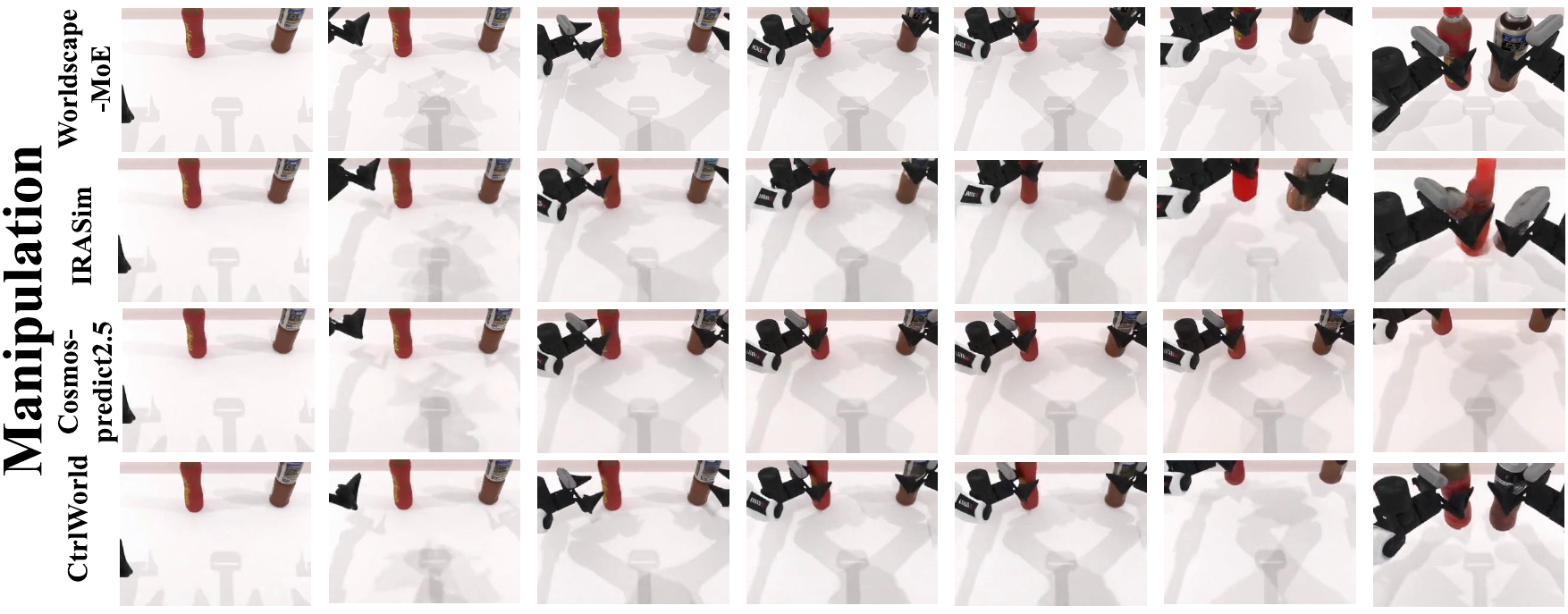}
    \caption{\textbf{Manipulation comparison.} Each row compares video frames generated by Worldscape-MoE, IRASim, Cosmos-predict2.5, and CtrlWorld under the same manipulation trajectory, with columns uniformly sampled from each rollout.}
	\label{fig:case2-6}
    \vspace{-5pt}
\end{figure}
\begin{figure}[htbp]
	\centering
	\includegraphics[width=\linewidth]{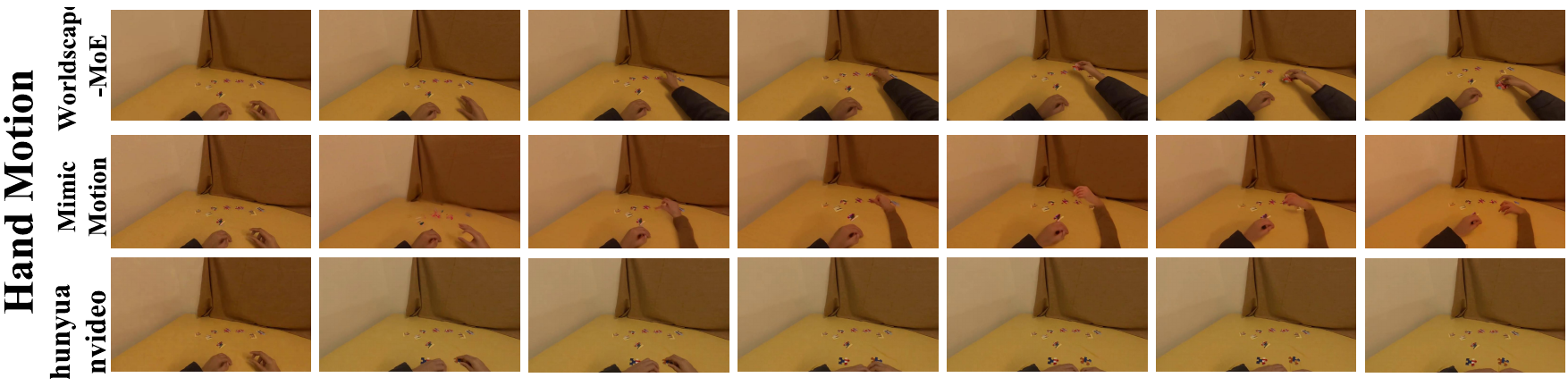}
    \caption{\textbf{Hand motion comparison.} Each row compares video frames generated by Worldscape-MoE, MimicMotion, and HunyuanVideo under the same hand motion trajectory, with columns uniformly sampled from each rollout.}
	\label{fig:case3}
    \vspace{-5pt}
\end{figure}
\begin{figure}[htbp]
	\centering
	\includegraphics[width=\linewidth]{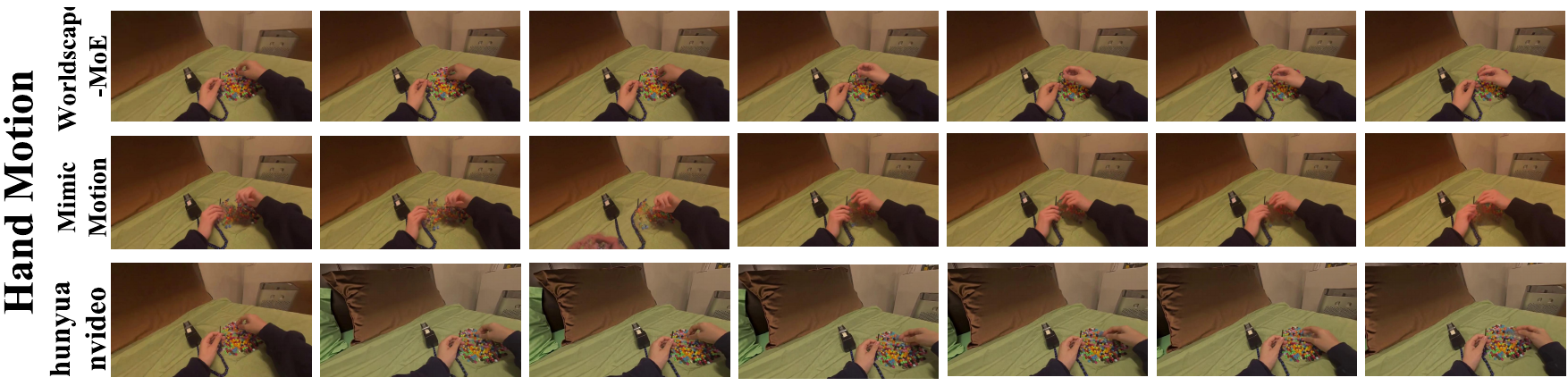}
    \caption{\textbf{Hand motion comparison.} Each row compares video frames generated by Worldscape-MoE, MimicMotion, and HunyuanVideo under the same hand motion trajectory, with columns uniformly sampled from each rollout.}
	\label{fig:case3-2}
    \vspace{-5pt}
\end{figure}
\begin{figure}[htbp]
	\centering
	\includegraphics[width=\linewidth]{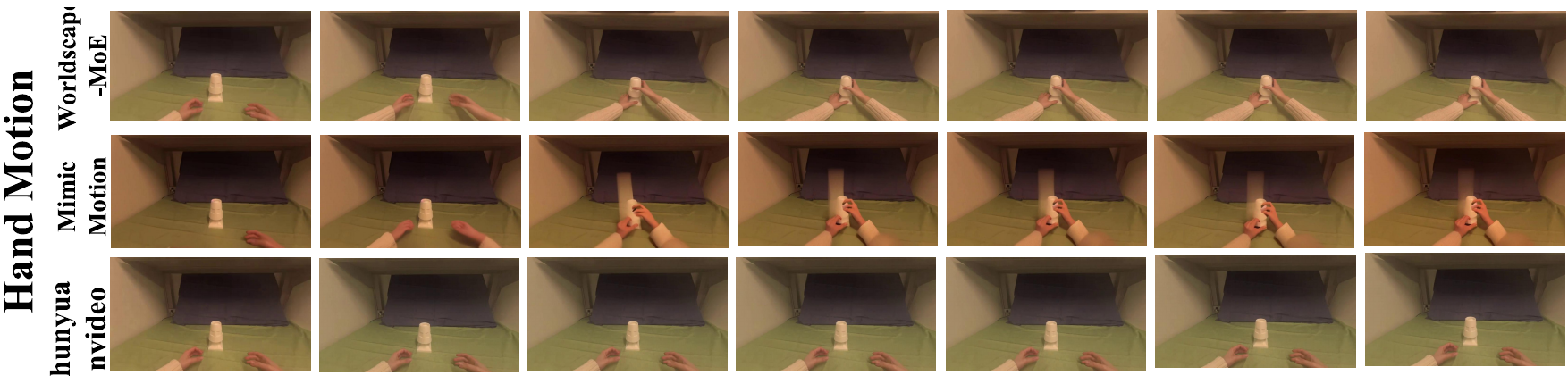}
    \caption{\textbf{Hand motion comparison.} Each row compares video frames generated by Worldscape-MoE, MimicMotion, and HunyuanVideo under the same hand motion trajectory, with columns uniformly sampled from each rollout.}
	\label{fig:case3-3}
    \vspace{-5pt}
\end{figure}
\begin{figure}[htbp]
	\centering
	\includegraphics[width=\linewidth]{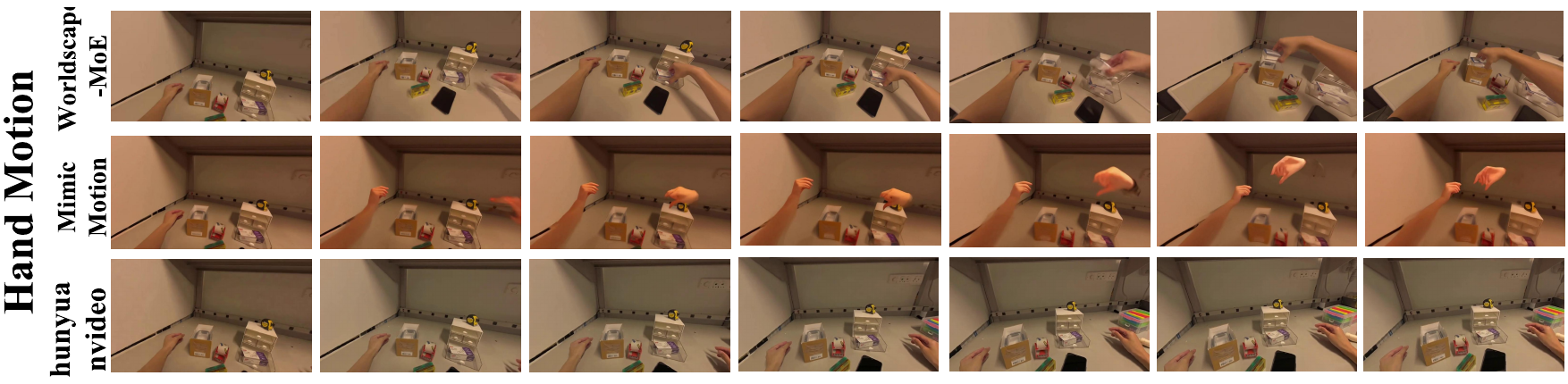}
    \caption{\textbf{Hand motion comparison.} Each row compares video frames generated by Worldscape-MoE, MimicMotion, and HunyuanVideo under the same hand motion trajectory, with columns uniformly sampled from each rollout.}
	\label{fig:case3-4}
    \vspace{-5pt}
\end{figure}
\begin{figure}[htbp]
	\centering
	\includegraphics[width=\linewidth]{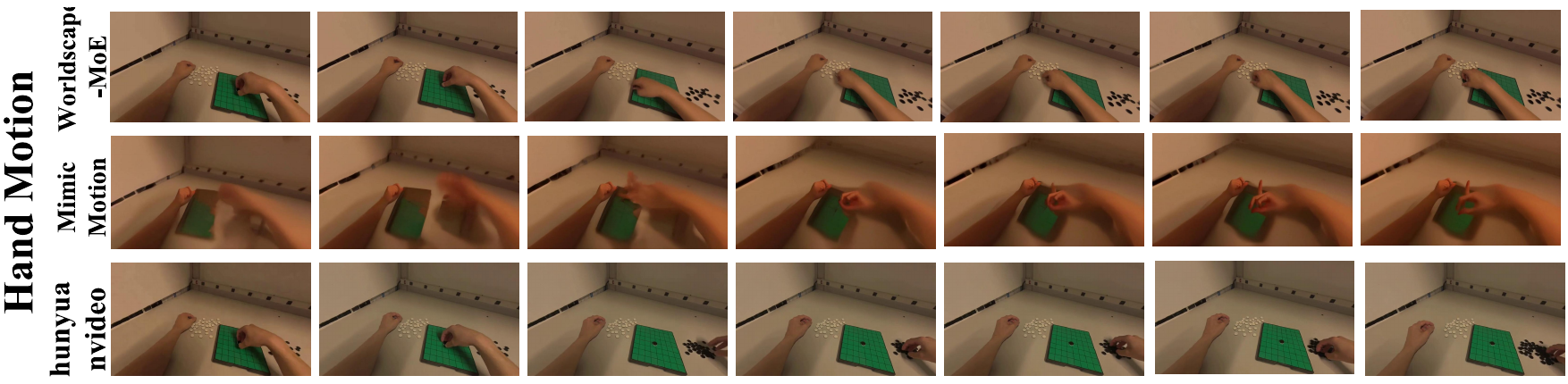}
    \caption{\textbf{Hand motion comparison.} Each row compares video frames generated by Worldscape-MoE, MimicMotion, and HunyuanVideo under the same hand motion trajectory, with columns uniformly sampled from each rollout.}
	\label{fig:case3-5}
    \vspace{-5pt}
\end{figure}
\begin{figure}[htbp]
	\centering
	\includegraphics[width=\linewidth]{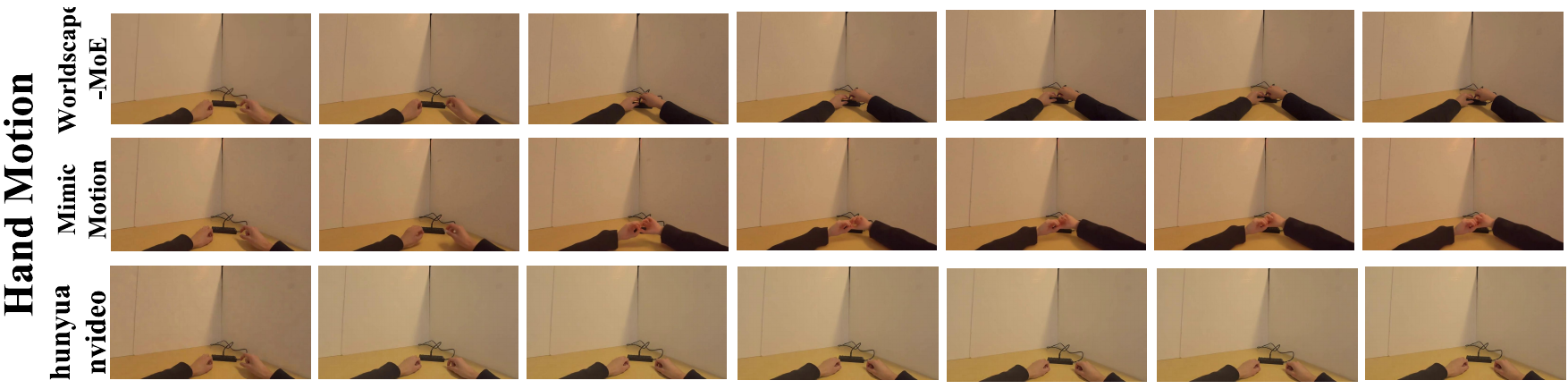}
    \caption{\textbf{Hand motion comparison.} Each row compares video frames generated by Worldscape-MoE, MimicMotion, and HunyuanVideo under the same hand motion trajectory, with columns uniformly sampled from each rollout.}
	\label{fig:case3-6}
    \vspace{-5pt}
\end{figure}
\begin{figure}[htbp]
	\centering
	\includegraphics[width=\linewidth]{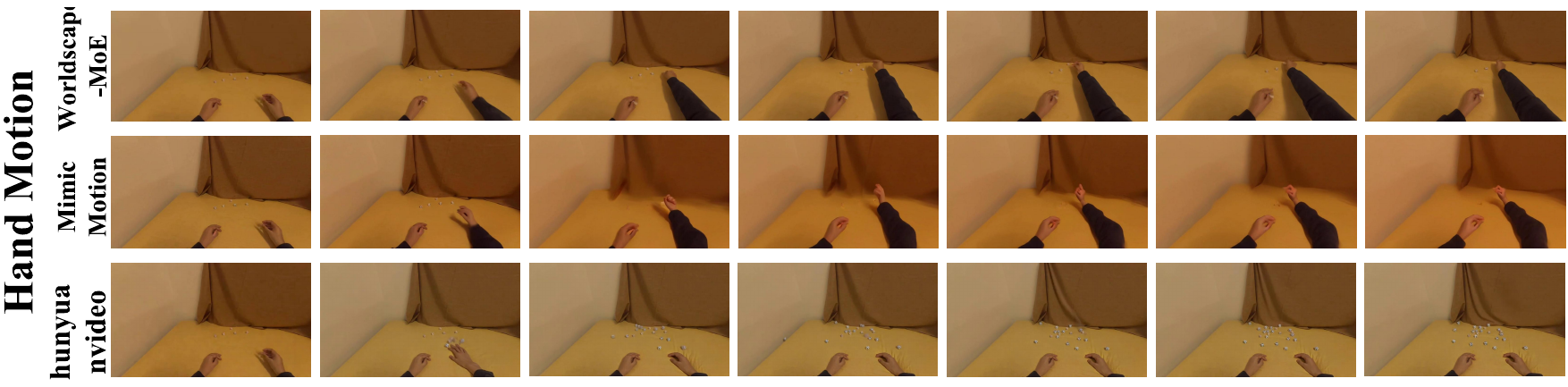}
    \caption{\textbf{Hand motion comparison.} Each row compares video frames generated by Worldscape-MoE, MimicMotion, and HunyuanVideo under the same hand motion trajectory, with columns uniformly sampled from each rollout.}
	\label{fig:case3-7}
    \vspace{-5pt}
\end{figure}

\begin{figure}[htbp]
	\centering
	\includegraphics[width=\linewidth]{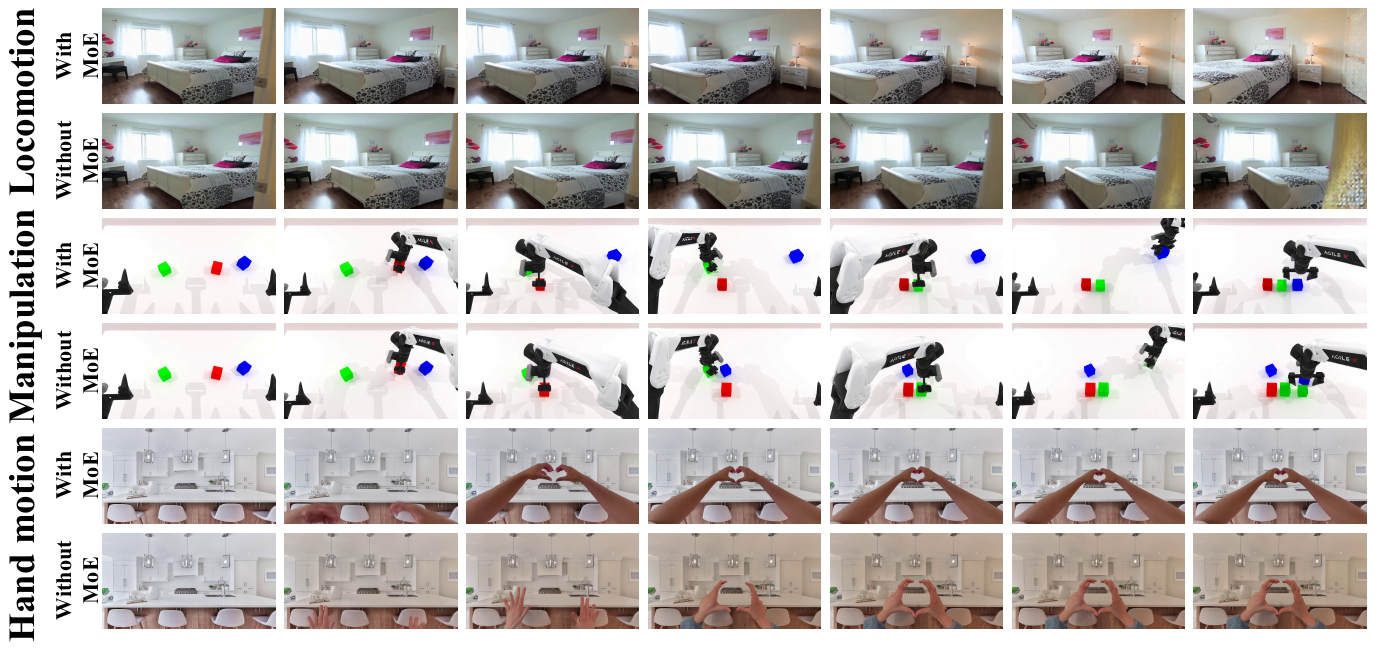}
    \caption{\textbf{Effect of MoE across diverse motion scenarios.} We compare generated rollouts with and without MoE under three representative settings: locomotion, manipulation, and hand motion. Each row shows uniformly sampled frames from one rollout, illustrating the effect of MoE on temporal consistency and motion-conditioned generation quality.}
	\label{fig:WOmoe}
    \vspace{-5pt}
\end{figure}

\begin{figure}[htbp]
	\centering
  \includegraphics[width=\linewidth]{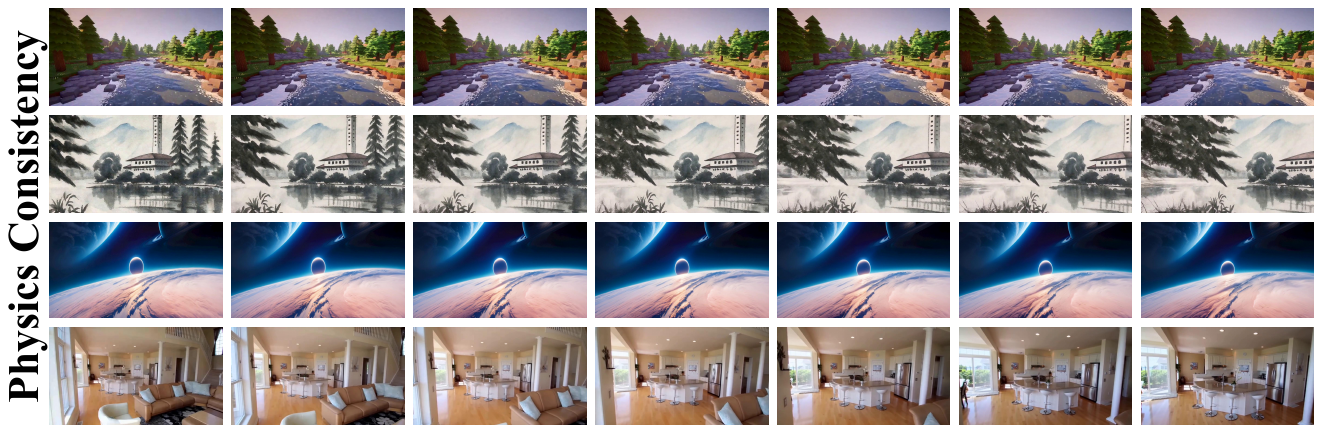}
    \vspace{-5pt}
    \caption{\textbf{Physics consistency across diverse visual domains.}
    The figure shows representative generated rollouts under different scene styles, including natural landscapes, ink-wash scenes, space environments, and indoor rooms. Each row contains uniformly sampled frames from one rollout, illustrating that the generated trajectories preserve temporal and physical consistency across visually distinct domains.}
    \label{fig:physics}
\end{figure}

\begin{figure}[htbp]
	\centering
	\includegraphics[width=\linewidth]{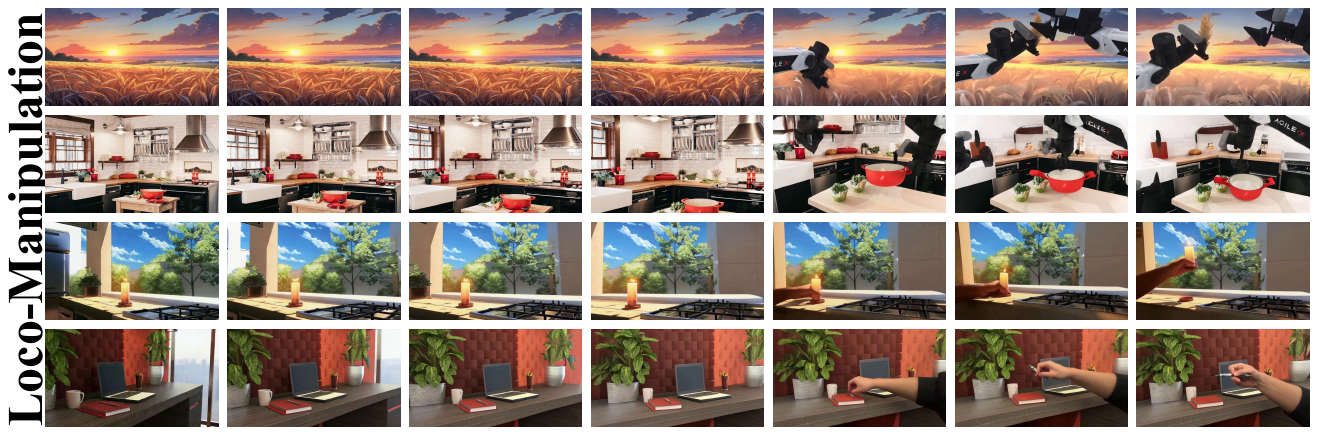}
    \caption{\textbf{Loco-manipulation generation across diverse scenes.}
The figure presents representative rollouts involving coordinated locomotion and object interaction. Each row contains uniformly sampled frames from one rollout, showing how the agent approaches and manipulates targets under different visual environments.}
    \label{fig:loco-mani}
    \vspace{-5pt}
\end{figure}

\end{document}